%% file: main.tex
\documentclass[numbers]{article}

\usepackage[final]{neurips_2024}
\usepackage{subfig}
\usepackage{natbib}
\usepackage{listings}
\usepackage{amsmath}

\usepackage[colorlinks,linkcolor=black]{hyperref}
\usepackage{bbding}
\usepackage[utf8]{inputenc} 
\usepackage[T1]{fontenc}    
\usepackage{hyperref}       
\usepackage{url}            
\usepackage{booktabs}       
\usepackage{amsfonts}       
\usepackage{nicefrac}       
\usepackage{microtype}      
\usepackage{xcolor}         
\usepackage{graphicx}
\usepackage{multirow}
\usepackage{array}
\usepackage{colortbl}
\usepackage{enumitem}

\title{Optimus-1\includegraphics[width=0.8cm]{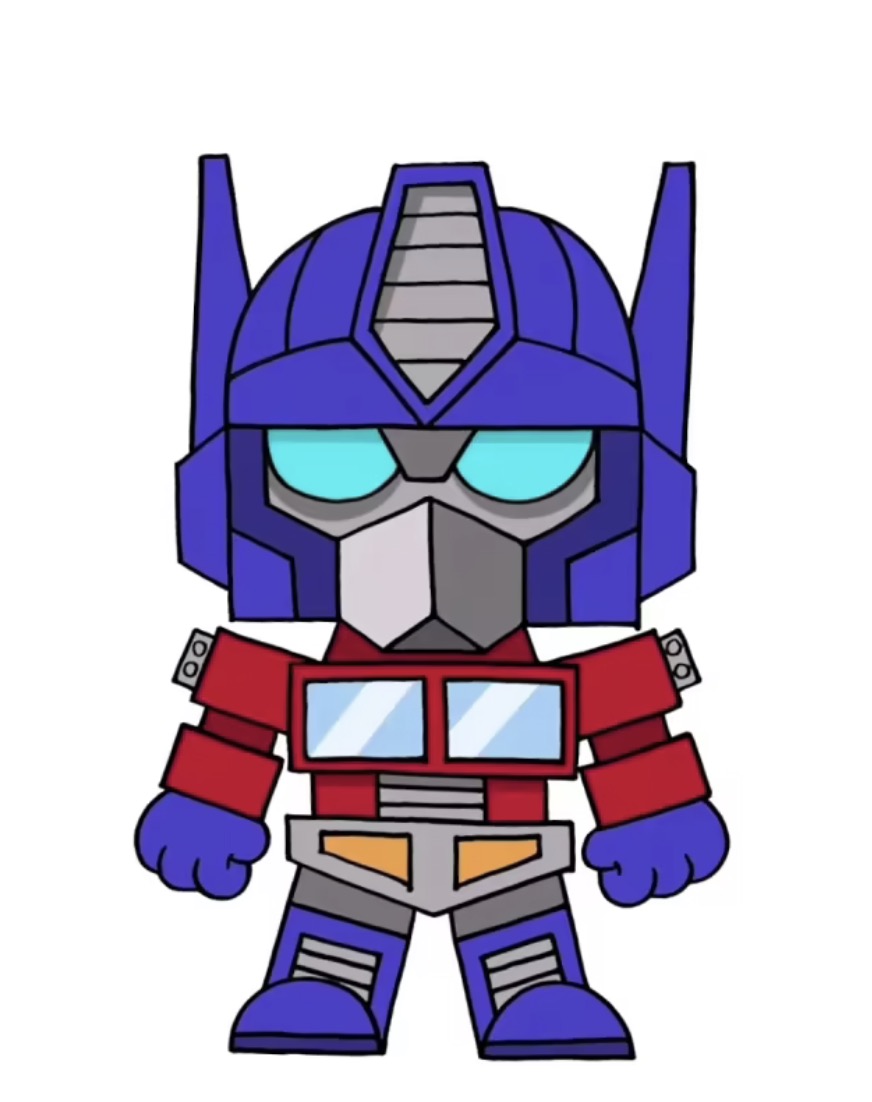}: Hybrid Multimodal Memory Empowered Agents Excel in Long-Horizon Tasks}

\author{
Zaijing Li$^{1\,2}$,
Yuquan Xie$^{1}$,
Rui Shao$^{1}$\footnotemark[1], Gongwei Chen$^{1}$, Dongmei Jiang$^{2}$, Liqiang Nie$^{1}$\footnotemark[1]
\\ 
    $^1$Harbin Institute of Technology, Shenzhen \\
      $^2$Peng Cheng Laboratory \\
       \texttt{\{lzj14011,xieyuquan20016,rshaojimmy,nieliqiang\}@gmail.com}
  }
\begin{document}

\maketitle
\renewcommand{\thefootnote}{\fnsymbol{footnote}} 
\footnotetext[1]{Corresponding authors}

\input{picture/fig-1}

\begin{abstract}
Building a general-purpose agent is a long-standing vision in the field of artificial intelligence. Existing agents have made remarkable progress in many domains, yet they still struggle to complete long-horizon tasks in an open world. We attribute this to the lack of necessary world knowledge and multimodal experience that can guide agents through a variety of long-horizon tasks. In this paper, we propose a \textbf{Hybrid Multimodal Memory} module to address the above challenges. It \textbf{1) }transforms knowledge into \textbf{Hierarchical Directed Knowledge Graph} that allows agents to explicitly represent and learn world knowledge, and \textbf{2) }summarises historical information into \textbf{Abstracted Multimodal Experience Pool} that provide agents with rich references for in-context learning. On top of the Hybrid Multimodal Memory module, a multimodal agent, Optimus-1, is constructed with dedicated \textbf{Knowledge-guided Planner} and \textbf{Experience-Driven Reflector}, contributing to a better planning and reflection in the face of long-horizon tasks in Minecraft. Extensive experimental results show that Optimus-1 significantly outperforms all existing agents on challenging long-horizon task benchmarks, and exhibits near human-level performance on many tasks. In addition, we introduce various Multimodal Large Language Models (MLLMs) as the backbone of Optimus-1. Experimental results show that Optimus-1 exhibits strong generalization with the help of the Hybrid Multimodal Memory module, outperforming the GPT-4V baseline on various tasks. Please see the project page at \href{https://cybertronagent.github.io/Optimus-1.github.io/}{https://cybertronagent.github.io/Optimus-1.github.io/}.
\end{abstract}

\section{Introduction}

Optimus Prime faces complex tasks alongside humans in Transformers to protect the peace of the planet. Creating an agent \cite{tan2024towards,huang2023embodied} like Optimus that can perceive, plan, reflect, and complete long-horizon tasks in an open world has been a longstanding aspiration in the field of artificial intelligence \cite{li2022emocaps,shao2019multi,shao2023detecting}. Early research developed simple policy through reinforcement learning \cite{fan2022minedojo} or imitation learning \cite{vpt,lifshitz2024steve}. A lot of work \cite{wang2023voyager,mc-planner} have utilized Large Language Models (LLMs) as action planners for agents, generating executable sub-goal sequences for low-level action controllers. Further, recent studies \cite{wang2023jarvis,qin2023mp5} employed Multimodal Large Language Models (MLLMs) \cite{chen2023lion,shen2024mome,ye2024cat} as planner and reflector. Leveraging the powerful instruction-following and logical reasoning capabilities of (Multimodal) LLMs \cite{li2024enhancing}, LLM-based agents have achieved remarkable success across multiple domains \cite{huang2022language,gramopadhye2023generating,gur2023real,yang2023appagent}. Nevertheless, the ability of these agents to complete long-horizon tasks still falls significantly short of human-level performance.

According to relevant studies \cite{makin2018amyloid,stuart2019comprehensive,vatansever2021varying}, the human ability to complete long-horizon tasks in an open world relies on long-term memory storage, which is divided into knowledge and experience. The storage and utilization of knowledge and experience play a crucial role in guiding human behavior and enabling humans to adapt flexibly to their environments in order to accomplish long-horizon tasks. Inspired by this theory, we summarize the challenges faced by current agents as follows:

\textbf{Insufficient Exploration of Structured Knowledge}: Structured knowledge, encompassing open world rules, object relationships, and interaction methods with the environment, is essential for agents to complete complex tasks \cite{raad2024scaling,tan2024towards}. However, MLLMs such as GPT-4V \footnote{https://openai.com/index/gpt-4v-system-card/} lack sufficient knowledge in Minecraft. Existing agents \cite{vpt,lifshitz2024steve,fan2022minedojo} only learn dispersed knowledge from video data and are unable to efficiently represent and learn this structured knowledge, rendering them incapable of performing complex tasks.

\textbf{Lack of Multimodal Experience}: Humans derive successful strategies and lessons from information on historical experience \cite{goddu2024development,parkes2019scientific}, which assists them in tackling current complex tasks. In a similar manner, agents can benefit from in-context learning with experience demonstrations \cite{sun2023generative,yang2024exploring}. 
However, existing agents \cite{wang2023voyager,wang2023describe,qin2023mp5} only consider unimodal information, which prevents them from learning from multimodal experience as humans do.

\renewcommand{\thefootnote}{\arabic{footnote}} 
To address the aforementioned challenges, we propose \textbf{Hybrid Multimodal Memory} module that consists of \textbf{Hierarchical Directed Knowledge Graph} (HDKG) and \textbf{Abstracted Multimodal Experience Pool} (AMEP). For HDKG, we map the logical relationships between objects into a directed graph structure, thereby transforming knowledge into high-level semantic representations. HDKG efficiently provides the agent with the necessary knowledge for task execution, without requiring any parameter updates. For AMEP, we dynamically summarize and store the multimodal information (\textit{e.g.,} environment, agent state, task plan, video frames, etc.) from the agent's task execution process, ensuring that historical information contains both a global overview and local details. Different from the method of directly storing successful cases as experience \cite{wang2023jarvis}, AMEP considers both successful and failed cases as references. This innovative approach of incorporating failure cases into in-context learning significantly enhances the performance of the agent.


On top of the Hybrid Multimodal Memory module, we construct a multimodal composable agent, \textbf{Optimus-1}. As shown in Figure \ref{fig1}, Optimus-1 consists of Knowledge-Guided Planner, Experience-Driven Reflector, and Action Controller. To enhance the ability of agents to cope with complex environments and long-horizon tasks, Knowledge-Guided Planner incorporates visual observation into the planning phase, leveraging HDKG to capture the knowledge needed. This allows the agent to efficiently transform tasks into executable sub-goals. Action Controller takes the sub-goal and the current observation as inputs and generates low-level actions, interacting with the game environment to update the agent's state. In open-world complex environments, agents are prone to be erroneous when performing long-horizon tasks. To address this, we propose Experience-Driven Reflector, which is periodically activated to retrieve relevant multimodal experiences from AMEP. This encourages the agent to reflect on its current actions and refine the plan.



We validate the performance of Optimus-1 in Minecraft, a popular open-world game environment. Experimental results show that Optimus-1 exhibits remarkable performance on long-horizon tasks, representing up to 30$\%$ improvement over existing agents. Moreover, we introduce various Multimodal Large Language Models (MLLMs) as the backbone of Optimus-1. Experimental results show that Optimus-1 has a 2 to 6 times performance improvement with the help of Hybrid Multimodal Memory, outperforming powerful GPT-4V baseline on lots of long-horizon tasks. Additionally, we verified that the plug-and-play Hybrid Multimodal Memory can drive Optimus-1 to incrementally improve its performance in a self-evolution manner. The extensive experimental results show that Optimus-1 makes a major step toward a general agent with a human-like level of performance. Main contributions of our paper:
\begin{itemize}[leftmargin=*]
    \item We propose \textbf{Hybrid Multimodal Memory} module which is composed of HDKG and AMEP. HDKG helps the agent make the planning of long-horizon tasks efficiently. AMEP provides refined historical experience and guides the agent to reason about the current situation state effectively.
    \item On top of the Hybrid Multimodal Memory module, we construct \textbf{Optimus-1}, which consists of \textbf{Knowledge-Guided Planner}, \textbf{Experience-Driven Reflector}, and Action Controller. Optimus-1 outperforms all baseline agents on long-horizon task benchmarks, and exhibits capabilities close to the level of human players.
    \item Driven by Hybrid Multimodal Memory, various MLLM-based Optimus-1 have demonstrated 2 to 6 times performance improvement, demonstrating the generalization of Hybrid Multimodal Memory.
\end{itemize}

\section{Optimus-1}
In this section, we first elaborate on how to implement the Hybrid Multimodal Memory in Sec 2.1. As a core innovation, it plays a crucial role in enabling Optimus-1 to execute long-horizon tasks. Next, we give an overview of Optimus-1 framework (Sec 2.2), which consists of Hybrid Multimodal Memory, Knowledge-Guided Planner, Experience-Driven Reflector, and Action Controller. Finally, we introduce a non-parametric learning approach to expand the hybrid multimodal memory (Sec 2.3), thereby enhancing the success rate of task execution for Optimus-1.

\subsection{Hybrid Multimodal Memory}
In order to endow agent with a long-term memory storage mechanism \cite{makin2018amyloid,vatansever2021varying}, we propose the Hybrid Multimodal Memory module, which consists of Abstracted Multimodal Experience Pool (AMEP) and Hierarchical Directed Knowledge Graph (HDKG). 

\subsubsection{Abstracted Multimodal Experience Pool}

Relevant studies \cite{li2023unisa,mitchell2021fast,li2023long,huang2023memory} highlight the importance of historical information for agents completing long-horizon tasks. Minedojo \cite{fan2022minedojo} and Voyager \cite{wang2023voyager} employed unimodal storage of historical information. Jarvis-1 \cite{wang2023jarvis} used a multimodal experience mechanism that stores task planning and visual information without summarization, posing challenges to storage capacity and retrieval speed. To address this issue, we propose AMEP, which aims to dynamically summarize all multimodal information during task execution. It preserves the integrity of long-horizon data while enhancing storage and retrieval efficiency.
\input{picture/fig-3}

 Specifically, as depicted in Figure \ref{fig3}, to conduct the static visual information abstraction, the video stream captured by Optimus-1 during task execution is first input to a video buffer, filtering the stream at a fixed frequency of 1 frame per second. Based on the filtered video frames, to further perform a dynamic visual information abstraction, these frames are then fed into an image buffer with a window size of 16, where the image similarity is dynamically computed and final abstracted frames are adaptively updated. To align such abstracted visual information with the corresponding textual sub-goal, we then utilize MineCLIP \cite{fan2022minedojo}, a pre-trained video-text alignment model, to calculate their multimodal correlation. When this correlation exceeds a threshold, the corresponding image buffer and textual sub-goal are saved as multimodal experience into a pool. Finally, we further incorporate environment information, agent initial state, and plan generated by Knowledge-Guided Planner, into such a pool, which forms the AMEP. In this way, we consider the multimodal information of each sub-goal, and summarise it to finally compose the multimodal experience of the given task.

\subsubsection{Hierarchical Directed Knowledge Graph}
In Minecraft, mining and crafting represent a complex knowledge network crucial for effective task planning. For instance, crafting a diamond sword \includegraphics[width=0.3cm]{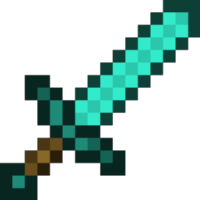} requires two diamonds \includegraphics[width=0.3cm]{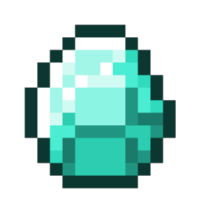} and one wooden stick \includegraphics[width=0.3cm]{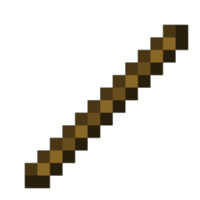}, while mining diamonds requires an iron pickaxe \includegraphics[width=0.3cm]{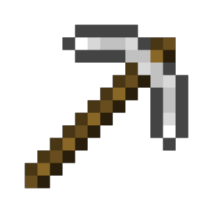}, which involving further materials and steps. Such knowledge is essential for an agent's ability to perform long-horizon complex tasks. Instead of implicit learning through fine-tuning \cite{qin2023mp5,zhou2024minedreamer}, we propose HDKG, which transforms knowledge into a graph representation. It enables the agent to perform explicit learning by retrieving information from the knowledge graph.

As shown in the Figure \ref{fig3}, we transform knowledge into a graph $\mathcal{D} (\mathcal{V},\mathcal{E})$, where nodes set $\mathcal{V}$ represent objects, and directed edges set $\mathcal{E}$ point to nodes that can be crafted by this object. An edge $e \in \mathcal{E}$ in the $\mathcal{D}$ can be represented as $e = (u, v)$, where $u, v \in \mathcal{V}$. The directed graph efficiently stores and updates knowledge. For a given object $x$, retrieving the corresponding node allows extraction of a sub-graph $\mathcal{D}_{j} (\mathcal{V}_{j}, \mathcal{E}_{j}) \in \mathcal{D}$, where nodes set $\mathcal{V}_{j}$ and edges set $\mathcal{E}_{j}$ can be formulated as:
\begin{equation}
 \mathcal{V}_{j} = \left \{ v \in \mathcal{V} \mid x \right \} , \; \; \; \; \; \;  \mathcal{E}_{j} = \left \{ e = (u,v) \in \mathcal{V} \mid u \in \mathcal{V}_{j} \cup  v \in \mathcal{V}_{j}\right \} ,
\end{equation}
Then by topological sorting, we can get all the materials and their relationships needed to complete the task. This knowledge is provided to the Knowledge-Guided Planner as a way to generate a more reasonable sequence of sub-goals. With HDKG, we can significantly enhance the world knowledge of the agent in a train-free manner.

\input{picture/fig-2}
\subsection{Optimus-1: Framework}
Relevant studies indicate that the human brain is essential for planning and reflection, while the cerebellum controls low-level actions, both crucial for complex tasks \cite{siddiqi2022causal,stein1986role}. Inspired by this, we divide the structure of Optimus-1 into Knowledge-Guided Planner, Experience-Driven Reflector, and Action Controller. In a given game environment with a long-horizon task, the Knowledge-Guided Planner senses the environment, retrieves knowledge from HDKG, and decomposes the task into executable sub-goals. The action controller then sequentially executes these sub-goals. During execution, the Experience-Driven Reflector is activated periodically, leveraging historical experience from AMEP to assess whether Optimus-1 can complete the current sub-goal. If not, it instructs the Knowledge-Guided Planner to revise its plan. Through iterative interaction with the environment, Optimus-1 ultimately completes the task.

\textbf{Knowledge-Guided Planner}. Open-world environments vary greatly, affecting task execution. Previous approaches \cite{wang2023describe} using LLMs for task planning failed to consider the environment, leading to the failure of tasks. For example, an agent in a cave aims to catch fish. It lacks visual information to plan conditions on the current situation, such as ``leave the cave and find a river''. Therefore, we integrate environmental information into the planning stage. Unlike Jarvis-1 \cite{wang2023jarvis} and MP5 \cite{qin2023mp5}, which convert observation to textual descriptions, Optimus-1 directly employs observation as visual conditions to generate environment-related plans, \textit{i.e.,} sub-goal sequences. This results in more comprehensive and reasonable planning. More importantly, Knowledge-Guided Planner retrieves the knowledge needed to complete the task from HDKG, allowing task planning to be done once, rather than generating the next step in each iteration. Given the task $t$, observation $o$, the sub-goals sequence $g_1$, $g_2$, $g_3$, ..., $g_n$ can be formulated as:
\begin{equation}
 g_1, g_2, g_3, ..., g_n =  p_\theta (o, t, p_\eta (t)),
\end{equation}
where $n$ is the number of sub-goals, $p_\eta$ denotes sub-graph retrieved from HDKG, $p_\theta$ denotes MLLM.
In this paper, we employ OpenAI’s GPT-4V as Knowledge-Guided Planner and Experience-Driven Reflector. We also evaluate other alternatives of GPT-4V, such as open-source models like Deepseek-VL \cite{lu2024deepseek} and InternLM-XComposer2-VL \cite{dong2024internlm} in Section \ref{sec:general}.

\textbf{Action Controller}. It takes the sub-goal and the current observation as inputs and then generates low-level actions, which are control signals for the mouse and keyboard. Thus, it can interact with the game environment to update the agent's state and the observation. The formulation is as follows:
\begin{equation}
 a_k =  p_\pi (o, g_i),
\end{equation}
where $a_k$ denotes low-level action at time $k$, $p_\pi$ denotes action controller. Unlike generating code \cite{wang2023voyager,qin2023mp5,mc-planner}, generating control actions for the mouse and keyboard \cite{vpt,lifshitz2024steve,wang2023jarvis,cai2023groot} more closely resembles human behavior. In this paper, we employ STEVE-1 \cite{lifshitz2024steve} as our Action Controller.

\textbf{Experience-Driven Reflector}. The sub-goals generated by Knowledge-Guided Planner are interdependent. The failure of any sub-goal halts the execution of subsequent ones, leading to overall task failure. Therefore, a reflection module is essential to identify and rectify errors promptly. During task execution, the Experience-Driven Reflector activates at regular intervals, retrieving historical experience from AMEP, and then analyzing the current state of Optimus-1. The reflection results of Optimus-1 are categorized as \texttt{COMPLETE}, \texttt{CONTINUE}, or \texttt{REPLAN}. \texttt{COMPLETE} indicates successful execution, prompting the action controller to proceed to the next sub-goal. \texttt{CONTINUE} signifies ongoing execution without additional feedback. \texttt{REPLAN} denotes failure, requiring the Knowledge-Guided Planner to revise the plan. The reflection $r$ generated by Experience-Driven Reflector can be formulated as:
\begin{equation}
 r =  p_\theta (o, g_i, p_\epsilon (t)),
\end{equation}
where $p_\epsilon$ denotes multimodal experience retrieved from AMEP. Experimental results in Section \ref{sec:ab} demonstrate that the Experience-Driven Reflector significantly enhances the success rate of long-horizon tasks. 

During task execution, even in cases where task failure necessitates \texttt{REPLAN}, multimodal experiences are stored in AMEP. Thus, during the reflection phase, Optimus-1 can retrieve the most relevant cases from each of the three scenarios \texttt{COMPLETE}, \texttt{CONTINUE}, and \texttt{REPLAN} from AMEP as references. Experimental Results in Section \ref{sec:ab} demonstrate the effectiveness of this innovative method of incorporating failure cases into in-context learning. 

\subsection{Non-parametric Learning of Hybrid Multimodal Memory}
\label{sec:hybrid-memory}
To implement the Hybrid Multimodal Memory and enhance Optimus-1's capacity, we propose a non-parametric learning method named ``free exploration-teacher guidance''. In the free exploration phase, Optimus-1's equipment and tasks are randomly initialized, and it explores random environments, acquiring world knowledge through environmental feedback. For example, it learns that ``a stone sword \includegraphics[width=0.3cm]{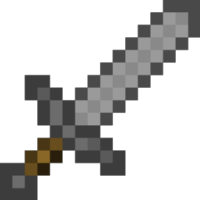} can be crafted with a wooden stick \includegraphics[width=0.3cm]{picture/logo/stick.pdf} and two cobblestones \includegraphics[width=0.3cm]{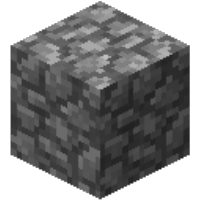}'', storing this in the HDKG. Additionally, successful and failed cases are stored in the AMEP, providing reference experience for the reflection phase. We initialize multiple Optimus-1, and they share the same HDKG and AMEP. Thus the memory is filled up efficiently. After free exploration, Optimus-1 has basic world knowledge and multimodal experience. In the teacher guidance phase, Optimus-1 needs to learn a small number of long-horizon tasks based on extra knowledge. For example, it learns ``a diamond sword \includegraphics[width=0.3cm]{picture/logo/diamond_sword.pdf} is obtained by a stick \includegraphics[width=0.3cm]{ picture/logo/stick.pdf} and two diamonds \includegraphics[width=0.3cm]{picture/logo/diamond.pdf}'' from the teacher, then perform the task ``craft diamond sword''. During the teacher guidance phase, Optimus-1's memory is further expanded and it gains the experience of executing complete long-horizon tasks. 

Unlike fine-tuning, this method enhances Optimus-1 incrementally without updating parameters, in a self-evolution manner. Starting with an empty Hybrid Multimodal Memory, Optimus-1 iterates between ``free exploration-teacher guidance'' learning and unseen task inference. With each iteration, its memory capacity grows, enabling mastery of tasks from easy to hard. 


\section{Experiments}
\subsection{Experiments Setting}
\textbf{Environment}.
To ensure realistic gameplay like human players, we employ MineRL \cite{guss2019minerl} with Minecraft 1.16.5 as our simulation environment. The agent operates at a fixed speed of 20 frames per second and only interacts with the environment via low-level action control signals of the mouse and keyboard. For more information about the detailed descriptions of the observation and action spaces, please refer to the \textbf{Appendix} \ref{mc}.

\input{table/tb_main}

\textbf{Benchmark}.
We constructed a benchmark of 67 tasks to evaluate the Optimus-1's ability to complete long-horizon tasks. As illustrated in Table \ref{tab:task}, we divide the 67 Minecraft tasks into 7 groups according to recommended categories in Minecraft. Please refer to \textbf{Appendix} \ref{bench} for more details.

\textbf{Baseline}.
We compare Optimus-1 with various agents, including GPT-3.5 \footnote{https://openai.com/research/gpt-3.5}, GPT-4V, DEPS \cite{wang2023describe}, and Jarvis-1 \cite{wang2023jarvis} on the challenging long-horizon tasks benchmark. In addition, we employed 10 volunteers to perform the same task on the benchmark, and their average performance served as a human-level baseline. Please refer to \textbf{Appendix} \ref{human} for more details about human-level baseline. For a more comprehensive comparison, we also report Optimus-1's performances on the benchmark used by Voyager \cite{wang2023voyager}, MP5 \cite{qin2023mp5}, and DEPS \cite{wang2023describe} in the \textbf{Appendix} \ref{other_bench}. Note that we initialize Optimus-1 with an empty inventory, while DEPS \cite{wang2023describe} and Jarvis-1 \cite{wang2023jarvis} have tools in their initial state. This makes it more challenging for Optimus-1 to perform the same tasks. 

\input{table/tb_ablation}

\input{picture/reflectiontex}
\textbf{Evaluation Metrics}.
The agent always starts in survival mode, with an empty inventory. We conducted at least 30 times for each task using different world seeds and reported the average success rate to ensure fair and thorough evaluation. Additionally, we add the average steps and average time of completing the task as evaluation metrics.

\subsection{Experimental Results}
The overall experimental results on benchmark are shown in Table \ref{tb_main}, see the accuracy for each task in \textbf{Appendix} \ref{experiment}. Optimus-1 has a success rate near 100$\%$ on the Wood Group \includegraphics[width=0.3cm]{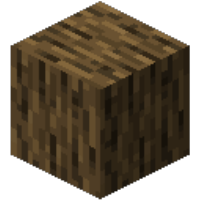}. Compared with Jarvis-1,  Optimus-1 has 29.28$\%$ and 53.40$\%$ improvement on the Diamond Group \includegraphics[width=0.3cm]{picture/logo/diamond.pdf} and Redstone Group \includegraphics[width=0.3cm]{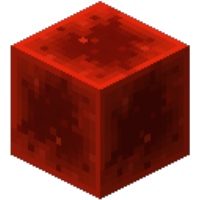}, respectively. Optimus-1 achieves the best performance and the shortest elapsed time among all task groups. It reveals the effectiveness and efficiency of our proposed Optimus-1 framework. Moreover, compared with all baselines, Optimus-1 performance was closer (average 5.37$\%$ improvement) to human levels on long-horizon task groups.

\input{picture/fig5}

\subsection{Ablation Study}
\label{sec:ab}
We conduct extensive ablation experiments on 18 tasks, experiment setting can be found in Table \ref{tab:abl_eval_task}. As shown in Table \ref{tb:ab_1}, we first remove Knowledge-Driven Planner and Experience-Driven Reflector, the performance of Optimus-1 on all task groups drops dramatically. It demonstrates the necessity of Knowledge-Guided Planner and Experience-Driven Reflector modules for performing long-horizon tasks. As for Hybrid Multimodal Memory, we remove HDKG from Optimus-1. Without the help of world knowledge, the performance of Optimus-1 decreased by an average of 20$\%$ across all task groups. We then removed AMEP, this resulted in the performance of Optimus-1 decreased by an average of 12$\%$. Finally, we performed ablation experiments on the way of retrieving cases from AMEP. As shown in Table \ref{tb:ab_2}, without retrieving cases from AMEP, the success rate shows an average of 10$\%$ decrease across all groups. It reveals that this reflection mechanism, which considers both success and failure cases, has a significant impact on the performance of Optimus-1. To illustrate the role of the reflection mechanism, we have shown some cases in Figure \ref{fig:fig-reflection}.

\subsection{Generalization Ability}
\label{sec:general}
In this section, we explore an interesting issue: whether generic MLLMs can effectively perform various long-horizon complex tasks in Minecraft using Hybrid Multimodal Memory. As shown in Figure \ref{fig4}, We employ Deepseek-VL \cite{lu2024deepseek} and InternLM-XComposer2-VL \cite{dong2024internlm} as Knowledge-Guided Planner and Experience-Driven Reflector. The experimental results show that the original MLLM has low performance on long-horizon tasks due to the lack of knowledge and experience of Minecraft. With the assistance of Hybrid Multimodal Memory, the performance of MLLMs has improved by 2 to 6 times across various task groups, outperforming the GPT-4V baseline. This encouraging result demonstrates the generalization of the proposed Hybrid Multimodal Memory.

\subsection{Self-Evolution via Hybrid Multimodal Memory}
\label{sec:evolution}
As shown in Section \ref{sec:hybrid-memory}, we  randomly initialize the Hybrid Multimodal Memory of Optimus-1, then update it multiple times by using the ``free exploration-teacher guidance'' learning method. We set the epoch to 4, and the number of learning tasks to 160. At each period, Optimus-1 performs free exploration on 150 tasks and teacher guidance learning on the remaining 10 tasks, we then evaluate Optimus-1's learning ability on the task groups same as ablation study. Experimental results are shown in Figure \ref{fig4}. It reveals that Optimus-1 keeps getting stronger through the continuous expansion of memory during the learning process of multiple periods. Moreover, it demonstrates that MLLM with Hybrid Multimodal Memory can incarnate an expert agent in a self-evolution manner \cite{tao2024survey}.

\section{Related Work}
\subsection{Agents in Minecraft}
We summarise the differences of existing Minecraft agents in the \textbf{Appendix} \ref{baseline}. Earlier work \cite{pmlr-v70-oh17a,yuan2023skill,cai2023open,cai2023groot} introduced policy models for agents to perform simple tasks in Minecraft. MineCLIP \cite{fan2022minedojo} used text-video data to train a contrastive video-language model as a reward model for policy, while VPT \cite{vpt} pre-trained on unlabelled videos but lacked instruction as input. Building on VPT and MineCLIP, STEVE-1 \cite{lifshitz2024steve} added text input to generate low-level action sequences from human instructions and images. However, these agents struggle with complex tasks due to limitations in instruction comprehension and planning. Recent work \cite{mc-planner,wang2023voyager,zhu2023ghost} incorporated LLMs as planning and reflection modules, but lacked visual information integration for adaptive planning. MP5 \cite{qin2023mp5}, MineDreamer \cite{zhou2024minedreamer}, and Jarvis-1 \cite{wang2023jarvis} enhanced situation-aware planning by obtaining textual descriptions of visual information, yet lacked detailed visual data. Optimus-1 addresses these issues by directly using observation as situation-aware conditions in the planning phase, enabling more rational, visually informed planning. Additionally, unlike other agents requiring multiple queries for task refinement, Optimus-1 generates a complete and effective plan in one step with the help of HDKG. This makes Optimus-1 planning more efficient.

\subsection{Memory in Agents}
In the agent-environment interaction process, memory is key to achieving experience accumulation \cite{li2024mamba}, environment exploration \cite{jiang2022vima}, and knowledge abstraction \cite{zhang2024survey}. There are two forms to represent memory content in LLM-based agents: textual form \cite{li2023long,huang2023memory,pal2023giraffe} and parametric form \cite{de2021editing,mitchell2021fast,wang2024detoxifying,li2024genview}. In textual form, the information is explicitly retained and recalled by natural languages. In parametric form, the memory information \cite{shao2024detecting} is encoded into parameters and implicitly influences the agent's actions. Recent work \cite{wang2024omnidrive,weng2024longvlm,he2024ma} has explored the long-term visual information storage \cite{li2023mask,li2023fine} and summarisation in MLLM. Our proposed hybrid multimodal memory module is plug-and-play and can provide world knowledge and multimodal experience for Optimus-1 efficiently.

\section{Conclusion}
In this paper, we propose Hybrid Multimodal Memory module, which consists of two parts: HDKG and AMEP. HDKG provides the necessary world knowledge for the planning phase of the agent, and AMEP provides the refined historical experience for the reflection phase of the agent. On top of the Hybrid Multimodal Memory, we construct the multimodal composable agent, Optimus-1, in Minecraft. Extensive experimental results show that Optimus-1 outperforms all existing agents on long-horizon tasks. Moreover, we validate that general-purpose MLLMs, based on Hybrid Multimodal Memory and without additional parameter updates, can exceed the powerful GPT-4V baseline. The extensive experimental results show that Optimus-1 makes a major step toward a general agent with a human-like level of performance.

\section{Limitation and Future Work}
In the framework of Optimus-1, we are dedicated to leverage proposed Hierarchical Directed Knowledge Graph and Abstracted Multimodal Experience Pool can be used to enhance the agent's ability to plan and reflect. For Action Controller, we directly introduce STEVE-1 \cite{lifshitz2024steve} as a generator of low-level actions. However, limited by STEVE-1's ability to follow instructions and execute complex actions, Optimus-1 is weak in completing challenging tasks such as ``beat ender dragon'' and ``build a house''. Therefore, a potential future research direction is to enhance the instruction following and action generation capabilities of action controller.

In addition, most of the work, including Optimus-1, utilize a multimodal large language model for planning and reflection, which then drives an action controller to perform the task. Building an end-to-end vision-language-action agent will be future work.

\section{Acknowledgement}
This study is supported by National Natural Science Foundation of China  (Grant No. 62236003 and 62306090), Shenzhen College Stability Support Plan (Grant No. GXWD20220817144428005), Natural Science Foundation of Guangdong Province of China (Grant No. 2024A1515010147), and Major Key Project of Peng Cheng Laboratory (Grant No. PCL2023A08).

\clearpage

\bibliography{Optimus}
\bibliographystyle{plainnat}
\clearpage

\appendix

\section{Broader Impact}
With the increasing capability level of Multimodal Large Language Models (MLLM) comes many potential benefits and also risks. On the positive side, we anticipate that the techniques that used to create Optimus-1 could be applied to the creation of helpful agents in robotics, video games, and the web. This plug-and-play architecture that we have created can be quickly adapted to different MLLMs, and the proposed methods also provide a viable solution for other application areas in the agent domain. However, on the negative side, it is imperative to acknowledge the inherent stochastic nature of MLLMs in text generation. If not addressed carefully, this could lead to devastating consequences for society. Prior to deploying MLLMs in conjunction with the Hybrid Multimodal Memory methodology, a comprehensive assessment of their potential risks must be undertaken. We hope that while the stakes are low, works such as ours can improve access to safety research on instruction-following models in multimodal agents domains.

\section{Minecraft}
\label{mc}

Minecraft is an extremely popular sandbox video game developed by Mojang Studios \footnote{https://www.minecraft.net/en-us/article/meet-mojang-studios}. It allows players to explore a blockly, procedurally generated 3D world with infinite terrain, discover and extract raw materials, craft tools and items, and build structures or earthworks (shown in Figure \ref{fig:mc-screen-shot}). Minecraft is a valuable and representative environment for evaluating long-horizon tasks, offering greater diversity and complexity compared to other environments. Unlike web/app navigation \cite{yang2023appagent} and embodied manipulation \cite{jiang2022vima}, Minecraft is an open world with a complex and dynamic environment (79 biomes, including ocean, plains, forest, desert, etc.). To complete long-horizon tasks, agents must achieve multiple sub-goals (e.g., 15 sub-goals to craft a diamond sword), making the construction of a Minecraft agent quite challenging. Many studies \cite{wang2023voyager,qin2023mp5,wang2023jarvis} have chosen Minecraft as the environment for validating performance on long-horizon tasks. Extensive experimental results in the paper show that Optimus-1 outperforms all baselines. Therefore, we chose Minecraft as open-world environment to evaluate the ability of agents to perform long-horizon tasks.

\input{picture/mc-intro}

\subsection{Basic Rules}

\textbf{Biomes.} The Minecraft world is divided into different areas called ``biomes''. Different biomes contain different blocks and plants and change how the land is shaped. There are 79 biomes in Minecraft 1.16.5, including ocean, plains, forest, desert, etc. Diverse environments have high requirements for the generalization of agents.

\textbf{Time.} Time passes within this world, and a game day lasts for 20 real-world minutes. Nighttime is much more dangerous than daytime: the game starts at dawn, and agents have 10 minutes of game time before nightfall. Hostile or neutral mobs spawn when night falls, and most of these mobs are dangerous, trying to attack agents. How to survive in such a dangerous world is an open problem for Minecraft agents research.

\textbf{Item.} In Minecraft 1.16.5, there are $975$ items can be obtained, such as wooden pickaxe \includegraphics[width=0.25cm]{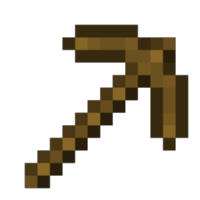}, iron sword \includegraphics[width=0.25cm]{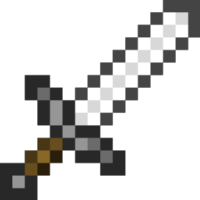}. Item can be obtained by crafting or destroying blocks or attacking entities. For example, agent can attack cows \includegraphics[width=0.25cm]{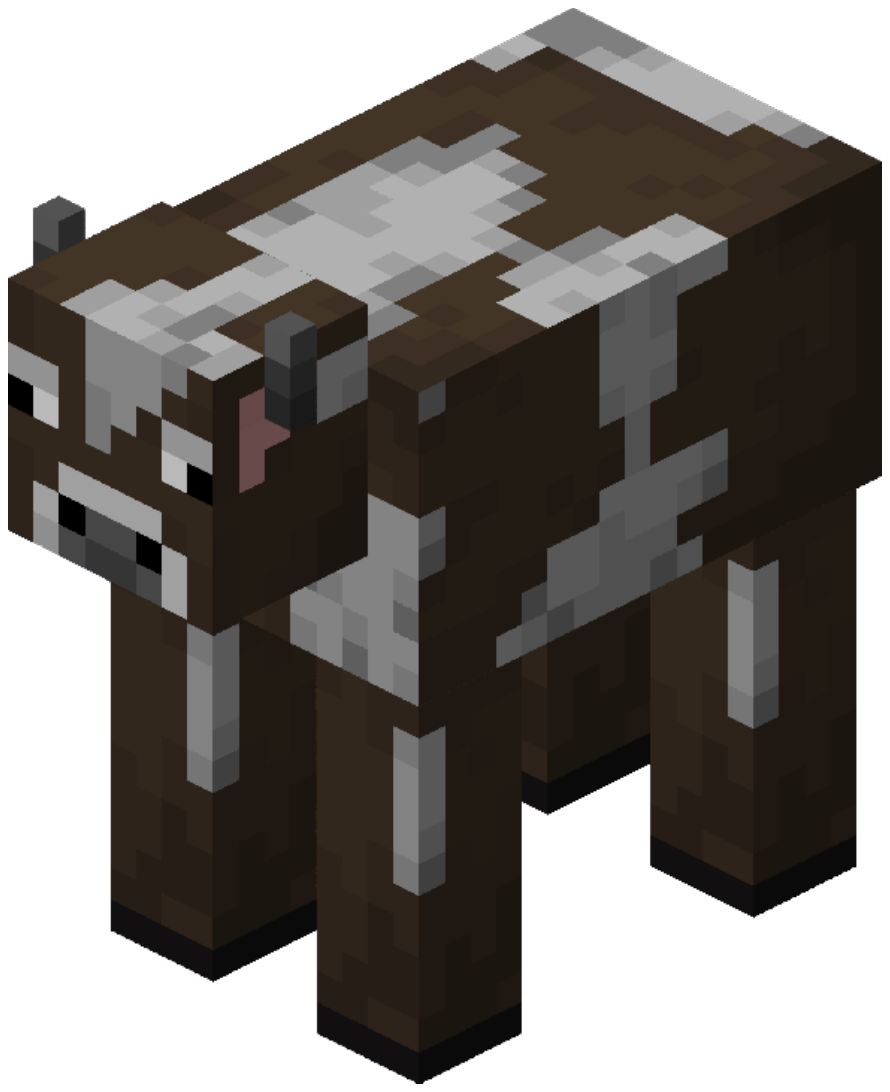}  to obtain leather \includegraphics[width=0.25cm]{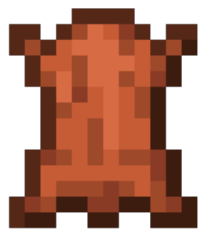} and beef \includegraphics[width=0.25cm]{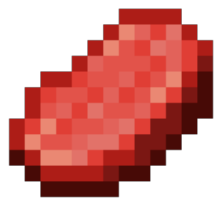}. Agent also can use $1$ stick \includegraphics[width=0.25cm]{picture/logo/stick.pdf} and $2$ diamonds \includegraphics[width=0.25cm]{picture/logo/diamond.pdf} to craft diamond sword \includegraphics[width=0.25cm]{picture/logo/diamond_sword.pdf}.

\textbf{Gameplay progress.} Progression primarily involves discovering and utilizing various materials and resources, each of which unlocks new capabilities and options. For instance, crafting a wooden pickaxe \includegraphics[width=0.25cm]{picture/logo/wooden_pickaxe.pdf} enables the player to mine stone \includegraphics[width=0.25cm]{picture/logo/cobblestone.pdf}, which can be used to create a stone pickaxe \includegraphics[width=0.25cm]{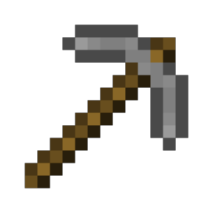} and a furnace \includegraphics[width=0.25cm]{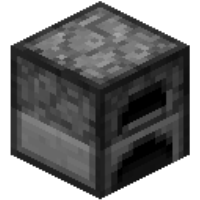}; these, in turn, allow for the mining and smelting of iron ore \includegraphics[width=0.25cm]{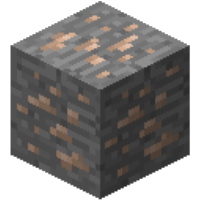}. Subsequently, an iron pickaxe \includegraphics[width=0.25cm]{picture/logo/iron_pickaxe.pdf} permits the extraction of diamonds \includegraphics[width=0.25cm]{picture/logo/diamond.pdf}, and a diamond pickaxe \includegraphics[width=0.25cm]{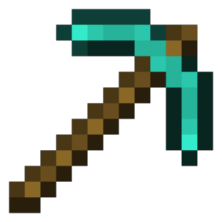} can mine virtually any block in the game. Similarly, cultivating different crops allows for the breeding of various animals, each providing distinct resources beyond mere sustenance. Enemy drops also have specific applications, with some being more beneficial than others. By integrating resources from mining, farming, and breeding, players can enchant their equipment. The collection and crafting of materials also facilitate construction, enabling players to build diverse structures. Beyond practical considerations such as secure bases and farms, the creative aspect of building personalized structures constitutes a significant part of the Minecraft experience.

\textbf{Freedom.} In Minecraft, player can do anything they can imagine. Player can craft tools, smelt ore, brew potions, trade with villagers and wandering traders, attack mobs, grow crops, raise animals in captivity, etc. Player even can use redstone \includegraphics[width=0.25cm]{picture/logo/readstone.pdf} to build a computer. This is a world of freedom and infinite possibilities.

\textbf{More Challenge than Diamond \includegraphics[width=0.25cm]{picture/logo/diamond.pdf}.} Progression beyond the Overworld is fairly limited: Eventually, you can build a nether portal to reach the Nether, where you can get materials for more complex crafting, the resources to brew potions, and the top tier of tools and armor. The Nether materials also let you reach the End dimension, where you must defeat the Ender Dragon to unlock the outer End Islands, where you can get an elytra that lets you fly, and shulker boxes for more storage.

\subsection{Observation and Action Spaces}

\textbf{Observation.}
Our observation space is completely consistent with human players. The agent only receives an RGB image with dimensions of $640 \times 360$ during the gameplay process, including the hotbar, health indicators, food saturation, and animations of the player's hands. It is worth helping the agent see more clearly in extremely dark environments, we have added a night vision effect for the agent, which increases the brightness of the environment during the night.

\textbf{Action Spaces.}
Our action space is almost similar to human players, except for craft and smelt actions. It consists of two parts: the mouse and the keyboard. The keypresses are responsible for controlling the movement of agents, such as jumping, forward, back, etc. The mouse movements are responsible for controlling the perspective of agents and the cursor movements when the GUI is opened. The left and right buttons of the mouse are responsible for attacking and using or placing items. In Minecraft, precise mouse movements are important when completing complex tasks that need open inventory or crafting table. In order to achieve both the same action space with MineDojo \cite{fan2022minedojo}, we abstract the craft and the smelt action into action space. The detailed action space is described in Table \ref{tab:action_space}.
\input{table/action_space}

\subsection{Long-horizon Tasks}

Long-horizon Tasks are complex tasks that require world knowledge to solve and consist of multiple indispensable subtask sequences. In Minecraft, technology has six levels, including wood \includegraphics[width=0.25cm]{picture/logo/wood.pdf}, stone \includegraphics[width=0.25cm]{picture/logo/cobblestone.pdf}, iron \includegraphics[width=0.25cm]{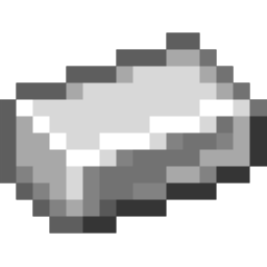}, golden \includegraphics[width=0.25cm]{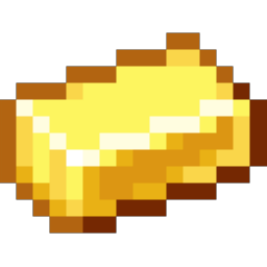}, diamond \includegraphics[width=0.25cm]{picture/logo/diamond.pdf}, and netherite \includegraphics[width=0.25cm]{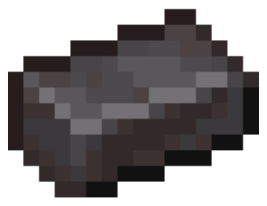}. Wooden tools can mine stone-level blocks, but can't mine iron-level and upper-level blocks. Stone tools can mine iron-level blocks, but can't mine diamond-level and upper-level blocks. Iron-level tools can mine diamond-level blocks, but can't mine netherite-level blocks. Diamond-level tools can mine any level blocks.

For example, the agent now wants to complete the task ``Craft iron sword \includegraphics[width=0.25cm]{picture/logo/iron_sword.pdf}''. The agent needs to craft wood-level tools to mine stone \includegraphics[width=0.25cm]{picture/logo/cobblestone.pdf}, and craft stone-level tools to mine iron ore \includegraphics[width=0.25cm]{picture/logo/iron_ore.pdf}. In order to craft tools, the agent needs a crafting table \includegraphics[width=0.25cm]{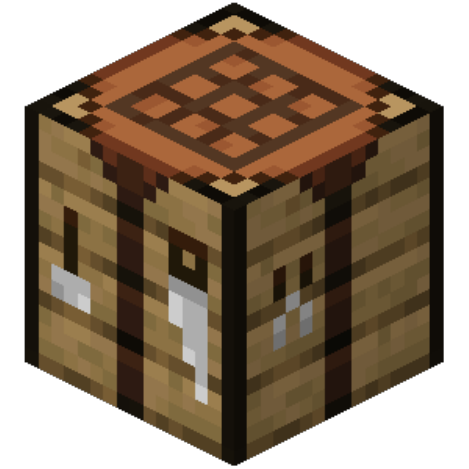}. To smelt iron ore \includegraphics[width=0.25cm]{picture/logo/iron_ore.pdf} into iron ingot \includegraphics[width=0.25cm]{picture/logo/iron_ingot.pdf}, the agent needs a furnace \includegraphics[width=0.25cm]{picture/logo/furnace.pdf}. Moreover, craft crafting table needs $4$ planks, and craft furnace needs $8$ cobblestone. In summary, the agent needs to obtain many raw materials,  wood-level and stone-level tools, $1$ crafting table, $1$ furnace, and most importantly, $2$ iron ingots. The process of this task is shown in Figure \ref{fig:example_long_horizon_task1}.

\input{picture/example_craft_iron_sword}

\section{Theory}
In this section, we briefly introduce the relevant theory of cognitive science. For more details, please refer to the original articles.

Our ability to understand and predict the world around us depends on our long-term memory stores, which have historically been divided into two distinct systems \cite{makin2018amyloid,stuart2019comprehensive,vatansever2021varying}. The semantic memory system provides a conceptual framework for describing the similar meanings of words and objects as they are encountered in different contexts (e.g., a bee is a flying insect with yellow and black stripes that produces honey), whereas the episodic memory system records our personal experiences characterized by the co-occurrence of words and objects at different times and places (e.g., being stung by a bee while eating honey at a picnic last weekend). These information stores and the interactions between them play a crucial role in guiding our behaviour and giving us the flexibility to adapt to the various demands of our environment.

In this paper, inspired by the above theory, we divide the agent memory module into two parts: knowledge and experience. Based on this, we propose Hierarchical Directed Knowledge Graph and Abstracted Multimodal Experience Pool to enable the agent to acquire, store, and utilize knowledge and experience during the execution of tasks.
Extensive experimental results demonstrate the effectiveness of the proposed methodology
\section{Benchmark Suite}
\label{bench}
\subsection{Benchmark}
We constructed a benchmark of 67 tasks to evaluate Optimus-1's ability to complete long-horizon tasks in Minecraft. According to recommended categories in Minecraft, we have classified these tasks into 7 groups: Wood \includegraphics[width=0.25cm]{picture/logo/wood.pdf}, Stone \includegraphics[width=0.25cm]{picture/logo/cobblestone.pdf}, Iron \includegraphics[width=0.25cm]{picture/logo/iron_ingot.pdf}, Gold \includegraphics[width=0.25cm]{picture/logo/gold_ingot.pdf}, Diamond \includegraphics[width=0.25cm]{picture/logo/diamond.pdf}, Redstone \includegraphics[width=0.25cm]{picture/logo/readstone.pdf}, Armor \includegraphics[width=0.25cm]{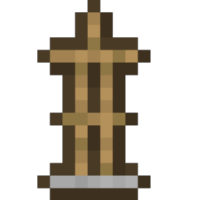}. The statistics for benchmark are shown in Table \ref{tab:task}. Due to the varying complexity of these tasks, we adopt different maximum gameplay steps (Max. Steps) for each task. The maximum steps are determined by the average steps that human players need to complete the task. Due to the randomness of Minecraft, the world and initial spawn point of the agent could vary a lot. In our benchmark setting, We initialize the agent with an empty inventory, which makes it necessary for the agent to complete a series of sub-goals (mining materials, crafting tools) in order to perform any tasks. This makes every task challenging, even for human players.

Note that Diamonds are a very rare item that only spawns in levels 2 to 16 and have a 0.0846$\%$ chance of spawning in Minecraft 1.16.5. Diamonds are usually found near level 9, or in man-made or natural mines no higher than level 16. In order to reduce the huge impact that diamond generation probability has on agent's likelihood of completing a task, we have adjusted the diamond generation probability to 20$\%$, spawns in levels 2 to 16. This setting applies to human players as well.

\input{table/eval_task}
In the ablation study, we select the subset of our benchmark as the test set (shown in Table \ref{tab:abl_eval_task}). The environment setting is the same as the benchmark.
\input{table/abl_eval_task}

\subsection{Baselines}
\label{human}

\textbf{Existing Baseline.} On the one hand, we employ GPT-3.5 and GPT-4V as baseline, which are evaluated without integrating hybrid multimodal memory modules. During the planning phase, they generate a plan for the action controller based on task prompt (and observation). During the reflection phase, they generate reflection results in a zero-shot manner. On the other hand, we compare existing SOTA Agents \cite{wang2023describe,wang2023jarvis} in Minecraft.

\textbf{Human-level Baseline.} To better demonstrate agent's performance level in Minecraft, we hired 10 volunteers to play the game as a human-level baseline. The volunteers played the game with the same environment and settings, and every volunteer asked to perform the each task on the benchmark 10 times. Ultimately, we used the average scores of 10 volunteers as the human-level baseline. The results of the human-level baseline are shown in Table \ref{tb_main}. To ensure the validity of the experiment, we ensured that each volunteer had at least 20 hours of Minecraft gameplay before conducting the experiment. For each volunteer, we pay $\$25$ as reward. 

\subsection{Minecraft Agents}
\label{baseline}
In this section, we summarise the differences between existing Minecraft agents. As shown in the Table \ref{tab:compare_baseline}, earlier work \cite{vpt,fan2022minedojo,lifshitz2024steve,cai2023groot} constructed Transformer-based policy network as agent. Recent work \cite{wang2023voyager,wang2023describe,wang2023jarvis,qin2023mp5} introduces the Multimodal Large Language Model, which empowers the agent to complete long-horizon tasks by exploiting the powerful language comprehension and planning capabilities of LLM. 

In the Mineflayer and Minedojo environments, agents \cite{fan2022minedojo,wang2023voyager,wang2023describe,qin2023mp5} can accomplish sub-goals by calling APIs (in the form of codes), which is a different behavioral pattern from humans. In MineRL \cite{guss2019minerl}, agents \cite{vpt,lifshitz2024steve,cai2023groot,wang2023jarvis} must generate low-level actions to perform tasks, which is more challenging to accomplish long-horizon tasks.

Moreover, existing agents lack knowledge and experience, and their performance in Minecraft is still vastly gapped from the human level. In this paper, we introduce Hybrid Multimodal Memory, which empowers Optimus-1 with hierarchical knowledge and multimodal experience. This makes Optimus-1 significantly outperform all existing agents on challenging long-horizon tasks benchmark, and exhibits near human-level performance on many tasks.

\input{table/compare_baseline}

\section{Implementation Details}
\subsection{Hybrid Multimodal Memory}
\subsubsection{Abstracted Multimodal Experience Pool}
Relevant studies \cite{de2021editing,mitchell2021fast,li2023long,huang2023memory} have demonstrated the importance of memory for agents to complete long-horizon tasks. To implement the memory mechanism, Minedojo \cite{fan2022minedojo} and Voyager \cite{wang2023voyager} only considered unimodal storage of historical information. Jarvis-1 \cite{wang2023jarvis} considered a multimodal memory mechanism to store task planning and visual information as experience, but it stores all historical information without summarisation. This approach stores all visual images, which poses a huge challenge in storage size and retrieval efficiency. To solve the problem, we propose the Abstracted Multimodal Experience Pool structure, which summarizes all historical information during the agent's execution of the task, which maintains the integrity of long sequential information and greatly improves the storage and retrieval efficiency of the experience.

As shown in Figure \ref{fig3}, we first input the visual image stream to the video buffer, which filters the image stream at a fixed frequency. It makes the length of the image stream substantially shorter. Empirically, we set the frequency of filtering to 1 second/frame, meaning that the video buffer takes one frame per second from the original image stream to compose the filtered image stream. We found that above this frequency makes the visual information redundant (too much similarity between images), and below this frequency does not preserve enough complete visual information. 

Then, we feed the filtered frames into an image buffer with a window size of 16. We dynamically compute the similarity between images in the image buffer, when a new image comes in, we compute the similarity between the new image and the most recent image, and then we remove the image with the highest similarity in order to keep the image buffer's window size to 16. 

Subsequently, we introduce MineCLIP \cite{fan2022minedojo}, a pre-trained model of video-text alignment with a structure similar to CLIP \cite{radford2021learning}, as our visual summariser. For a given sub-goal, it calculates the correlation between the visual content within the current memory bank and the sub-goal, and when this correlation exceeds a pre-set threshold, the frames within the memory bank are saved as the visual memories corresponding to that sub-goal. Finally, we store the visual memories with the sub goal's textual description into the Abstracted Multimodal Experience Pool. In addition, we incorporate the environment information, agent initial state, plan from Knowledge-Guided Planner, etc. into the experience memory of the given task. In this way, we consider the history information of each sub-goal and summaries and summarise it to finally compose the multimodal experience of the given task. 

Note that we also store these visual memories as failure cases when the feedback from the reflection phase is REPLAN. Therefore, when Optimus-1 executes a long-horizon task, it can retrieve past successes and failures as references and update memory after the task is finished. In the reflection phase, Optimus-1 retrieve the most relevant cases from Abstracted Multimodal Experience Pool, which contains the three scenarios COMPLETE, CONTINUE, and REPLAN, to help the agent better assess which state the current situation belongs to. This approach of considering both successful and failed cases for in-context learning is inspired by related research \cite{goddu2024development,parkes2019scientific}, and its effectiveness is validated in Section \ref{sec:ab}. 

\subsubsection{Hierarchical Directed Knowledge Graph}
 As shown in the Figure \ref{fig3}, crafting a diamond sword \includegraphics[width=0.3cm]{picture/logo/diamond_sword.pdf} requires two diamonds \includegraphics[width=0.3cm]{picture/logo/diamond.pdf} and a wooden stick \includegraphics[width=0.3cm]{picture/logo/stick.pdf}, while mining diamonds requires an iron pickaxe \includegraphics[width=0.3cm]{picture/logo/iron_pickaxe.pdf}, which in turn requires additional raw materials and crafting steps. We transform this mine and craft knowledge into a graph structure, where the nodes of the graph are objects, and the nodes point to objects that can be crafted or completed by that object. With directed graph, we show that connections between objects are established, and that this knowledge can be stored and updated efficiently. For a given object, we only need to retrieve the corresponding node to extract the corresponding subgraph from the knowledge graph. Then by topological sorting, we can get the antecedents and required materials for the object, and this information is provided to the Knowledge-Guided Planner as a way to generate a more reasonable sequence of sub-goals. With Hierarchical Directed Knowledge Graph, we can significantly enhance the world knowledge of the agent in a train-free manner, as shown in the experimental results in Section \ref{sec:ab}.

Our HDKG can be efficiently updated and expanded. When adding new nodes, the HDKG can be updated by simply merging the nodes and relationships into the graph. This method involves local linear modifications to the graph rather than altering the entire graph, making the process efficient and time-saving. For example, when M new nodes and N edges are added, the HDKG can be updated with M+N times of operations. Moreover, an HDKG containing 851 objects (nodes) requires less than 1 MB of memory. Thus, the HDKG can be efficiently updated and maintained.

\subsection{Hybrid Multimodal Memory Driven Optimus-1}
In order to implement the proposed Hybrid Multimodal Memory and to progressively increase the capacity of Optimus-1 in a self-evolution manner, we propose a non-parametric learning method named ``free exploration-teacher guidance''. 

In the free exploration phase, we randomly initialize the environment, materials, and tasks. For the task ``craft a wooden pickaxe'', we provide initial materials (three planks, two sticks), and then Optimus-1 (only the action controller activated) attempts to complete the task. If the environment feedback indicates the task is successful, the knowledge \{3 planks, 2 sticks → wooden pickaxe\} is added to the HDKG. Note that we randomly initialize materials and their quantities, which means that the task may not always succeed. As a result, each free exploration may not acquire the corresponding knowledge, but it can record the relevant experience (whether successful or fail). In the free exploration phase, Optimus-1 learns simple atomic operations, such as crafting sticks in the Wooden Group and mining diamonds in the Diamond Group. 

In the teacher guidance phase, Optimus-1 need to learn a small number of long-horizon tasks based on extra knowledge. For example, during the free exploration phase, Optimus-1 mastered crafting stick \includegraphics[width=0.3cm]{picture/logo/stick.pdf} and mining diamond \includegraphics[width=0.3cm]{picture/logo/diamond.pdf}, but did not know that ``a diamond sword \includegraphics[width=0.3cm]{picture/logo/diamond_sword.pdf} is obtained by a stick \includegraphics[width=0.3cm]{picture/logo/stick.pdf} and two diamonds \includegraphics[width=0.3cm]{picture/logo/diamond.pdf}''. So we provide some task plans, which will serve as extra knowledge to guide Optimus-1 to complete the task of ``craft diamond sword''. We built the following automated process to get the task plan needed for “free exploration”: 

\begin{itemize}
    \item We randomly select 5 tasks for each Group (7 groups in total) that are not included in the benchmark.
    \item For each selected task, we use a script to automatically obtain the crafting relationships from the Minecraft Wiki \footnote{https://minecraft.wiki/}. Taking the task “craft a wooden sword” as an example, we use the script to automatically obtain the crafting relationships: {1 wooden stick, 2 planks, 1 crafting table → 1 wooden sword}, {1 log → 4 planks}, {2 planks → 4 sticks}, {4 planks → 1 crafting table}.
    \item These relationships are converted into a directed acyclic graph through an automated script. By performing a topological sort, the graph can be converted into tuples of materials and their quantities: (wooden sword, 1), (crafting table, 1), (wooden stick, 1) (planks, 8), (log, 2).
    \item We prompt GPT-4 to construct a plan in order from basic materials to advanced materials.
    \item Finally, we get the plan: {1. Get two logs 2. Craft eight planks 3. Craft a crafting table 4. Craft a wooden stick 5. Craft a wooden sword}
\end{itemize}

During the teacher guidance phase, Optimus-1's memory is further expanded and it gains the experience of executing complete long-horizon tasks. Teacher guidance phase allows Optimus-1 to acquire advanced knowledge and learn multimodal experiences through complete long-horizon tasks.

\subsection{Backbone of Optimus-1}
Optimus-1 consists of Knowledge-Guided Planner, Experience-Driven Reflector, and Action Controller. In this paper, we employ OpenAI’s GPT-4V (\texttt{gpt-4-turbo}) \footnote{https://openai.com/index/gpt-4v-system-card/} as Knowledge-Guided Planner and Experience-Driven Reflector, and STEVE-1 \cite{lifshitz2024steve} as Action Controller. We also employ open-source models like Deepseek-VL \cite{lu2024deepseek} and InternLM-XComposer2-VL \cite{dong2024internlm} as Knowledge-Guided Planner and Experience-Driven Reflector.

All experiments were implemented on 4x NVIDIA A100 GPUs. We employ multiple Optimus-1 to perform different tasks at the same time, and this parallelized inference greatly improves our experimental efficiency. In the free exploration and teacher guidance phases, there is no need to access OpenAI's API, and the learning process takes approximately 16 hours on 4x A100 80G GPUs. During the inference phase, it takes about 20 hours on 4x A100 80G GPUs.

Throughout the experiment, we spent about $\$5,000$ to access the GPT-4V API. However, we also offer more cost-effective solutions. As shown in Figure \ref{fig4}, if we employ Deepseek-VL \cite{lu2024deepseek} or InternLM-XComposer2-VL \cite{dong2024internlm} as Optimus-1's backbone, we can get comparable performance with low-cost!

\subsection{Prompt for Optimus-1}
We show the prompt templates for Experience-Driven Reflector and Action Controller as follows.

\input{picture/appendix/prompt_plan}
\input{picture/appendix/prompt_reflect}
\section{Additional Experimental Results}
\label{experiment}
\subsection{Full Results on Our Benchmark}
We list the results of each task on the benchmark below, with details including task name, sub-goal numbers, success rate (SR), average number of steps (AS), average time (AT), and eval times. All tasks are evaluated in Minecraft 1.16.5 Survival Mode. Note that each time Optimus-1 performs a task, we initial it with an empty initial inventory and a random start point. This makes it challenging for Optimus-1 to perform each task. 

Moreover, in MineRL \cite{guss2019minerl} environment, 'steps' refers to the number of interactions between the agent and the environment, occurring at a frequency of 20 times per second. For example, if an agent takes 2 seconds to complete the task “chop a tree”, it interacts with the environment 40 times, resulting in a recorded steps number of 40. Experimental results show that Optimus-1's average task completion step (AS) is significantly lower than other baselines.

\input{table/our_benchmark_result/wooden_result}
\input{table/our_benchmark_result/stone_result}
\input{table/our_benchmark_result/iron_result}
\input{table/our_benchmark_result/gold_result}
\input{table/our_benchmark_result/diamond_result}
\input{table/our_benchmark_result/redstone_result}
\input{table/our_benchmark_result/armor_result}

\subsection{Results on Other Benchmark}
\label{other_bench}
For a more comprehensive comparison with current Minecraft Agents, we also report Optimus-1's performances on the benchmark used by Voyager \cite{wang2023voyager}, MP5 \cite{qin2023mp5}, and DEPS \cite{wang2023describe} below. Due to the different environments and settings, agents perform tasks with varying degrees of difficulty. For example, Optimus-1 requires low-level action to perform any task in MineRL \cite{guss2019minerl}, and we initialize its inventory to be empty. While Voyager \cite{wang2023voyager} performs tasks in Mineflayer \footnote{https://github.com/PrismarineJS/mineflayer} environment only through encapsulated code, MP5 \cite{qin2023mp5} performs tasks in MineDOJO \cite{fan2022minedojo} environment only needs a specific control signal to craft tools, no low-level actions (mouse movement and click) are needed.

Optimus-1's success rate in completing tasks with these baselines is shown in the Table \ref{tab:mp5_result} and Table \ref{tab:deps_result}, and Optimus-1's efficiency in unlocking the tech tree in Minecraft is shown in the Figure \ref{tech_tree}. These results reveal that Optimus-1 outperforms a variety of powerful baseline agents, even in challenging environmental settings!
\input{table/other_benchmark/mp5_result}
\input{table/other_benchmark/deps_result}
\input{table/other_benchmark/tech-tree}

\clearpage

\section{Case Study}
This section introduces several cases to comprehensively demonstrate Optimus-1's capabilities. 

Figures \ref{fig:case-replan}, \ref{fig:case-done}, and \ref{fig:case-continue}  demonstrate the superiority of our reflection mechanism, which dynamically adjusts the plan based on the current game progress.
\begin{itemize}
    \item Figure \ref{fig:case-replan} illustrates Optimus-1's replanning ability. When Optimus-1 realizes it cannot complete a task (such as a craft failure shown in the figure), it will replan the current task and continue execution. 
    \item Figures \ref{fig:case-done} and \ref{fig:case-continue} showcase Optimus-1's ability to make judgments based on visual signals. When Optimus-1 determines that it has completed a task (such as ``kill a cow \includegraphics[width=0.3cm]{picture/logo/cow.pdf}'' in Figure \ref{fig:case-done}), it will finish the current task and move on to the next one. If Optimus-1 discovers that it has not yet completed the task and the task has not failed(as shown in Figure \ref{fig:case-continue}), it will continue executing the task.
\end{itemize}

Figures \ref{fig:case-plan-knowledge} and \ref{fig:case-plan-one} illustrate the advantages of planning with knowledge. With the Hierarchical Directed Knowledge Graph, we can generate a high-quality plan in one step and dynamically adjust the plan based on current visual signals.

\begin{itemize}
    \item Figure \ref{fig:case-plan-knowledge} demonstrates the importance of knowledge. For a long-horizon task such as ``Mine 1 diamond \includegraphics[width=0.3cm]{picture/logo/diamond.pdf},'' Optimus-1 first generates a plan based on the Hierarchical Directed Knowledge Graph. However, this plan needs to be adjusted based on the current visual signals. For example, in this figure, Optimus-1 appears in a cave, so the primary task is not to ``chop a tree'' but to ``leave the cave'' first. Only after exiting the cave can Optimus-1 proceed with the initial plan.
    \item Figure \ref{fig:case-plan-one} demonstrates the high efficiency of our method. Agents like MP5 \cite{qin2023mp5} and Voyager \cite{wang2023voyager} use an iterative planning approach, which is very time-consuming, generating the final plan step by step. During this process, agent does not take any action. As shown in Figure \ref{fig:case-plan-one}, a zombie is gradually approaching the agent, but the agent is still iterating on its plan. Optimus-1, however, generates the plan in one step based on the Hierarchical Directed Knowledge Graph and makes reasonable plans based on the current visual signals.
\end{itemize}

\input{picture/case-study}

\clearpage

\end{document}

%% file: picture/fig-1.tex
\begin{figure*}[!h]
    \centering
    \includegraphics[width=1.0\textwidth]{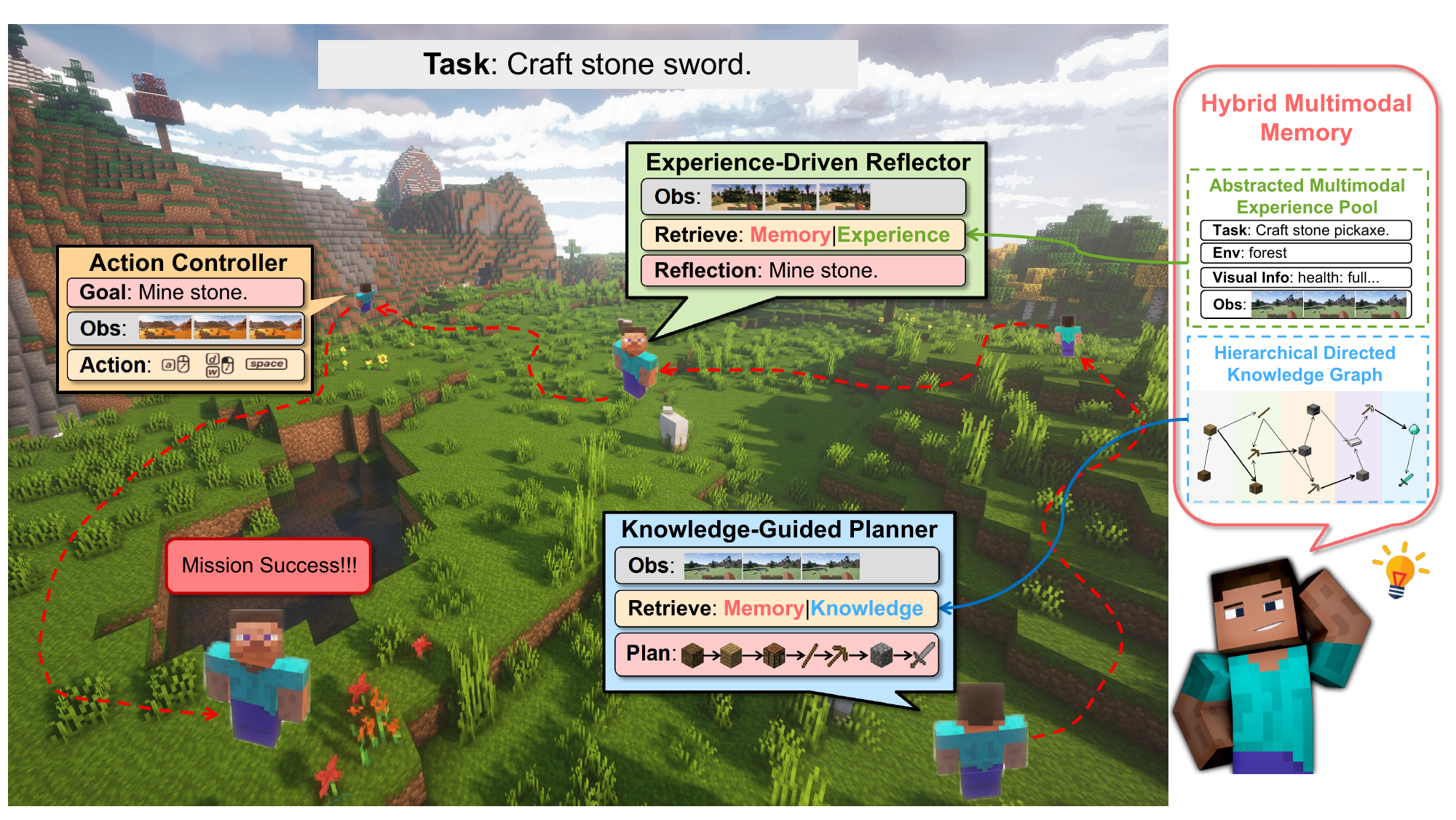}
    \caption{An illustration of Optimus-1 performing long-horizon tasks in Minecraft. Given the task ``Craft stone sword'', Knowledge-Guided Planner incorporates knowledge from Hierarchical Directed Knowledge Graph into planning, then Action Controller executes these planning sequences step-by-step. During the execution of the task, the Experience-Driven Reflector is periodically activated and retrieve experience from Abstracted Multimodal Experience Pool to make reflection.}
    \label{fig1}
\end{figure*}

%% file: picture/fig-3.tex
\begin{figure}[ht]
  \centering
  \includegraphics[width=1\linewidth]{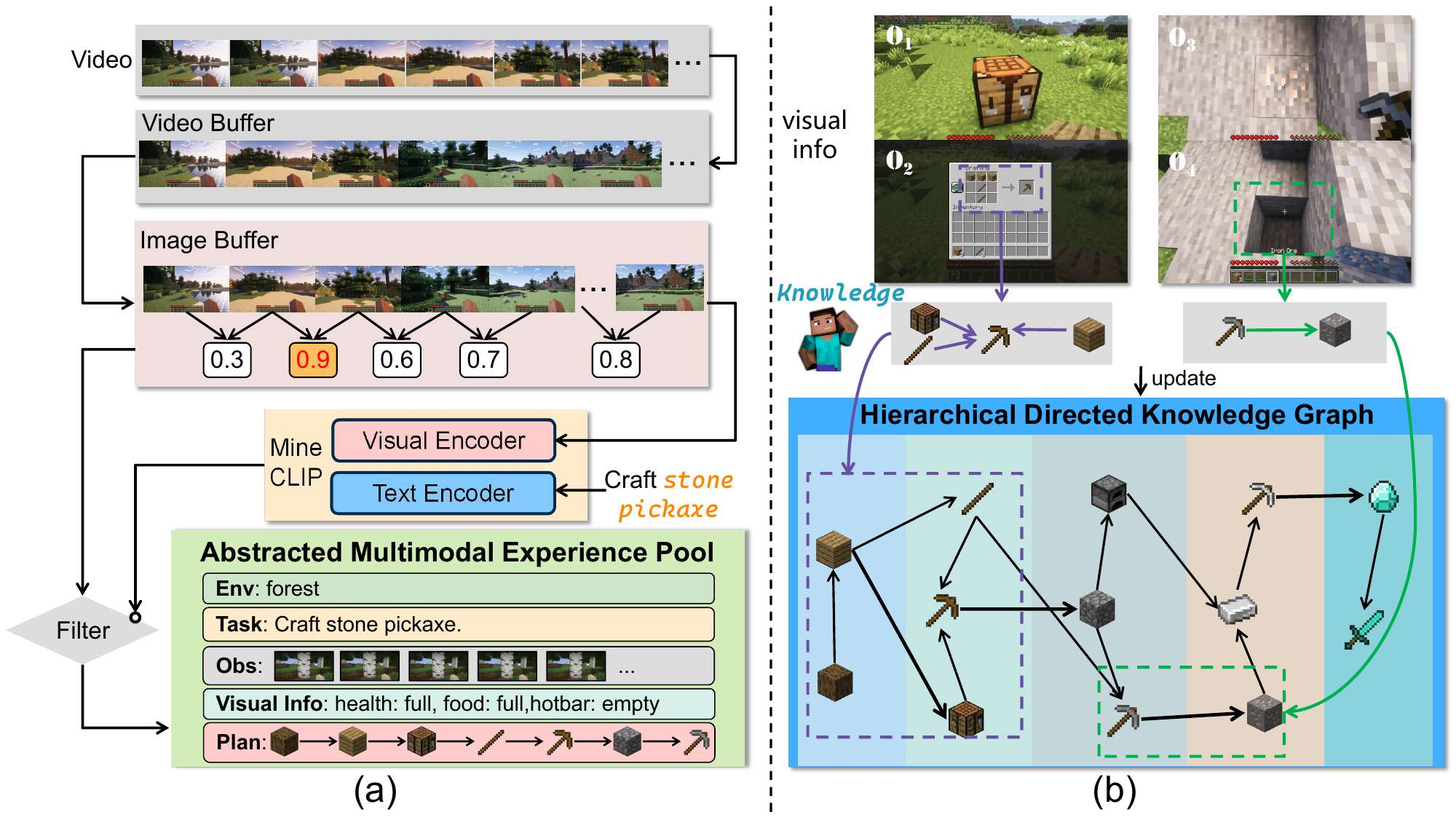}
  \caption{\textbf{(a)} Extraction process of multimodal experience. The frames are filtered through video buffer and image buffer, then MineCLIP \cite{fan2022minedojo} is employed to compute the visual and sub-goal similarities and finally they are stored in Abstracted Multimodal Experience Pool. \textbf{(b)} Overview of Hierarchical Directed Knowledge Graph. Knowledge is stored as a directed graph, where its nodes represent objects, and directed edges point to materials that can be crafted by this object.}
  \label{fig3}
\end{figure}

%% file: picture/fig-2.tex
\begin{figure}[t]
  \centering
  \includegraphics[width=1\linewidth]{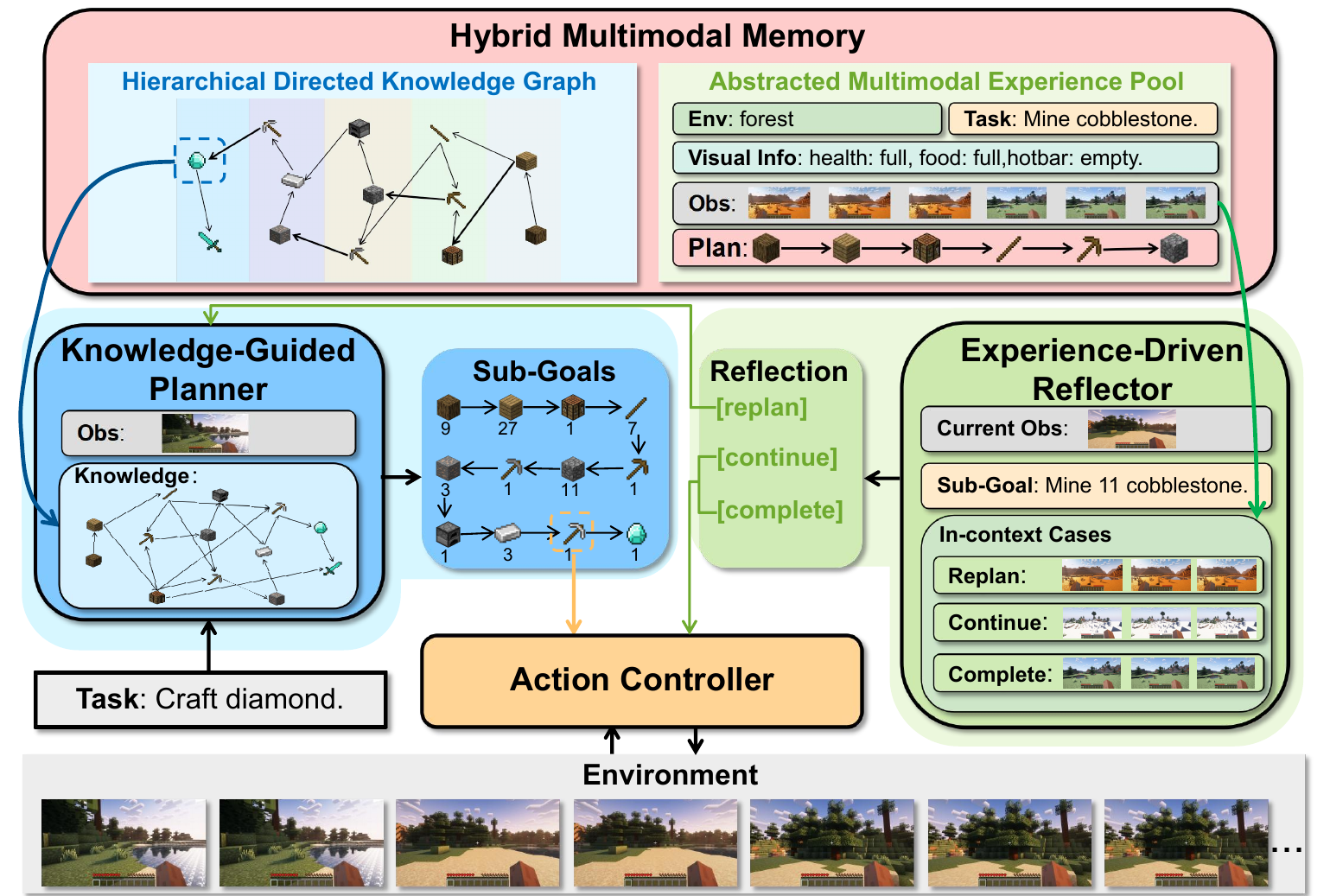}
  \caption{Overview framework of our Optimus-1. Optimus-1 consists of Knowledge-Guided Planner, Experience-Driven Reflector, Action Controller, and Hybrid Multimodal Memory architecture. Given the task ``craft stone sword'', Optimus-1 incorporates the knowledge from HDKG into Knowledge-Guided Planning, then Action Controller generates low-level actions. Experience-Driven Reflector is periodically activated to introduce multimodal experience from AMEP to determine if the current task can be executed successfully. If not, it will ask the Knowledge-Guided Planner to refine the plan.}
  \label{fig2}
\end{figure}

%% file: table/tb_main.tex
\begin{table}[]
\centering
\caption{Main Result of Optimus-1 on long-horizon tasks benchmark. We report the average success rate (SR), average number of steps (AS), and average time (AT) on each task group, the results of each task can be found in the Appendix \ref{experiment}. Lower AS and AT metrics mean that the agent is more efficient at completing the task, while $+\infty$ indicates that the agent is unable to complete the task. Overall represents the average result on the five groups of Iron, Gold, Diamond, Redstone, 
 and Armor.}
\label{tb_main}
\resizebox{\textwidth}{!}{%
\renewcommand\arraystretch{1.3}
\begin{tabular}{llllll>{\columncolor[HTML]{fdf9ea}}l>{\columncolor[HTML]{EDDEDE}}l}

\toprule[1.3pt]
Group                     & Metric     & GPT-3.5 & GPT-4V  & DEPS     & Jarvis-1          & Optimus-1              & Human-level \\ \hline
\multirow{3}{*}{\includegraphics[width=0.4cm]{picture/logo/wood.pdf} Wood}     & SR $\uparrow$    & 40.16   & 41.42   & 77.01    & 93.76             & \textbf{98.60}    & 100.00      \\
                          & AT $\downarrow$  & 56.39   & 55.15   & 85.53    & 67.76             & \textbf{47.09}    & 31.08       \\
                           & AS $\downarrow$ & 1127.78 & 1103.04 & 1710.61  & 1355.25           & \textbf{841.94}   & 621.59      \\
                          \hline
\multirow{3}{*}{\includegraphics[width=0.4cm]{picture/logo/cobblestone.pdf} Stone}    & SR $\uparrow$    & 20.40   & 20.89   & 48.52    & 89.20             & \textbf{92.35}    & 100.00      \\
                
                          & AT $\downarrow$  & 135.71  & 132.77  & 138.71    & 141.50            & \textbf{129.94}   & 80.85       \\
                          & AS $\downarrow$ & 2714.21 & 2655.47 & 2574.30  & 2830.05           & \textbf{2518.88}  & 1617.00     \\
                          \hline
\multirow{3}{*}{\includegraphics[width=0.4cm]{picture/logo/iron_ingot.pdf} Iron}     & SR $\uparrow$    & 0.00    & 0.00    & 16.37    & 36.15             & \textbf{46.69}    & 86.00      \\
                          & AT $\downarrow$  & $+\infty$   & $+\infty$   & 944.61   & 722.78            & \textbf{651.33}   & 434.38      \\ 
                          & AS $\downarrow$ & $+\infty$   & $+\infty$   & 8892.24  & 8455.51           & \textbf{6017.85}  & 5687.60     \\
                          \hline
\multirow{3}{*}{\includegraphics[width=0.4cm]{picture/logo/gold_ingot.pdf} Gold}     & SR $\uparrow$    & 0.00    & 0.00    & 0.00     & 7.20              & \textbf{8.51}     &     17.31        \\
                          & AT $\downarrow$  & $+\infty$   & $+\infty$   & $+\infty$    & 787.37            & \textbf{726.35}   & 557.08      \\ 
                          & AS $\downarrow$ & $+\infty$   & $+\infty$   & $+\infty$    & 15747.13          & \textbf{15527.07} & 13141.60    \\
                          \hline
\multirow{3}{*}{\includegraphics[width=0.4cm]{picture/logo/diamond.pdf} 
 Diamond}  & SR $\uparrow$    & 0.00    & 0.00    & 0.60     & 8.98              & \textbf{11.61}    &       16.98      \\
                          & AT $\downarrow$  & $+\infty$   & $+\infty$   & 1296.96   & 1255.06           & \textbf{1150.98}  & 744.82      \\ 
                          & AS $\downarrow$ & $+\infty$   & $+\infty$   & 23939.30 & 25101.25          & \textbf{23019.64} & 16237.54    \\
                          \hline
\multirow{3}{*}{\includegraphics[width=0.4cm]{picture/logo/readstone.pdf} Redstone} & SR $\uparrow$    & 0.00    & 0.00    & 0.00     & 16.31             & \textbf{25.02}    &      33.27       \\
                          & AT $\downarrow$  & $+\infty$   & $+\infty$   & $+\infty$    & 1070.42            & \textbf{932.50}   & 617.89      \\ 
                          & AS $\downarrow$ & $+\infty$   & $+\infty$   & $+\infty$    & 17408.40          & \textbf{12709.99} & 12357.00    \\
                          \hline
\multirow{3}{*}{\includegraphics[width=0.4cm]{picture/logo/armor.pdf} Armor}    & SR $\uparrow$    & 0.00    & 0.00    & 9.98     & 15.82             & \textbf{19.47}    &       28.48      \\
                          & AT $\downarrow$  & $+\infty$   & $+\infty$   & 997.59   & 924.60   & \textbf{824.53}            & 551.30      \\ 
                           & AS $\downarrow$ & $+\infty$   & $+\infty$   & 17951.95  & 16492.96 & \textbf{16350.56}          & 11026.00    \\
                          \hline
    Overall & SR $\uparrow$    & 0.00    & 0.00    & 5.39     & 16.89             & \textbf{22.26}    &       36.41 \\ 
                          \bottomrule[1.3pt]
\end{tabular}
}
\end{table}

%% file: table/tb_ablation.tex
\begin{table*}[t]
\footnotesize
\begin{minipage}[t]{0.50\textwidth}
\centering
\caption{Ablation study results. We report average success rate (SR) on each task group. \texttt{P.}, \texttt{R.}, \texttt{K.}, \texttt{E.} represent Planning, Reflection, Knowledge, and Experience, respectively.}
\resizebox{\textwidth}{!}{%
\renewcommand\arraystretch{1.3}
\begin{tabular}{cccc|ccccc}
\toprule[1.2pt]
\multicolumn{4}{c|}{Ablation Setting}                                                                         & \multicolumn{5}{c}{Task Group}          \\ \hline
P.                        & R.                        & K.                        & E.                        & Wood  & Stone & Iron  & Gold  & Diamond \\ \hline
                          &                           &                           &                           & 14.29 & 0.00     & 0.00     & 0.00     & 0.00       \\
\Checkmark &                           &                           &                           & 42.95 & 25.67 & 0.00     & 0.00     & 0.00       \\
\Checkmark & \Checkmark &                           &                           & 55.00    & 47.37 & 18.11 & 2.08  & 1.11    \\
\Checkmark & \Checkmark &                           & \Checkmark & 73.53 & 64.20  & 24.19 & 3.08  & 1.86    \\
\Checkmark & \Checkmark & \Checkmark &                           & 92.37 & 69.63 & 38.33 & 3.49  & 2.42    \\
\rowcolor[HTML]{fdf9ea}
\Checkmark & \Checkmark & \Checkmark & \Checkmark & \textbf{97.49} & \textbf{94.26} & \textbf{53.33} & \textbf{11.54} & \textbf{9.59}    \\ 
\bottomrule[1.2pt]
\end{tabular}
}
\label{tb:ab_1}
\end{minipage}
\begin{minipage}[t]{0.5\textwidth}
\centering
\caption{Ablation study on AMEP. We report the average success rate (SR) on each task group. \texttt{Zero}, \texttt{Suc.}, and \texttt{Fail.} represent retrieving from AMEP without getting the case, getting the success case, and getting the failure case, respectively.}
\resizebox{\textwidth}{!}{%
\renewcommand\arraystretch{1.3}
\begin{tabular}{ccc|ccccc}
\toprule[1.2pt]
\multicolumn{3}{c|}{Ablation Setting}                                             & \multicolumn{5}{c}{Task Group}          \\ \hline
Zero                      & Suc.                      & Fai.                      & Wood  & Stone & Iron  & Gold  & Diamond \\ \hline
\Checkmark &                           &                           & 92.00 & 79.26 & 36.32 & 4.25  & 3.25    \\
                          & \Checkmark &                           & 95.00 & 84.29 & 46.98 & 9.36  & 7.89    \\
                          &                           & \Checkmark & 95.00 & 81.10 & 45.47 & 7.50  & 6.39    \\
                          \rowcolor[HTML]{fdf9ea}
                          & \Checkmark & \Checkmark & \textbf{97.49} & \textbf{94.26} & \textbf{53.33} & \textbf{11.54} & \textbf{9.59}    \\
\bottomrule[1.2pt]
\end{tabular}
}
\label{tb:ab_2}
\end{minipage}
\end{table*}

%% file: picture/reflectiontex.tex
\begin{figure}[ht]
  \centering
  \includegraphics[width=1\linewidth]{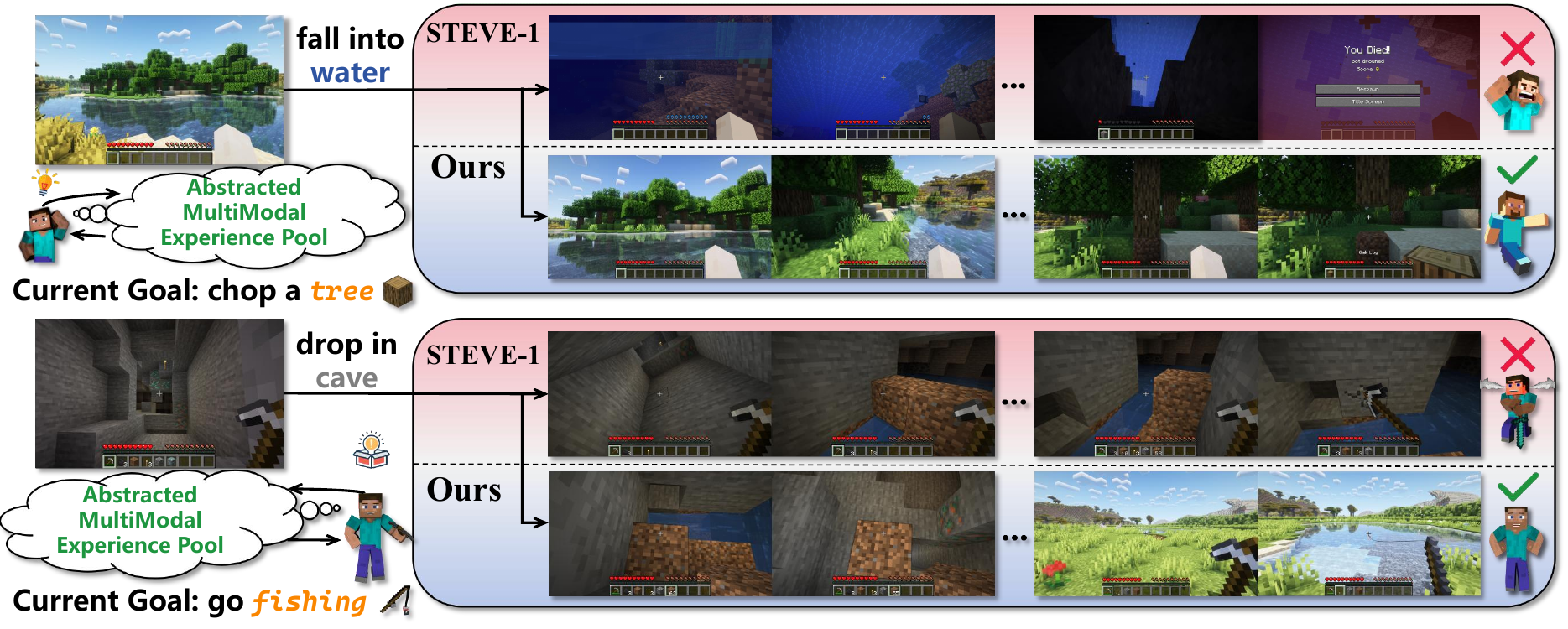}
  \caption{Illustration of the role of reflection mechanism. Without the help of reflective mechanisms, STEVE-1 \cite{lifshitz2024steve} often gets into trouble and fails to complete the task. While Optimus-1, with the help of the Experience-Driven Reflector, leverages the AMEP to retrieve relevant experience, reflect current situation and correct errors. This improves Optimus-1's success rate on long-horizon tasks.}
  \label{fig:fig-reflection}
\end{figure}

%% file: picture/fig5.tex
\begin{figure}[!htb]
  \centering
  \subfloat[Generalisation of Hybrid Multimodal Memory]
  {\includegraphics[width=0.5\textwidth]{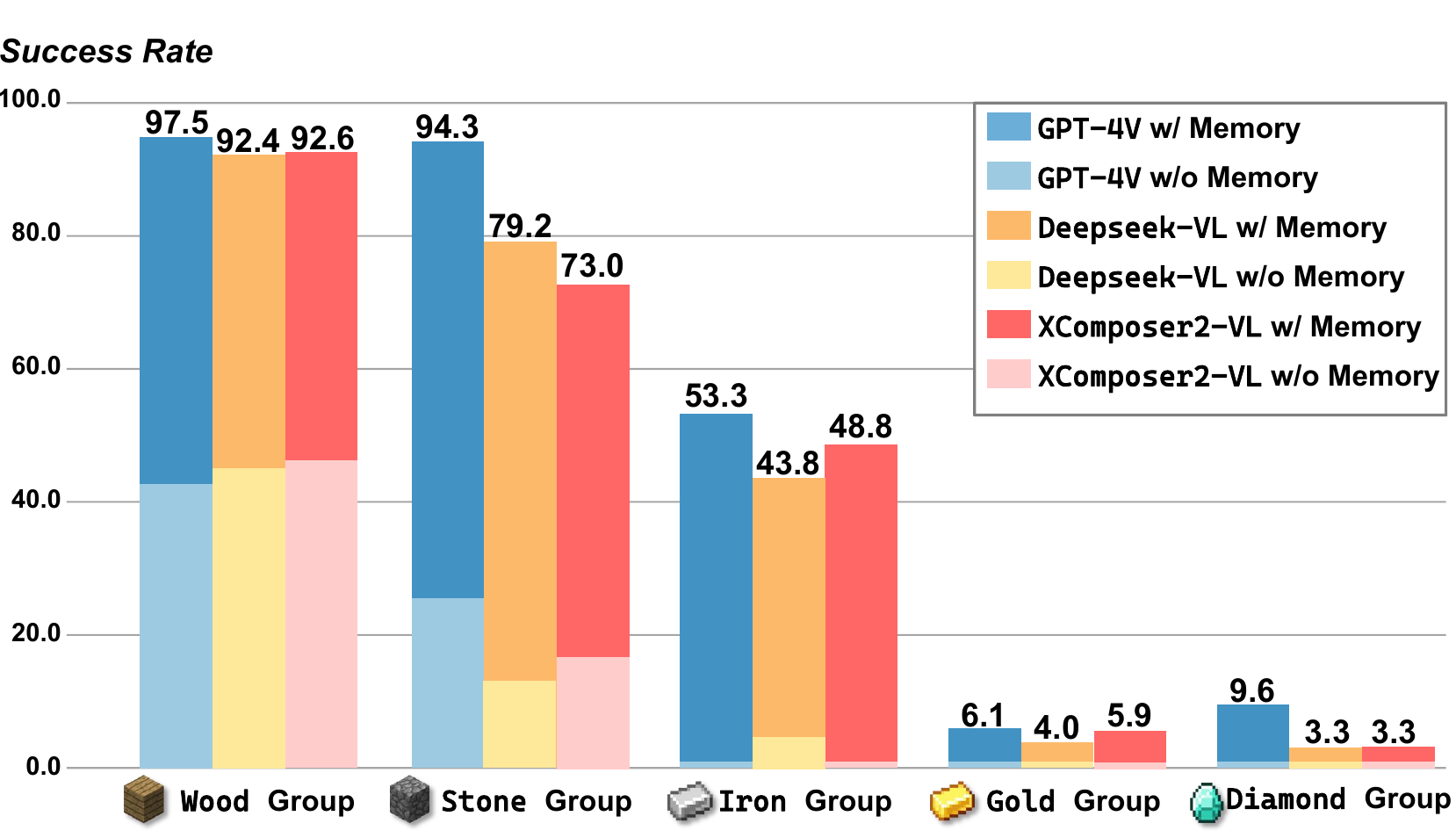}\label{fig41}}
  \subfloat[Self-Evolution]
  {\includegraphics[width=0.5\textwidth]{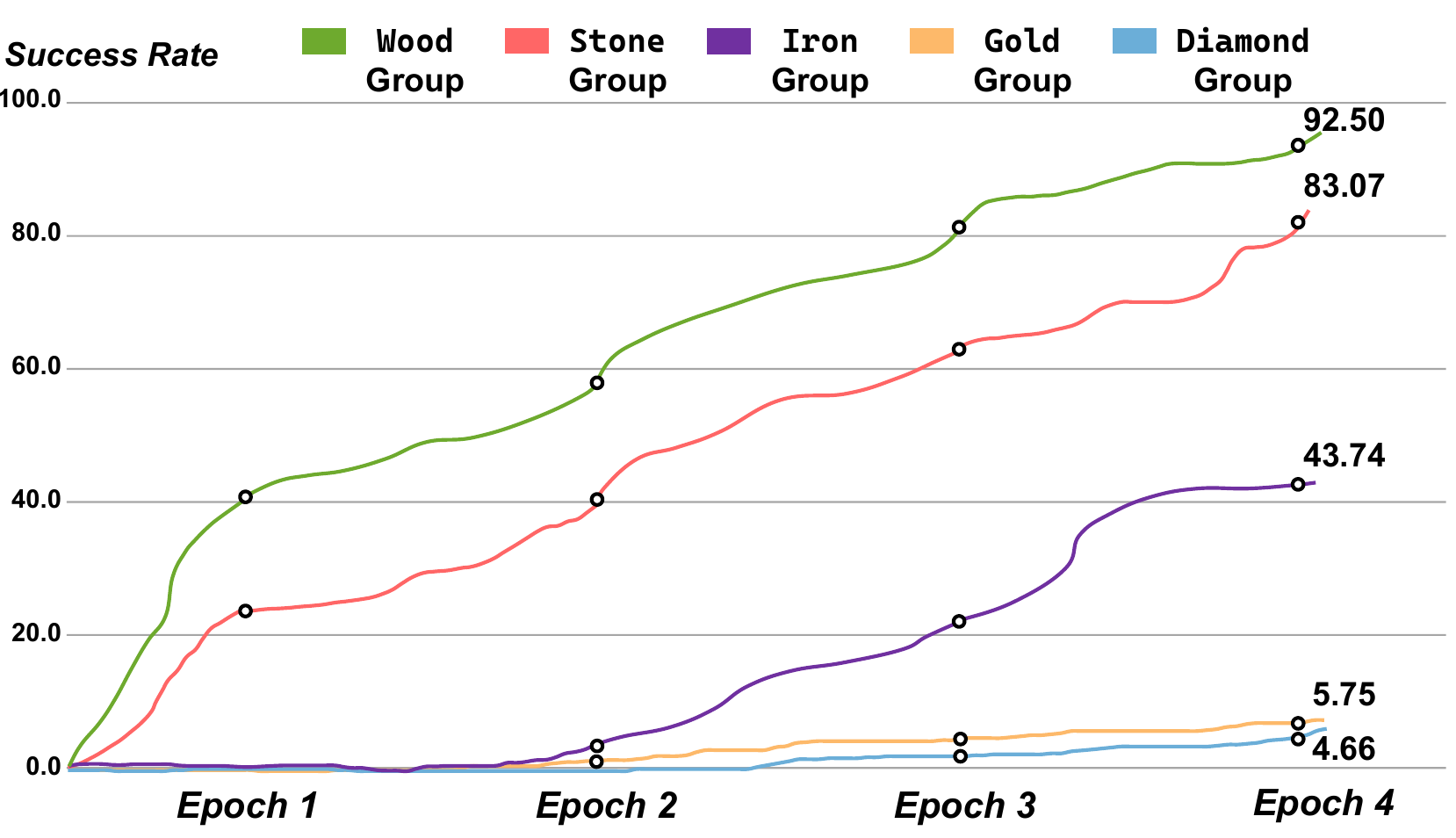}\label{fig42}}
  \caption{\textbf{(a)} With the help of Hybrid Multimodal Memory, various MLLM-based Optimus-1 have demonstrated 2 to 6 times performance improvement. \textbf{(b)} Illustration of the change in Optimus-1 success rate on the unseen task over 4 epochs.}
  \label{fig4}
\end{figure}

%% file: picture/mc-intro.tex
\begin{figure}[!ht]
    \subfloat[\label{screen_shot:fig1}]{%
      \includegraphics[width=0.5\textwidth]{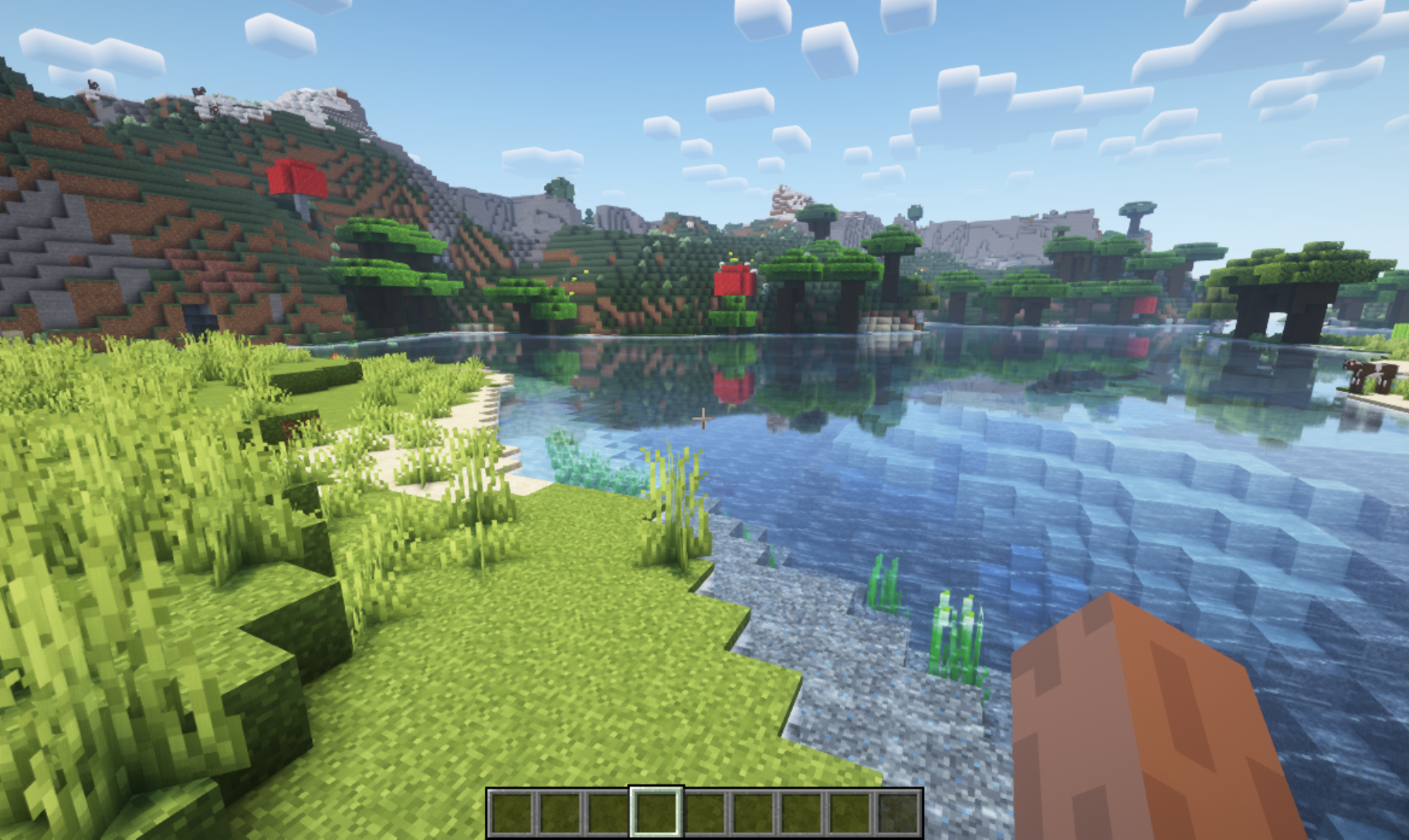}
    }
    \hfill
    \subfloat[\label{screen_shot:fig2}]{%
      \includegraphics[width=0.5\textwidth]{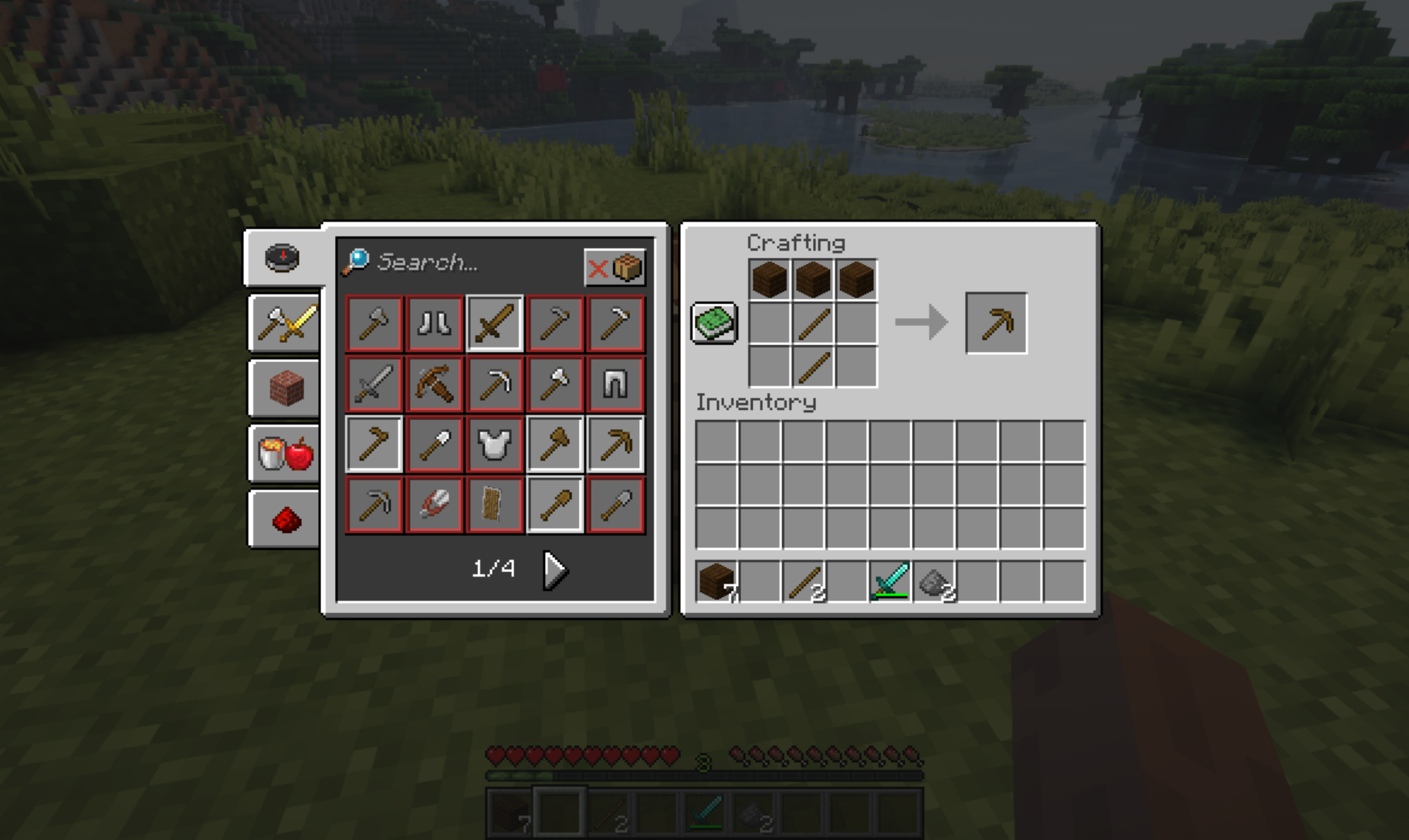}
    }
    \vfill
    \subfloat[\label{screen_shot:fig3}]{%
      \includegraphics[width=0.5\textwidth]{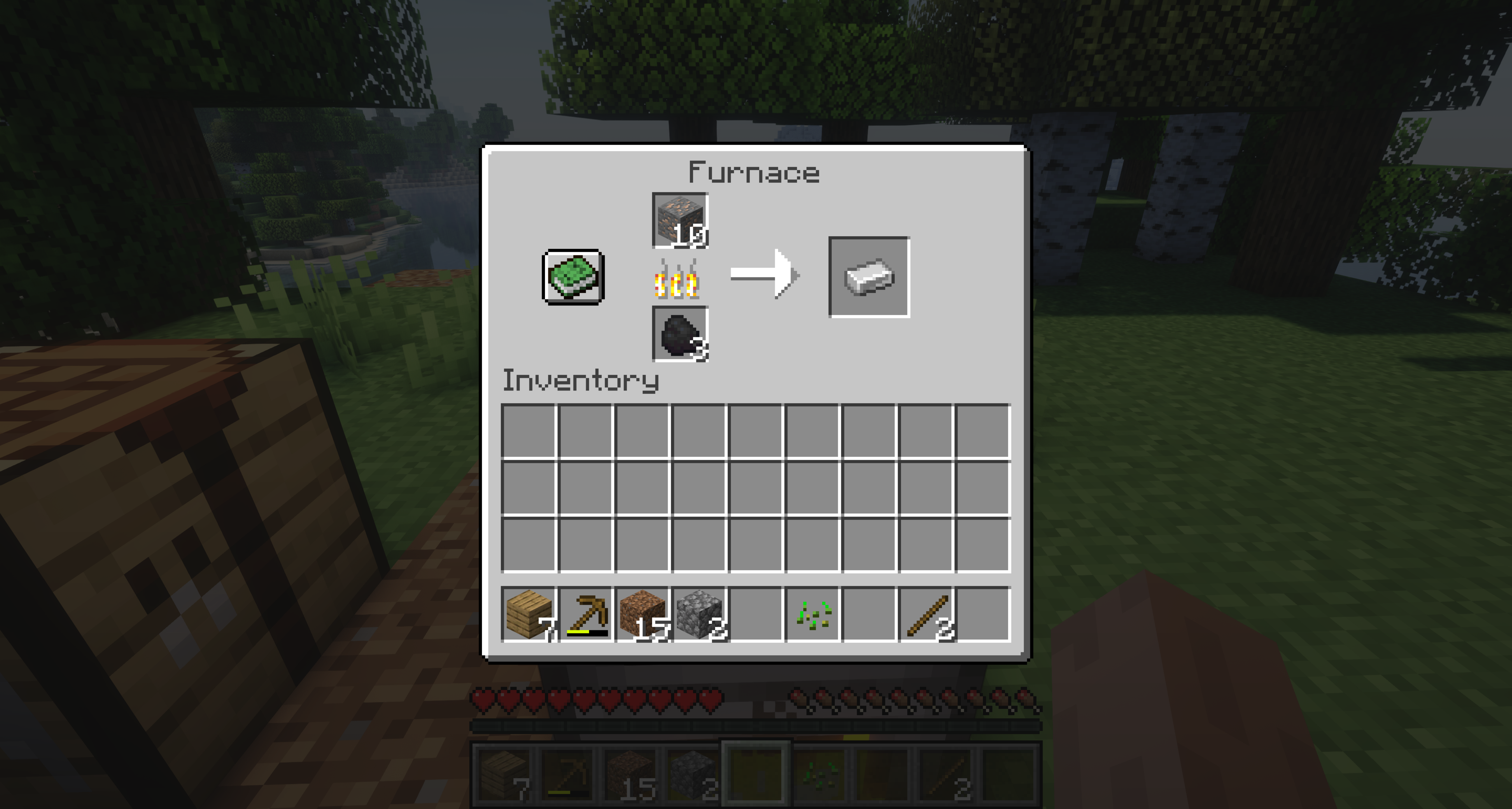}
    }
    \hfill
    \subfloat[\label{screen_shot:fig4}]{%
      \includegraphics[width=0.5\textwidth]{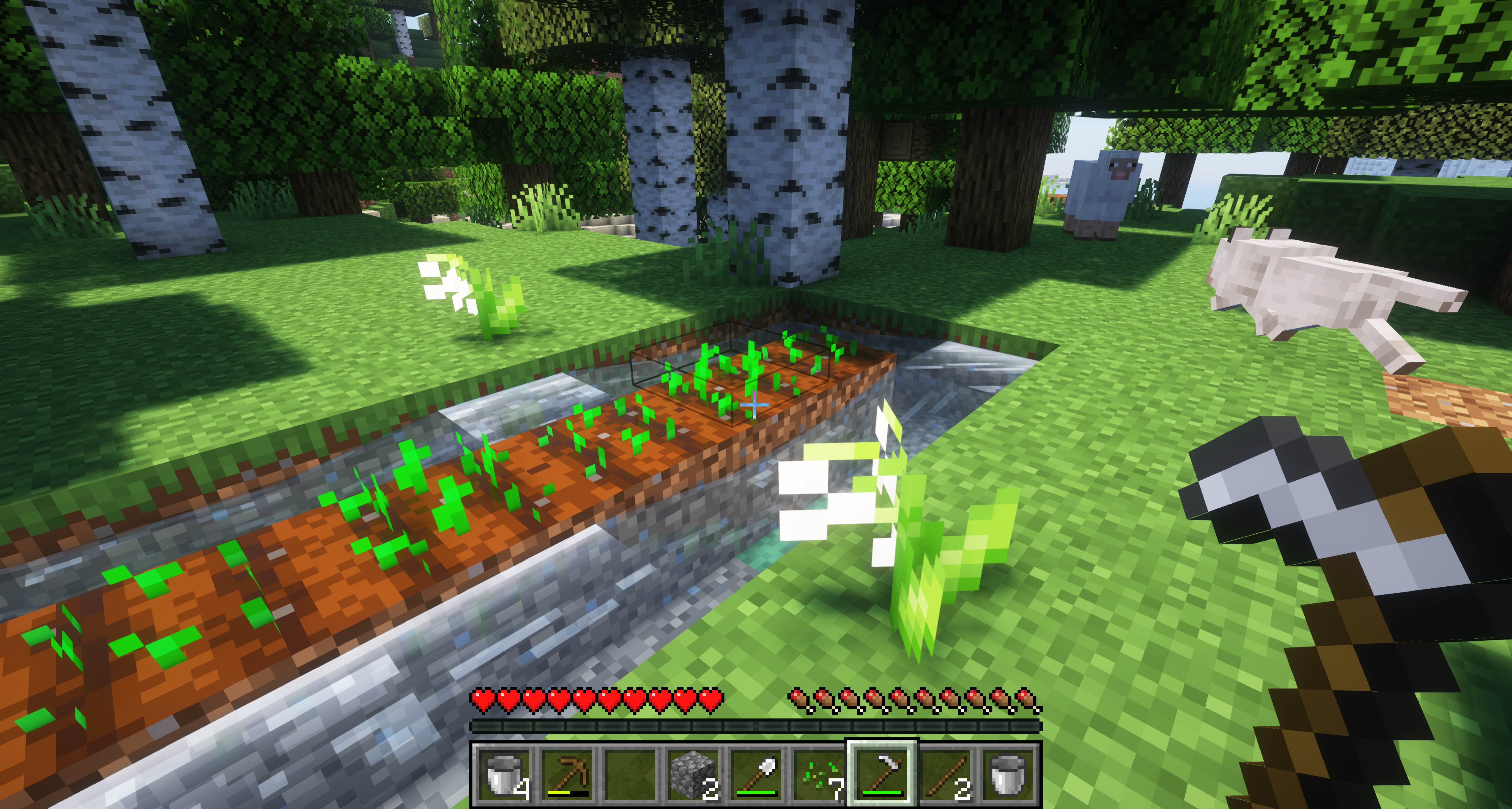}
    }
    \caption{The screenshots in Minecraft. \textbf{(a).} The world has different complex terrains, including plains, river, forest and mine. \textbf{(b).} The agent can use crafting table to craft tools and items with recipes. \textbf{(c).} The agent can use the furnace to smelt ore to obtain precious ingot. \textbf{(d).} The agent can grow wheat near the river.}
    \label{fig:mc-screen-shot}
\end{figure}

%% file: table/action_space.tex
\begin{table}[]
\centering
\caption{Our action space.}
\label{tab:action_space}
\resizebox{\textwidth}{!}{%
\renewcommand\arraystretch{1.2}
\begin{tabular}{cccccc}
\toprule[1.5pt]
\textbf{Index} & \textbf{Action}      & \textbf{Human Action} & \textbf{Description}                               \\ \hline
1              & Forward              & key W                 & Move forward.                                      \\
2              & Back                 & key S                 & Move back.                                         \\
3              & Left                 & key A                 & Strafe left.                                       \\
4              & Right                & key D                 & Strafe right.                                      \\
5              & Jump                 & key Space             & Jump. When swimming, keeps the player afloat.      \\
6              & Sneak                & key left Shift        & Slowly move in the current direction of movement.  \\
7              & Sprint               & key left Ctrl         & Move quickly in the direction of current movement. \\
8  & Attack & left Button  & Destroy blocks (hold down); Attack entity (click once).                     \\
9  & Use    & right Button & Place blocks, entity, open items or other interact actions defined by game. \\
10             & hotbar {[}1-9{]}     & keys 1-9              & Selects the appropriate hotbar item.               \\
11             & Open/Close Inventory & key E                 & Opens the Inventory. Close any open GUI.           \\
12 & Yaw    & move Mouse X & Turning; aiming; camera movement.Ranging from -180 to +180.                 \\
13 & Pitch  & move Mouse Y & Turning; aiming; camera movement.Ranging from -180 to +180.                 \\
14             & Craft                & -                     & Execute a crafting recipe to obtain new item       \\
15             & Smelt                & -                     & Execute a smelting recipe to obtain new item.      \\ 
\bottomrule[1.5pt]
\end{tabular}%
}
\end{table}

%% file: picture/example_craft_iron_sword.tex
\begin{figure}[!ht]
    \subfloat[Chop $7$ logs\label{example:fig1}]{%
      \includegraphics[width=0.3\textwidth]{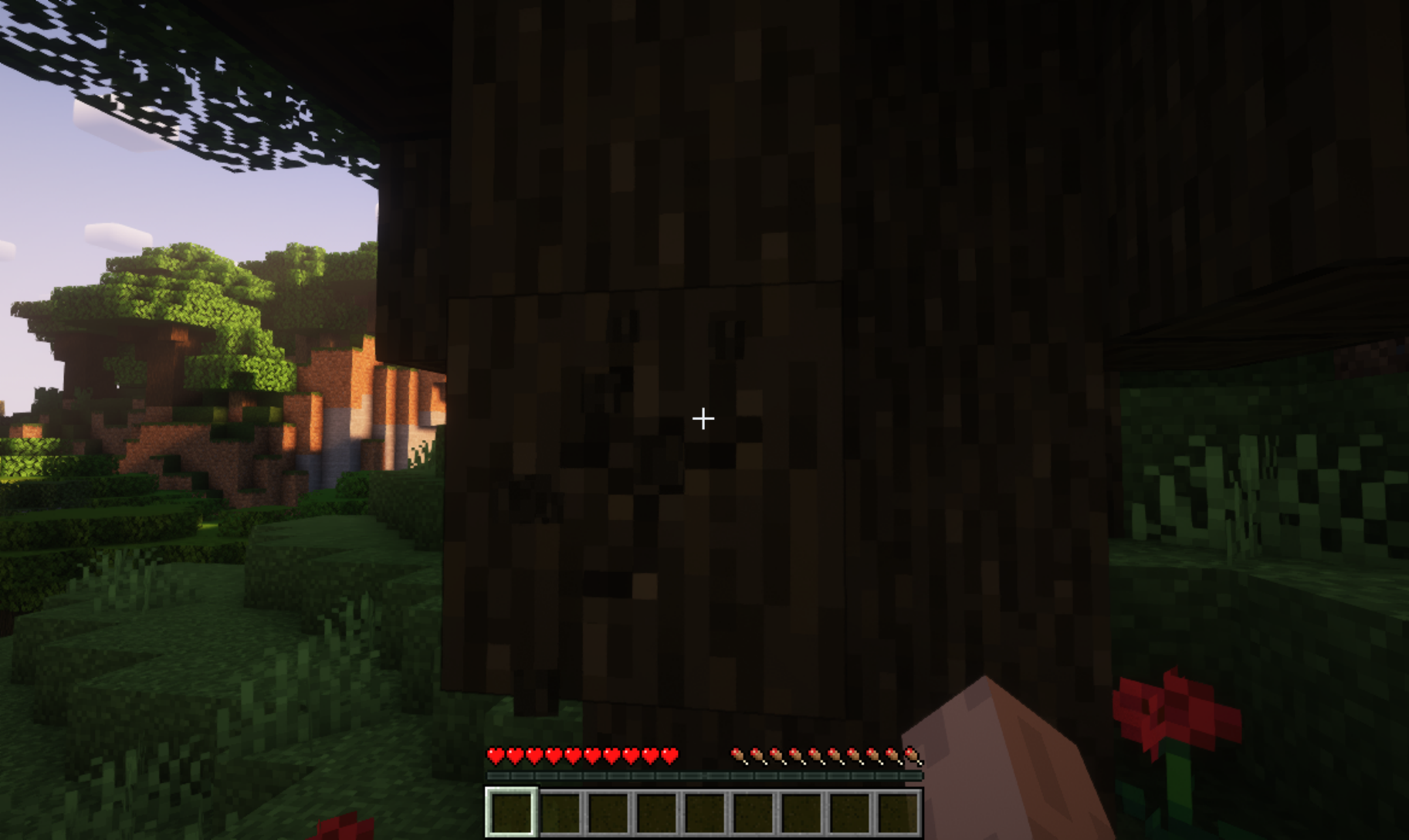}
    }
    \hfill
    \subfloat[Craft $21$ planks\label{example:fig2}]{%
      \includegraphics[width=0.3\textwidth]{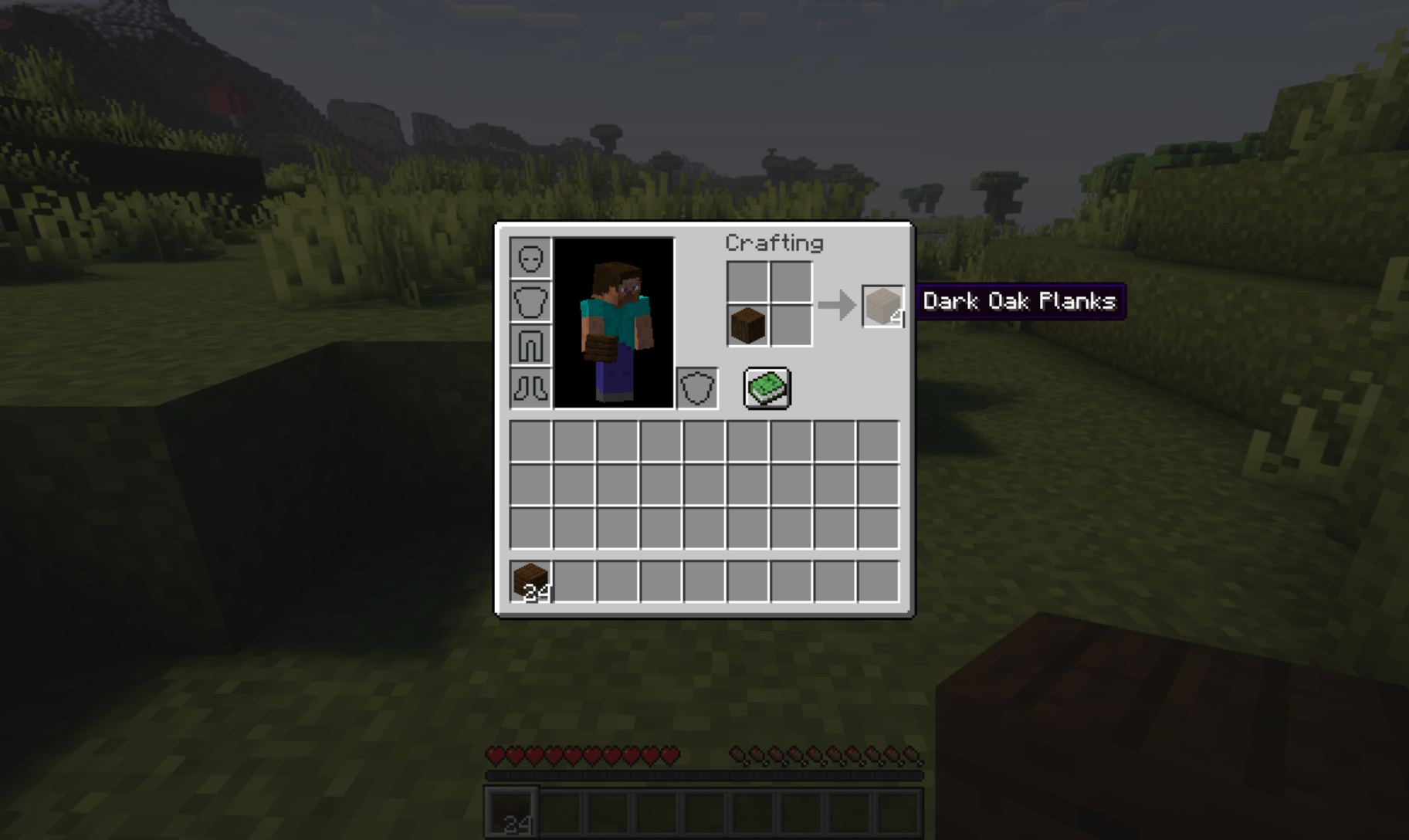}
    }
    \hfill
    \subfloat[Craft $5$ sticks\label{example:fig3}]{%
      \includegraphics[width=0.3\textwidth]{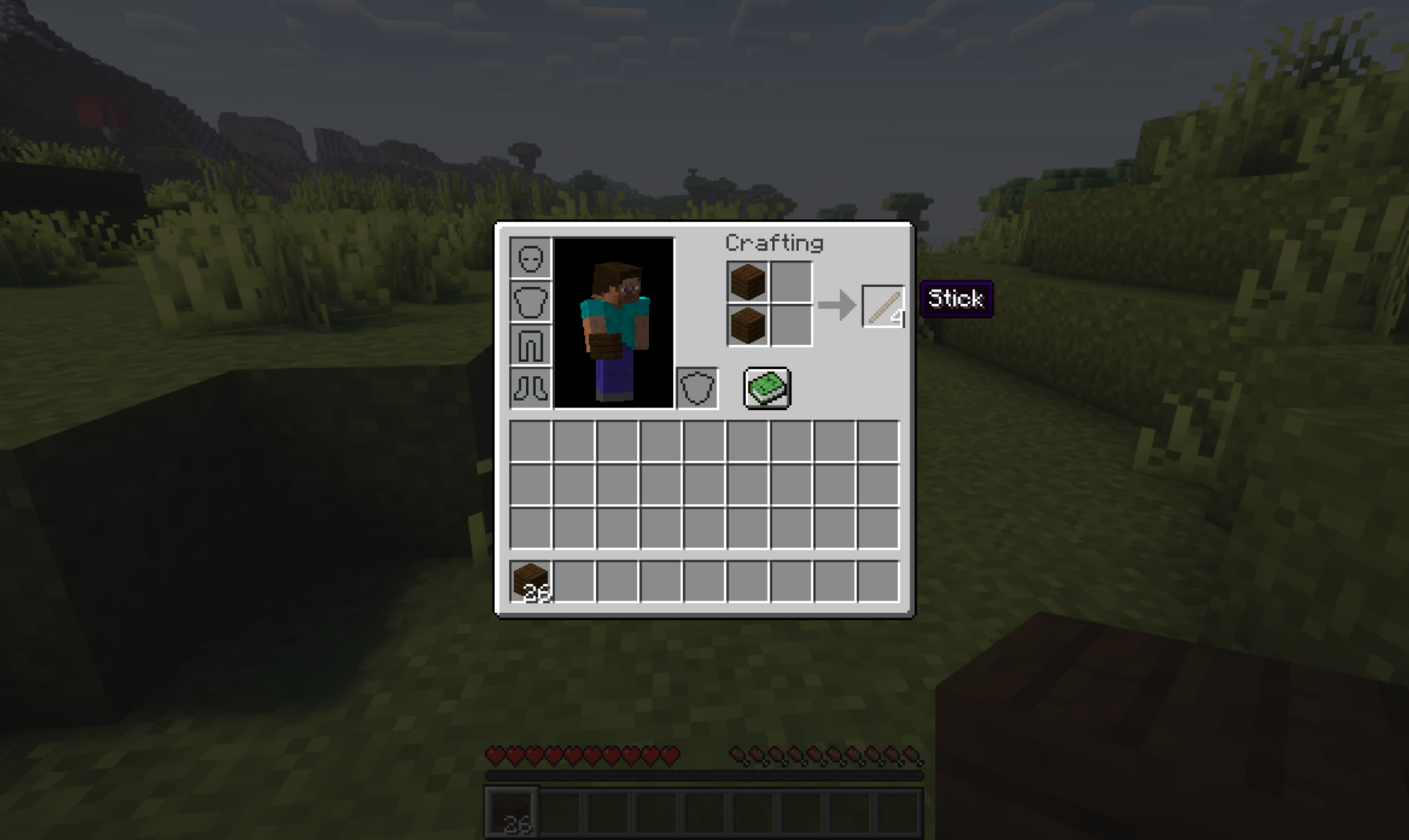}
    }
    \hfill
    \subfloat[Craft $1$ crafting table\label{example:fig11}]{%
      \includegraphics[width=0.3\textwidth]{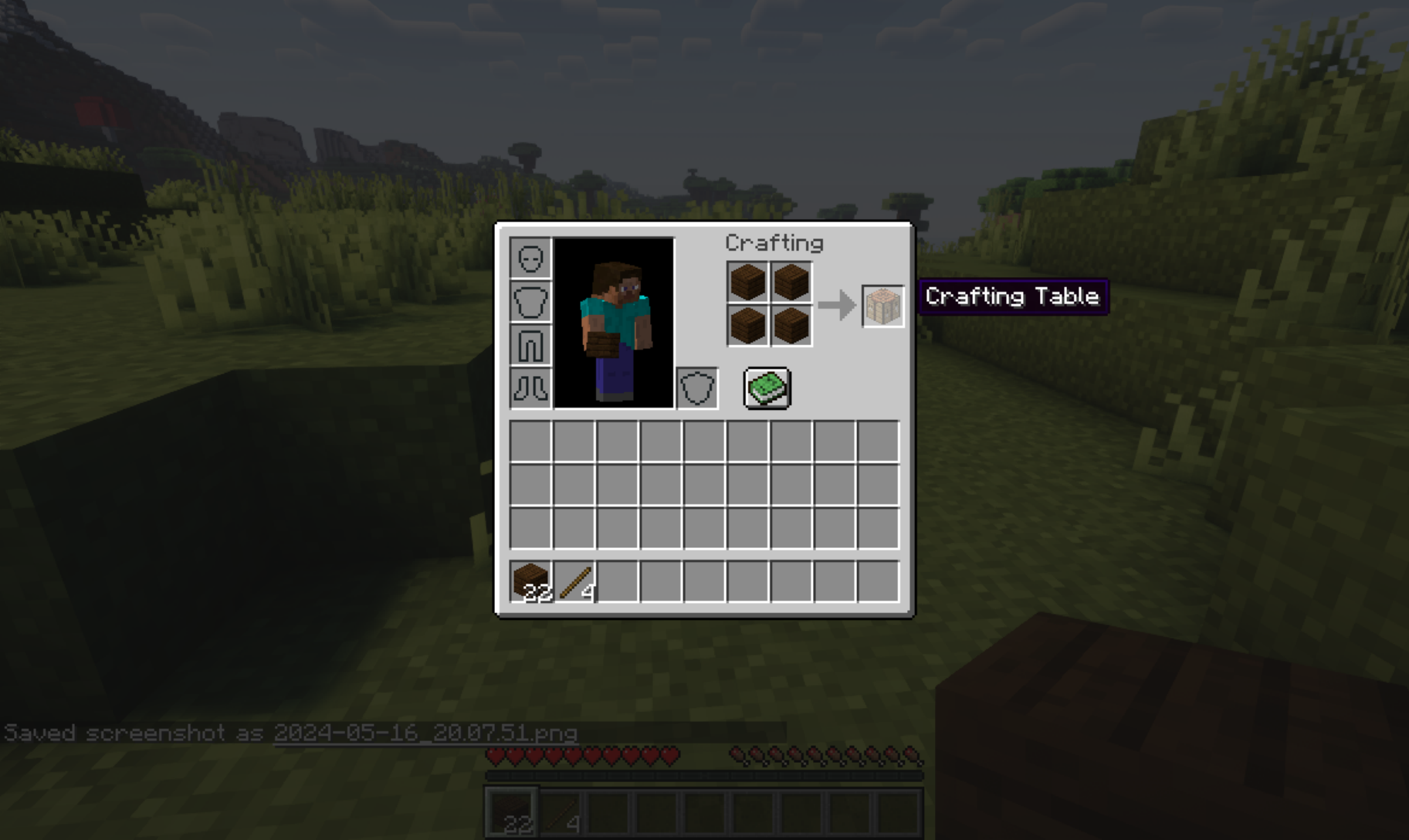}
    }
    \hfill
    \subfloat[Craft $1$ wooden pickaxe\label{example:fig4}]{%
      \includegraphics[width=0.3\textwidth]{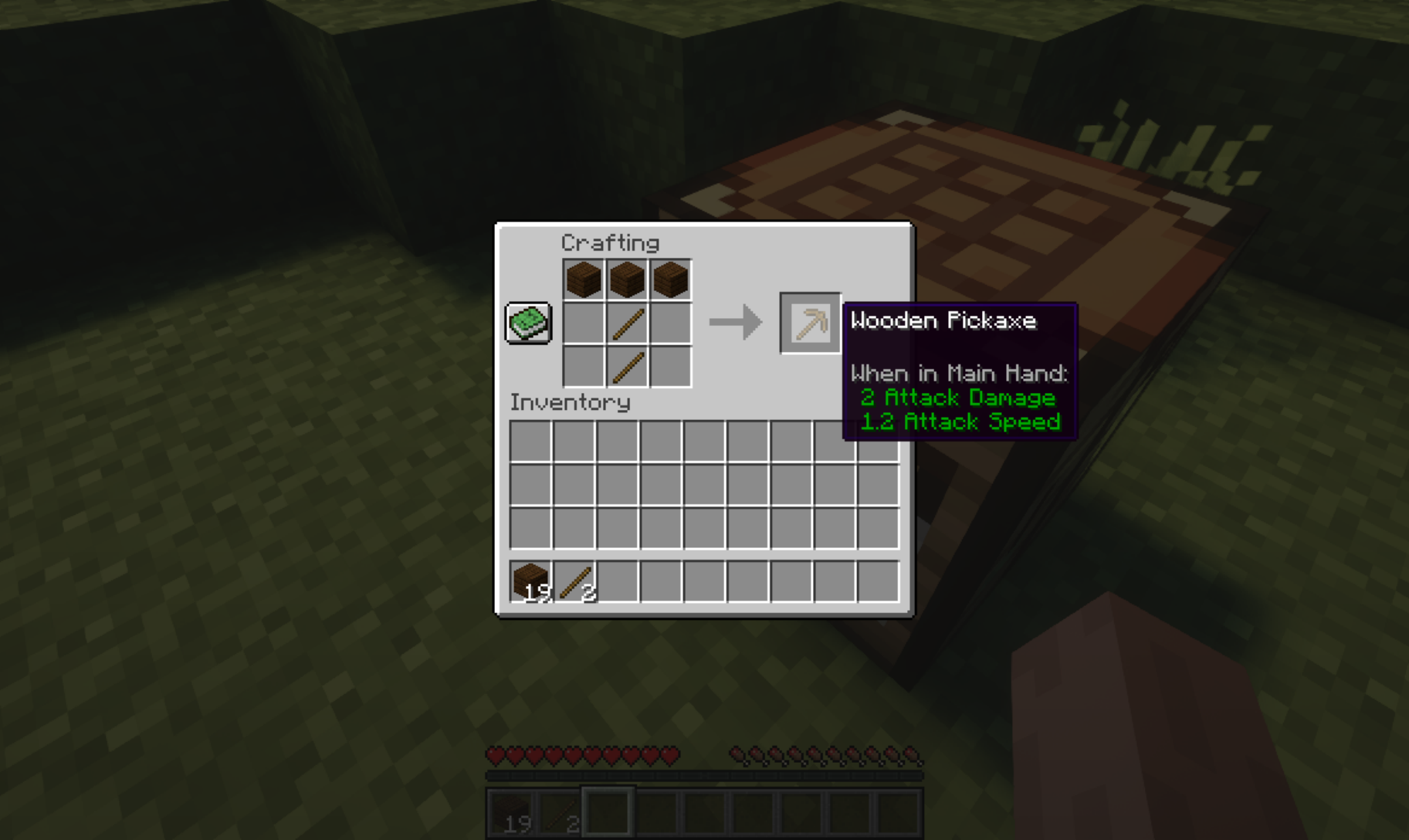}
    }
    \hfill
    \subfloat[Mine $11$ cobblestone\label{example:fig5}]{%
      \includegraphics[width=0.3\textwidth]{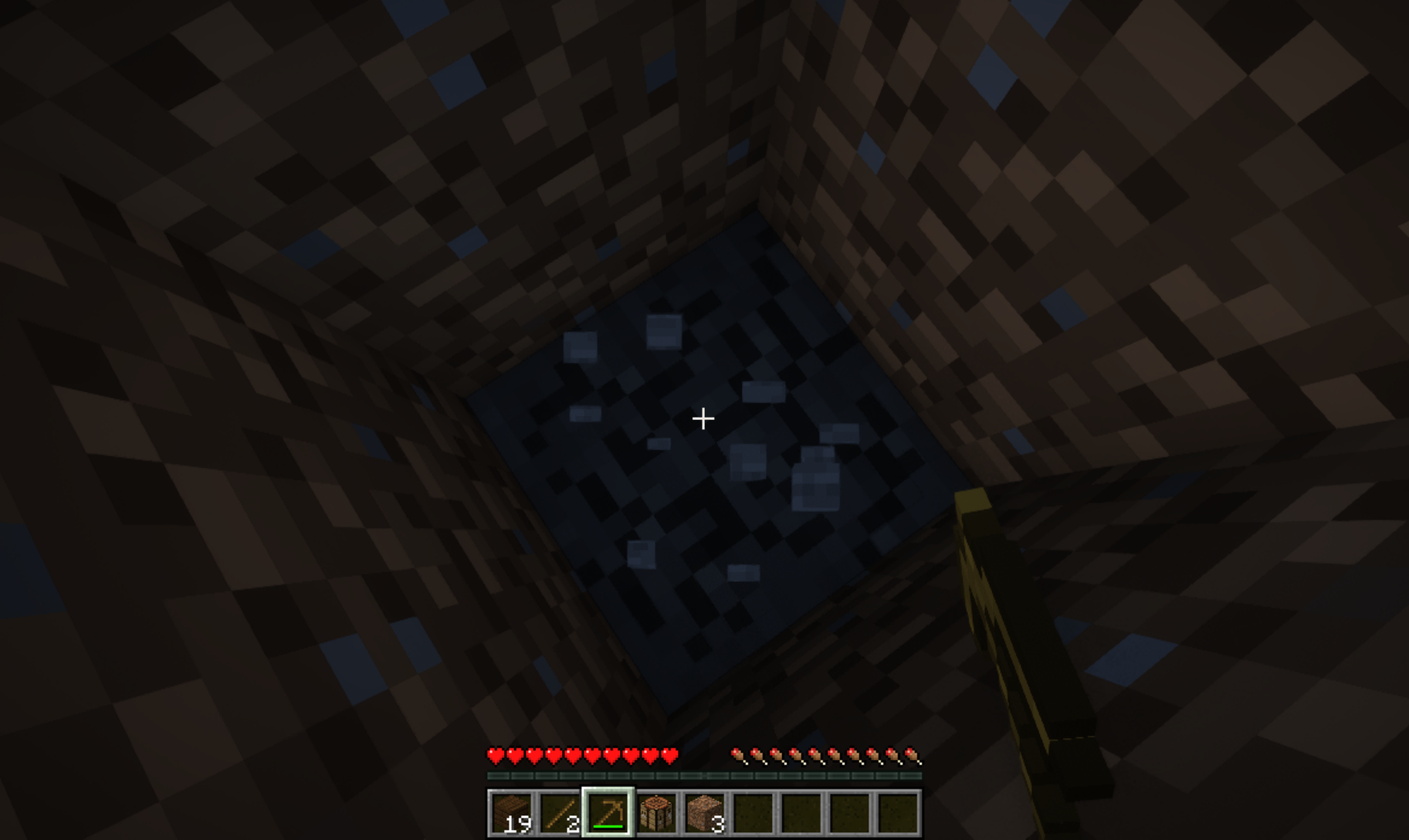}
    }
    \hfill
    \subfloat[Craft $1$ furnace\label{example:fig6}]{%
      \includegraphics[width=0.3\textwidth]{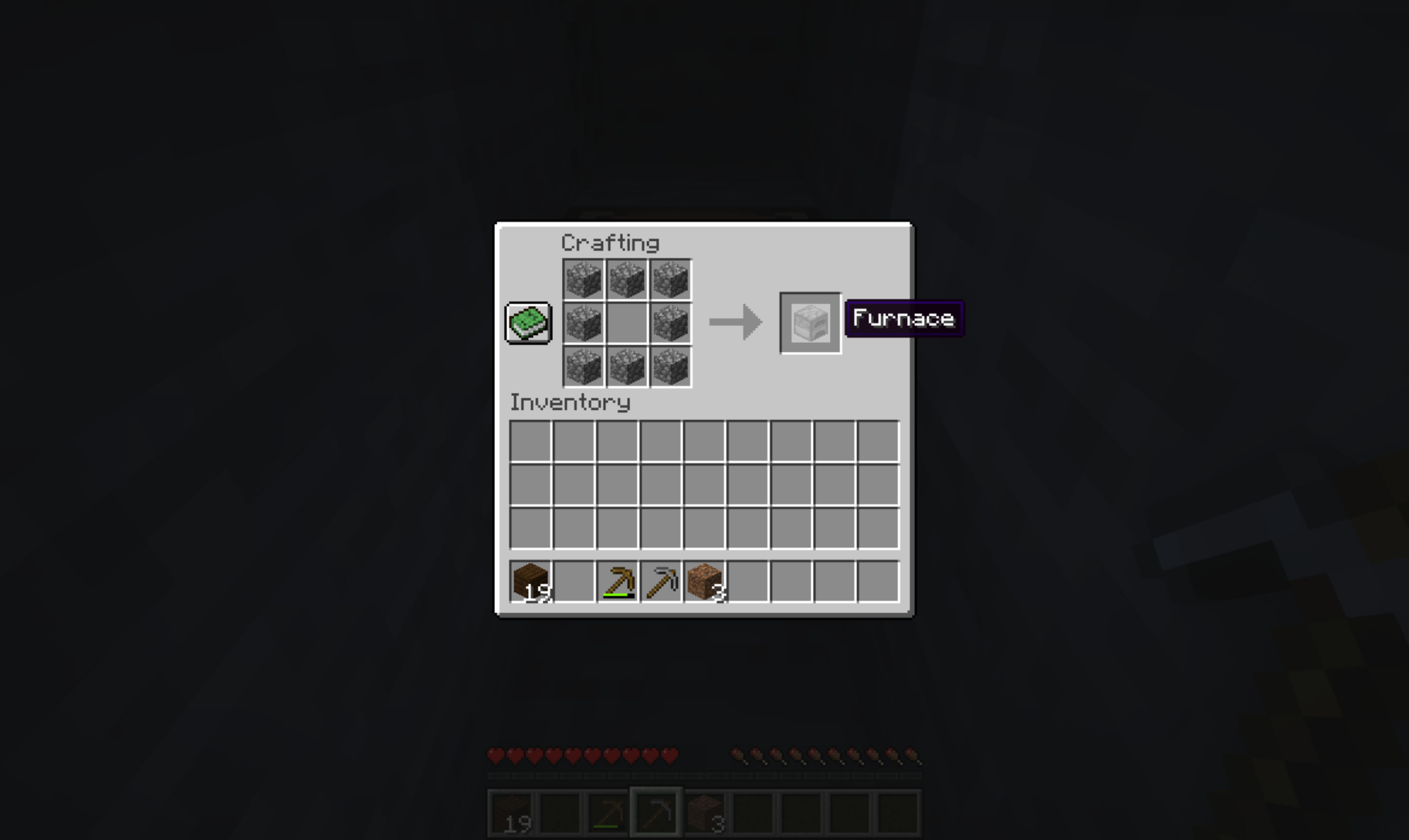}
    }
    \hfill
    \subfloat[Craft $1$ stone pickaxe\label{example:fig7}]{%
      \includegraphics[width=0.3\textwidth]{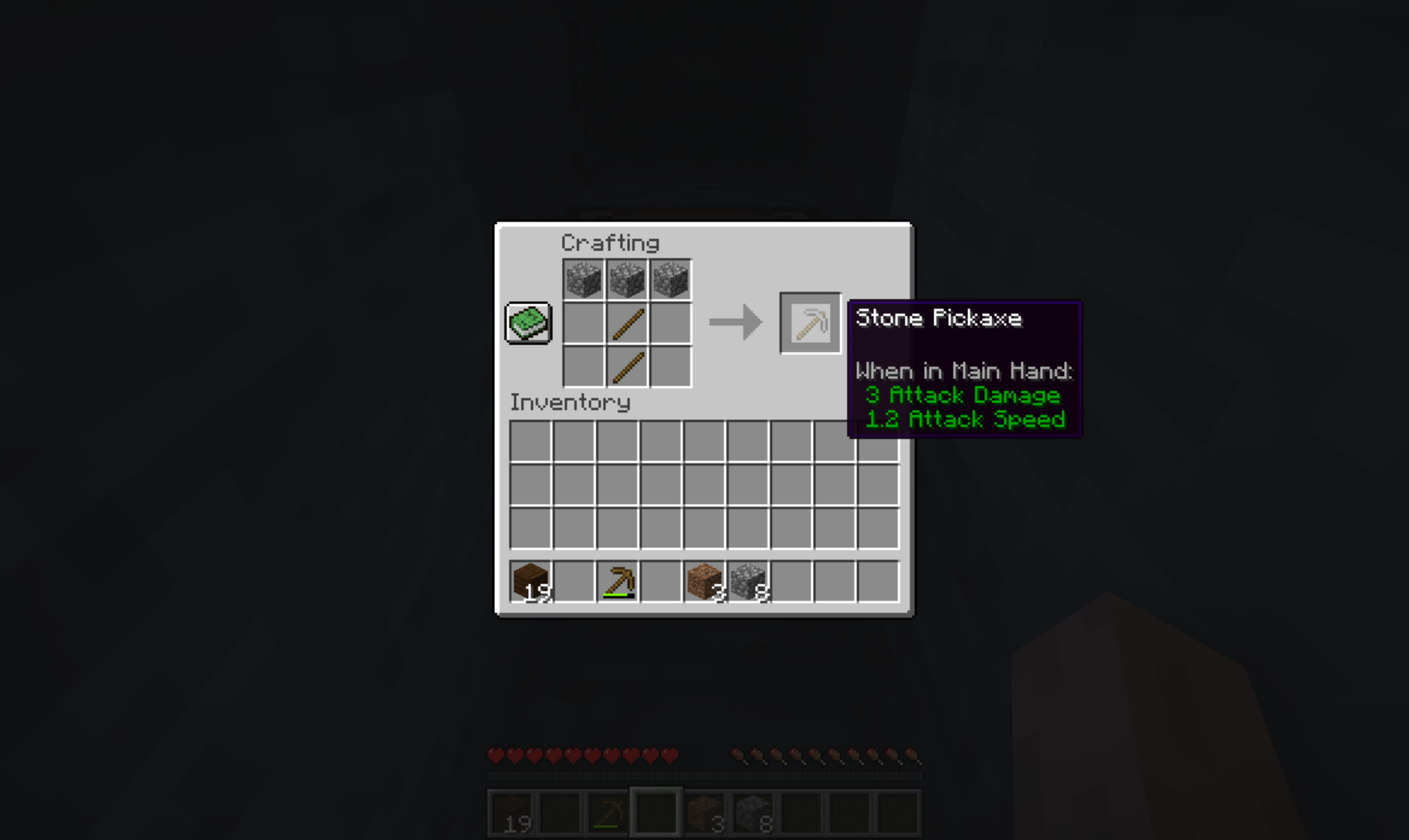}
    }
    \hfill
    \subfloat[Dig down more deeper to find iron ore\label{example:fig12}]{%
      \includegraphics[width=0.3\textwidth]{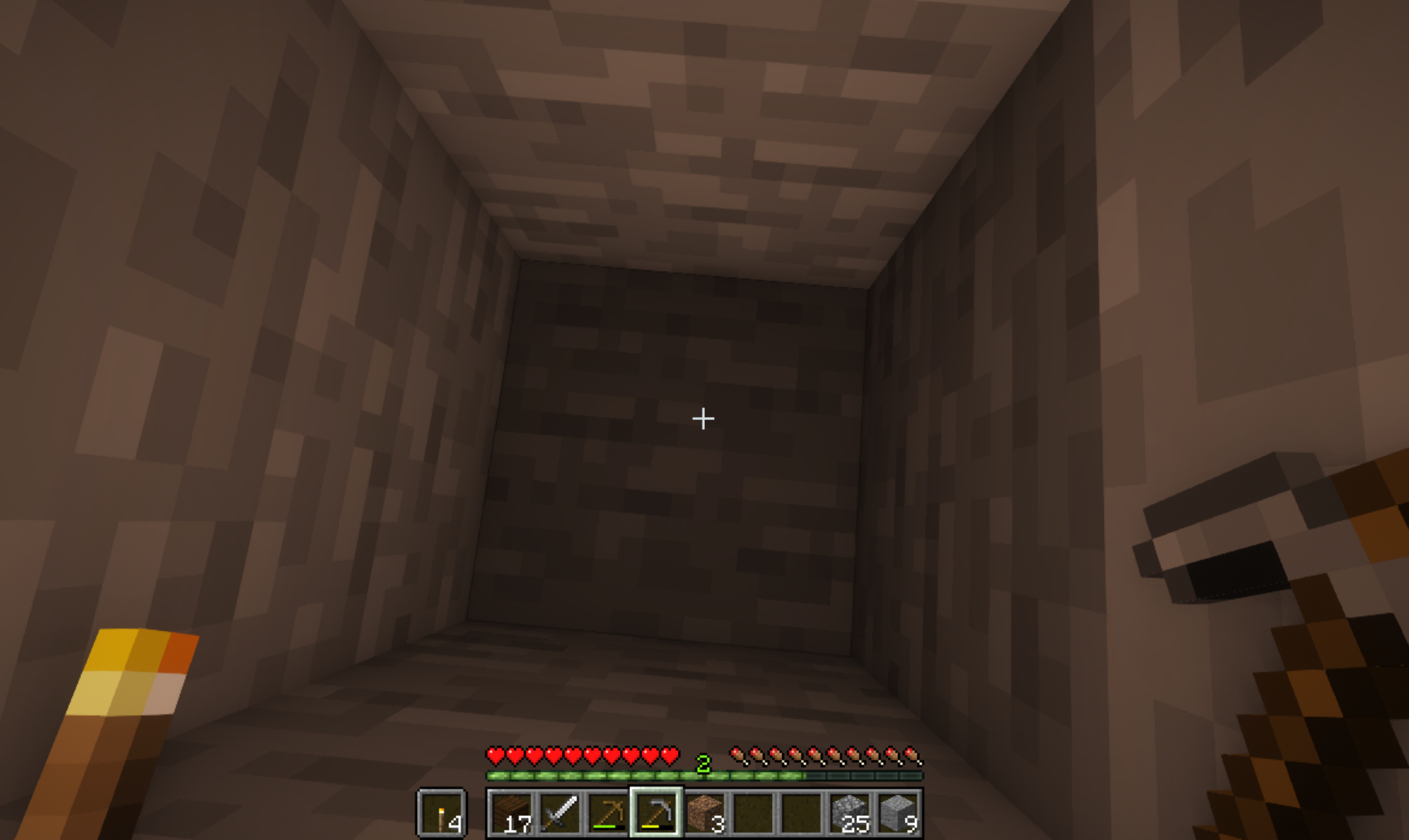}
    }
    \hfill
    \subfloat[Mine $2$ iron ores\label{example:fig8}]{%
      \includegraphics[width=0.3\textwidth]{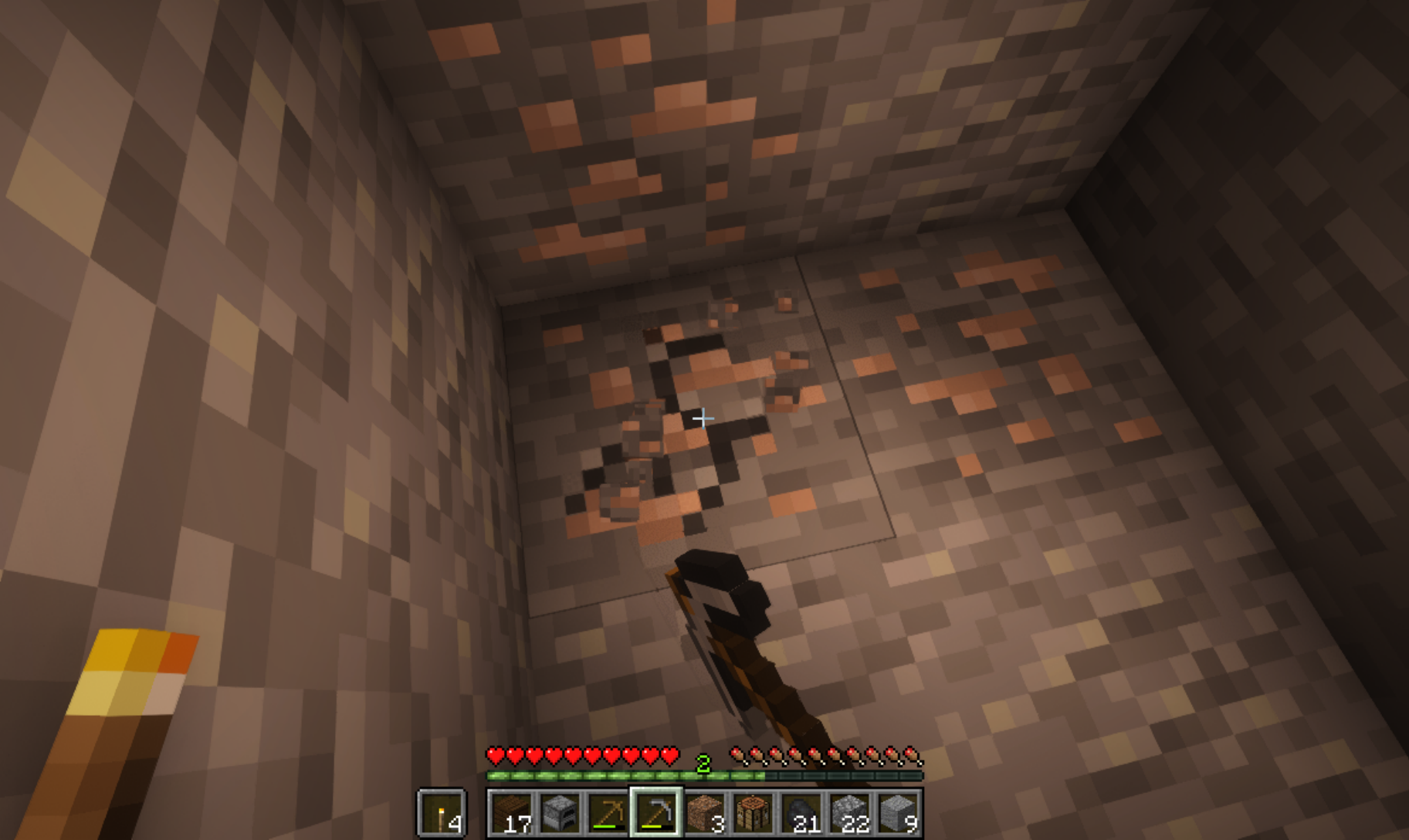}
    }
    \hfill
    \subfloat[Smelt $2$ iron ingots\label{example:fig9}]{%
      \includegraphics[width=0.3\textwidth]{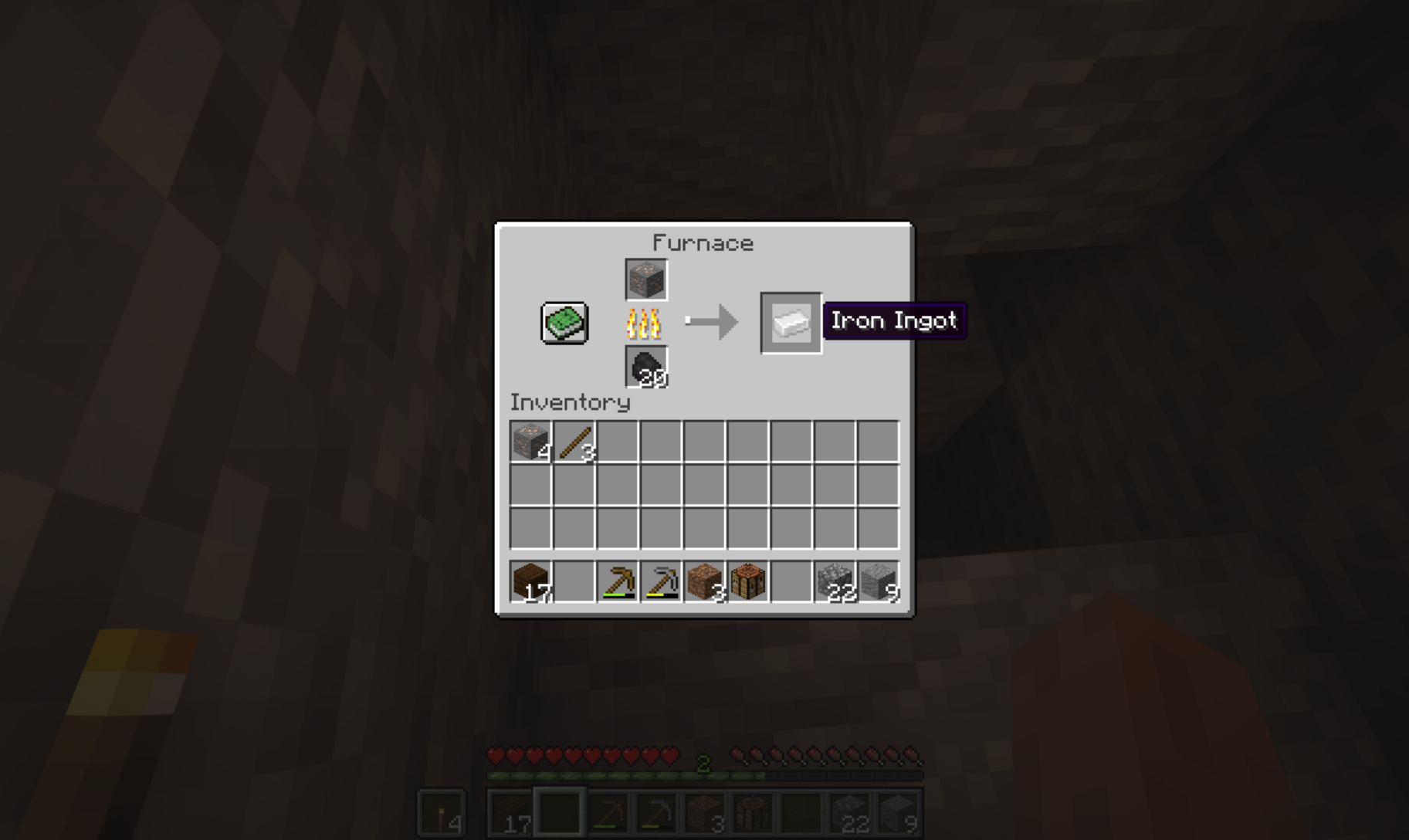}
    }
    \hfill
    \subfloat[Craft $1$ iron sword\label{example:fig10}]{%
      \includegraphics[width=0.3\textwidth]{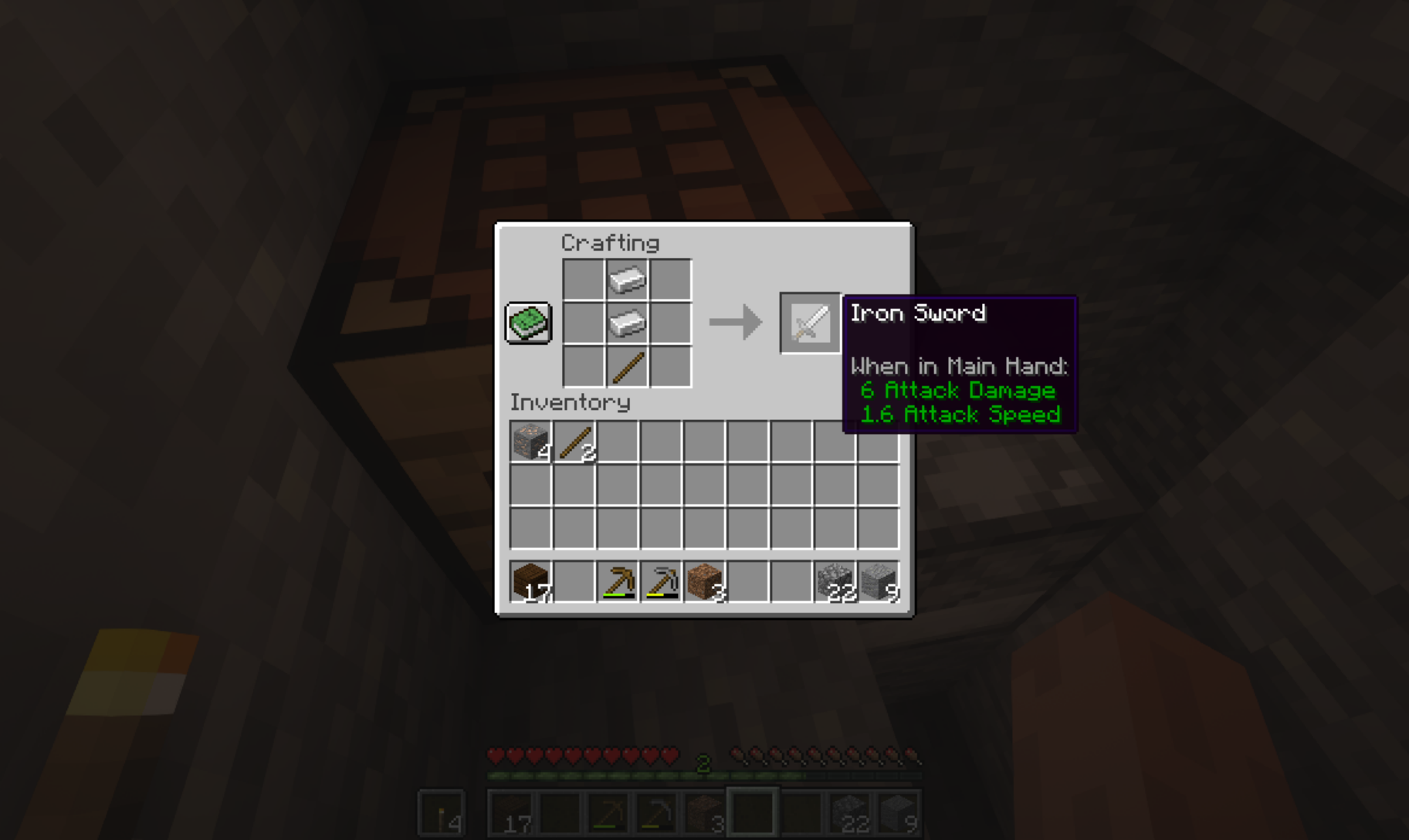}
    }
    \caption{Processing of task "Craft $1$ iron sword". Optimus-1 needs thousands of steps to complete this task. To craft and smelt precisely, the mouse movements action can't have any error.}
    \label{fig:example_long_horizon_task1}
\end{figure}

%% file: table/eval_task.tex
\begin{table}[]
\centering
\caption{Setting of 7  groups encompassing 67 Minecraft long-horizon tasks.}
\label{tab:task}
\resizebox{\textwidth}{!}{%
\renewcommand\arraystretch{1.2}
\begin{tabular}{lccccc}
\toprule[1.5pt]
Group    & Task Num. & Example Task                          & Max. Steps & Initial Inventory & Avg. Sub-goal Num. \\ \hline
\includegraphics[width=0.4cm]{picture/logo/wood.pdf} Wooden   & 10        & Craft a wooden axe                    & 3600       & Empty             & 4.60               \\
\includegraphics[width=0.4cm]{picture/logo/cobblestone.pdf} Stone    & 9         & Craft one stone pickaxe               & 7200       & Empty             & 8.78               \\
\includegraphics[width=0.4cm]{picture/logo/iron_ingot.pdf} Iron     & 16        & Craft a iron pickaxe                  & 12000      & Empty             & 12.75              \\
\includegraphics[width=0.4cm]{picture/logo/gold_ingot.pdf} Golden   & 6         & Mine gold and smelt into golden ingot & 36000      & Empty             & 15.83              \\
\includegraphics[width=0.4cm]{picture/logo/readstone.pdf} Redstone & 6         & Craft a piston                        & 36000      & Empty             & 17.50              \\
\includegraphics[width=0.4cm]{picture/logo/diamond.pdf} Diamond  & 7         & Dig down and mine a diamond           & 36000      & Empty             & 15.14              \\
\includegraphics[width=0.4cm]{picture/logo/armor.pdf} Armor    & 13        & Craft one iron helmet                 & 36000      & Empty             & 15.85              \\ 
\bottomrule[1.5pt]
\end{tabular}%
}
\end{table}

%% file: table/abl_eval_task.tex
\begin{table}[]
\centering
\caption{We evaluate Optimus-1 on these tasks in ablation study which are the subset of our benchmark.}
\label{tab:abl_eval_task}
\resizebox{\textwidth}{!}{%
\renewcommand\arraystretch{1.2}
\begin{tabular}{cccccc}
\toprule[1.5pt]
Group & Task & Sub-Goal Num. & Max. Step & Initial Inventory \\ \hline
\multirow{2}{*}{ \includegraphics[width=0.4cm]{picture/logo/wood.pdf} Wooden}  & Craft a wooden axe          & 5  & 3600  & Empty \\
                         & Craft a crafting table      & 3  & 3600  & Empty \\ \hline
\multirow{3}{*}{\includegraphics[width=0.4cm]{picture/logo/cobblestone.pdf} Stone }   & Craft a stone pickaxe       & 10 & 7200  & Empty \\
                         & Craft a stone axe           & 10 & 7200  & Empty \\
                         & Craft a furnace             & 9  & 7200  & Empty \\ \hline
\multirow{5}{*}{ \includegraphics[width=0.4cm]{picture/logo/iron_ingot.pdf} Iron }    & Craft a iron pickaxe        & 13 & 12000 & Empty \\
                         & Craft a bucket              & 13 & 12000 & Empty \\
                         & Craft a rail                & 13 & 12000 & Empty \\
                         & Craft a iron sword          & 12 & 12000 & Empty \\
                         & Craft a shears              & 12 & 12000 & Empty \\ \hline
\multirow{3}{*}{ \includegraphics[width=0.4cm]{picture/logo/gold_ingot.pdf} Golden }  & Craft a golden pickaxe      & 16 & 36000 & Empty \\
                         & Craft a golden axe          & 16 & 36000 & Empty \\
                         & Smelt a golden ingot        & 15 & 36000 & Empty \\ \hline
\multirow{5}{*}{\includegraphics[width=0.4cm]{picture/logo/diamond.pdf} Diamond } & Craft a diamond pickaxe     & 15 & 36000 & Empty \\
                         & Craft a diamond axe         & 16 & 36000 & Empty \\
                         & Craft a diamond hoe         & 15 & 36000 & Empty \\
                         & Craft a diamond sword       & 15 & 36000 & Empty \\
                         & Dig down and mine a diamond & 15 & 36000 & Empty \\
\bottomrule[1.5pt]
\end{tabular}%
}
\end{table}

%% file: table/compare_baseline.tex
\begin{table}[]
\centering
\caption{Statistics for various Minecraft agents.}
\label{tab:compare_baseline}
\resizebox{\textwidth}{!}{%
\renewcommand\arraystretch{1.2}
\begin{tabular}{lcccccccc}
\hline
Agent     & Pub.         & Env.       & Input & Output           & Planning                  & Reflection                & Knowledge                 & Experience                \\ \hline
VPT \cite{vpt}       & NeurIPS' 22 & MineRL     & V     & low-level action &     &     &     &     \\
MineDOJO \cite{fan2022minedojo}  & NeurIPS' 22 & MineDOJO   & T+V   & low-level action &     &     &     &     \\
STEVE-1 \cite{lifshitz2024steve}   & NeurIPS' 23 & MineRL     & T+V   & low-level action &     &     &     &     \\
Voyager \cite{wang2023voyager}   & NeurIPS' 23 & Mineflayer & T+V   & code             & \Checkmark & \Checkmark &     & \Checkmark \\
DEPS \cite{wang2023describe}     & NeurIPS' 23 & MineDOJO   & T+V   & code             & \Checkmark & \Checkmark &     &     \\
GROOT \cite{cai2023groot}    & ICLR' 24    & MineRL     & T+V   & low-level action &     &     &     &     \\
MP5  \cite{qin2023mp5}     & CVPR' 24    & MineDOJO   & T+V   & code             & \Checkmark & \Checkmark &     &     \\
Jarvis-1 \cite{wang2023jarvis} & -            & MineRL      & T+V   & low-level action & \Checkmark & \Checkmark &     & \Checkmark \\
\rowcolor[HTML]{fdf9ea}
Optimus-1 & -            & MineRL     & T+V   & low-level action & \Checkmark & \Checkmark & \Checkmark & \Checkmark \\ \hline
\end{tabular}
}
\end{table}

%% file: picture/appendix/prompt_plan.tex
\definecolor{codeblue}{rgb}{0.25,0.5,0.5}
\definecolor{codekw}{rgb}{0.85, 0.18, 0.50}
\definecolor{keywordgreen}{rgb}{0,0.6,0}
\lstset{
  backgroundcolor=\color{gray!10},
  basicstyle=\fontsize{8pt}{9pt}\ttfamily\selectfont,
  columns=fullflexible,
  breaklines=true,
  captionpos=b,
  commentstyle=\fontsize{6.5pt}{7.5pt}\color{codeblue},
  keywords = {User, Assistant, System}, 
  keywordstyle = {\textbf},
  caption={Prompt for Knowledge-Guided Planner.},
  label={lst:planning_prompt}
}

\begin{lstlisting} % [language=python]  

System: You are a MineCraft game expert and you can guide agents to complete complex tasks. 
User: For a given game screen and task, you need to complete <goal inference> and <visual inference>.
<goal inference>: According to the task, you need to infer the weapons, equipment, or materials required to complete the task.
<visual inference>: According to the game screen, you need to infer the following aspects: health bar, food bar, hotbar, environment.
I will give you an example as follow:
[Example]
<task>: craft a stone sword.
<goal inference>: stone sword
<visual inference>
health bar: full
food bar: full
hotbar: empty
environment: forest
Here is a game screen and task, you MUST output in example format.
<task>: {task}.
<game screen>: {image}

Assistant: 

==============================
User: Now you need to make a plan with the help of <visual info> and <craft graph>.
<visual info>: Consists of the following aspects: health bar, food bar, hotbar, environment. Based on the current visual information, you need to consider whether prequel steps needed to ensure that agent can complete the task.
<craft graph>: a top-down list of all the tools and materials needed to complete the task. 
I will give you an example of planning under specific visual conditions as follow:
[Example]
{example}
Here is a game screen and task, you MUST output in example format. Remember <task planning> MUST output in example format.
<task>: {task}
<game screen>: {image}
<craft graph>: {graph}
Assistant:
\end{lstlisting}

%% file: picture/appendix/prompt_reflect.tex
\definecolor{codeblue}{rgb}{0.25,0.5,0.5}
\definecolor{codekw}{rgb}{0.85, 0.18, 0.50}
\definecolor{keywordgreen}{rgb}{0,0.6,0}
\lstset{
  backgroundcolor=\color{gray!10},
  basicstyle=\fontsize{8pt}{9pt}\ttfamily\selectfont,
  columns=fullflexible,
  breaklines=true,
  captionpos=b,
  commentstyle=\fontsize{6.5pt}{7.5pt}\color{codeblue},
  keywords = {User, Assistant, System}, 
  keywordstyle = {\textbf},
  caption={Prompt for Experience-Driven Reflector.},
  label={lst:planning_prompt}
}

\begin{lstlisting} % [language=python]  

System: You are a MineCraft game expert and you can guide agents to complete complex tasks. Agent is executing the task: {task}.
Given two images about agent's state before executing the task and its current state, you should first detection the environment (forest, cave, ocean, etc.,) in which the agent is located, then determine whether the agent's current situation is done, continue, or replan.
<done>: Comparing the image before the task was performed, the current image reveals that the task is complete.
<continue>: Current image reveals that the task is NOT complete, but agent is in good state (good health, not hungry) with high likelihood to complete task.
<replan>: Current image reveals that the task is NOT complete, and agent is in bad state (bad health, or hungry) or situation (in danger, or in trouble), need for replanning. For replan, you need to further determine whether the agent's predicament is "drop_down" or "in_water". "drop_down" means that the agent has fallen into a cave or is trapped in a mountain or river, while "in_water" means that the agent is in the ocean and needs to return to land immediately.

User: I'll give you some examples to illustrate the different situations. Each example consists of two images, where the first image is the state of the agent before performing the task and the second image is the current state of the agent.

[Examples]
<done>: {image1},{image2}
<continue>: {image1},{image2}
<replan>: {image1},{image2}

Now given two images about agent's state before executing the task and its current state, you MUST and ONLY output in following format:
Enviroment: <environment>
Situation: <situation>
(if situation is replan) Predicament: <predicament>
\end{lstlisting}

%% file: table/our_benchmark_result/wooden_result.tex

\begin{table}[]
\centering
\caption{The results of Optimus-1 on various tasks in the Wood group. SR, AS, AT denote success rate, average number of steps, and average time (seconds), respectively.}
\label{tab:wooden_result}
\resizebox{\textwidth}{!}{%
\renewcommand\arraystretch{1.2}
\begin{tabular}{cccccc}
\toprule[1.5pt]
\textbf{Task} & \textbf{Sub-Goal Num.} & \textbf{SR} & \textbf{AS} & \textbf{AT(s)} & \textbf{Eval Times} \\ \hline
Craft a wooden shovel  & 6 & 95.00  & 995.58  & 49.78 & 40 \\
Craft a wooden pickaxe & 5 & 100.00 & 1153.91 & 57.70 & 30 \\
Craft a wooden axe     & 5 & 96.67  & 1010.28 & 50.51 & 30 \\
Craft a wooden hoe     & 5 & 100.00 & 1042.80 & 52.14 & 30 \\
Craft a stick          & 4 & 97.14  & 372.97  & 18.65 & 70 \\
Craft a crafting table & 3 & 98.55  & 448.63  & 22.43 & 69 \\
Craft a wooden sword   & 5 & 100.00 & 1214.90 & 60.74 & 30 \\
Craft a chest          & 4 & 100.00 & 573.80  & 28.69 & 30 \\
Craft a bowl           & 4 & 100.00 & 744.30  & 37.21 & 30 \\ \bottomrule[1.5pt]
\end{tabular}%
}
\end{table}

%% file: table/our_benchmark_result/stone_result.tex

\begin{table}[]
\centering
\caption{The results of Optimus-1 on various tasks in the Stone group. SR, AS, AT denote success rate, average number of steps, and average time (seconds), respectively.}
\label{tab:stone_result}
\resizebox{\textwidth}{!}{%
\renewcommand\arraystretch{1.2}
\begin{tabular}{cccccc}
\toprule[1.5pt]
\textbf{Task} & \textbf{Sub-Goal Num.} & \textbf{SR} & \textbf{AS} & \textbf{AT(s)} & \textbf{Eval Times} \\ \hline
Craft a stone shovel  & 8  & 90.32 & 2221.00 & 111.05 & 31 \\
Craft a stone pickaxe & 10 & 96.77 & 2310.09 & 115.50 & 31 \\
Craft a stone axe     & 10 & 96.88 & 2112.59 & 105.63 & 32 \\
Craft a stone hoe     & 8  & 94.64 & 2684.60 & 134.23 & 56 \\
Craft a charcoal      & 9  & 88.57 & 3083.35 & 154.17 & 35 \\
Craft a smoker        & 9  & 90.24 & 3118.89 & 155.94 & 41 \\
Craft a stone sword   & 8  & 94.29 & 2067.92 & 103.40 & 35 \\
Craft a furnace       & 9  & 93.75 & 2842.71 & 142.14 & 32 \\
Craft a torch         & 8  & 85.71 & 2109.00 & 105.45 & 95 \\ \bottomrule[1.5pt]
\end{tabular}%
}
\end{table}

%% file: table/our_benchmark_result/iron_result.tex
\begin{table}[]
\centering
\caption{The results of Optimus-1 on various tasks in the Iron group. SR, AS, AT denote success rate, average number of steps, and average time (seconds), respectively.}
\label{tab:iron_result}
\resizebox{\textwidth}{!}{%
\renewcommand\arraystretch{1.2}
\begin{tabular}{cccccc}
\toprule[1.5pt]
\textbf{Task} & \textbf{Sub Goal Num.} & \textbf{SR} & \textbf{AS} & \multicolumn{1}{c}{\textbf{AT(s)}} & \textbf{Eval Times} \\ \hline
Craft an iron shovel   & 13 & 54.79 & 5677.35 & 637.81  & 73 \\
Craft an iron pickaxe  & 13 & 59.42 & 6157.39 & 591.81  & 69 \\
Craft an iron axe      & 13 & 54.29 & 6026.26 & 676.97  & 70 \\
Craft a stone\_hoe     & 13 & 52.70 & 6650.97 & 743.82  & 74 \\
Craft a bucket         & 13 & 54.29 & 6124.61 & 591.35  & 70 \\
Craft a hopper         & 14 & 46.67 & 7242.14 & 710.17  & 60 \\
Craft a rail           & 13 & 42.19 & 6713.07 & 754.48  & 64 \\
Craft an iron sword    & 12 & 57.14 & 5625.49 & 633.91  & 70 \\
Craft a shears         & 12 & 53.62 & 5058.00 & 570.35 & 69 \\
Craft a smithing table & 12 & 44.93 & 5317.39 & 594.81  & 69 \\
Craft a tripwire hook  & 13 & 48.57 & 4968.74 & 562.66  & 70 \\
Craft a chain          & 13 & 44.93 & 5764.42 & 645.33  & 69 \\
Craft an iron bars     & 12 & 42.00 & 6508.43 & 723.13  & 50 \\
Craft an iron nugget   & 12 & 30.99 & 4697.23 & 525.29  & 71 \\
Craft a blast furnace  & 14 & 25.71 & 7760.67 & 711.05  & 35 \\
Craft a stonecutter    & 13 & 34.78 & 5993.38 & 675.52  & 46 \\ \bottomrule[1.5pt]
\end{tabular}%
}
\end{table}

%% file: table/our_benchmark_result/gold_result.tex

\begin{table}[]
\centering
\caption{The results of Optimus-1 on various tasks in the Gold group. SR, AS, AT denote success rate, average number of steps, and average time (seconds), respectively.}
\label{tab:gold_result}
\resizebox{\textwidth}{!}{%
\renewcommand\arraystretch{1.2}
\begin{tabular}{cccccc}
\toprule[1.5pt]
\textbf{Task} & \textbf{Sub Goal Num.} & \textbf{SR} & \textbf{AS} & \textbf{AT(s)} & \textbf{Eval Times} \\ \hline
Craft a golden shovel         & 16 & 9.80  & 13734.75 & 686.74 & 51 \\
Craft a golden pickaxe        & 16 & 13.75 & 9672.00  & 783.60 & 80 \\
Craft a golden axe            & 16 & 4.44  & 10158.75 & 707.94 & 45 \\
Craft a golden hoe            & 16 & 3.33  & 13120.50 & 756.03 & 27 \\
Craft a golden sword          & 16 & 3.33  & 9792.00  & 789.60 & 26 \\
Smelt and craft a gold\_ingot & 15 & 16.42 & 9630.27  & 681.51 & 67 \\ \bottomrule[1.5pt]
\end{tabular}%
}
\end{table}

%% file: table/our_benchmark_result/diamond_result.tex
\begin{table}[]
\centering
\caption{The results of Optimus-1 on various tasks in the Diamond group. SR, AS, AT denote success rate, average number of steps, and average time (seconds), respectively.}
\label{tab:diamond_result}
\resizebox{\textwidth}{!}{%
\renewcommand\arraystretch{1.2}
\begin{tabular}{cccccc}
\toprule[1.5pt]
\textbf{Task} & \textbf{Sub Goal Num.} & \textbf{SR} & \textbf{AS} & \textbf{AT(s)} & \textbf{Eval Times} \\ \hline
Craft a diamond shovel      & 15 & 18.75 & 23696.75 & 1184.84 & 64 \\
Craft a diamond pickaxe     & 15 & 15.71 & 32189.50 & 1609.46 & 70 \\
Craft a diamond axe         & 16 & 4.00  & 21920.50 & 1096.03 & 75 \\
Craft a diamond hoe         & 15 & 4.61  & 24031.00 & 1201.55 & 65 \\
Craft a diamond sword       & 15 & 14.52 & 27555.50 & 1377.78 & 62 \\
Dig down and mine a diamond & 15 & 9.09  & 20782.13 & 1039.11 & 64 \\
Craft a jukebox             & 15 & 14.58 & 25056.00 & 1252.80  & 48 \\ \bottomrule[1.5pt]
\end{tabular}%
}
\end{table}

%% file: table/our_benchmark_result/redstone_result.tex

\begin{table}[]
\centering
\caption{The results of Optimus-1 on various tasks in the Redstone group. SR, AS, AT denote success rate, average number of steps, and average time (seconds), respectively.}
\label{tab:redstone_result}
\resizebox{\textwidth}{!}{%
\renewcommand\arraystretch{1.2}
\begin{tabular}{cccccc}
\toprule[1.5pt]
\textbf{Language Instruction} & \textbf{Sub-Goal Num.} & \textbf{SR} & \textbf{AS} & \textbf{AT(s)} & \textbf{Eval Times} \\ \hline
Craft a piston          & 16 & 28.57 & 6457.10  & 822.85 & 35 \\
Craft a redstone torch  & 16 & 29.63 & 6787.87  & 939.39 & 27 \\
Craft an activator rail & 18 & 15.68 & 8685.62  & 934.28 & 51 \\
Craft a compass         & 23 & 15.00 & 14908.67 & 845.43 & 40 \\
Craft a dropper         & 16 & 37.50 & 7272.80  & 1063.64 & 40 \\
Craft a note block      & 16 & 24.32 & 6727.89  & 936.39 & 37 \\ \bottomrule[1.5pt]
\end{tabular}%
}
\end{table}

%% file: table/our_benchmark_result/armor_result.tex
\begin{table}[]
\centering
\caption{The results of Optimus-1 on various tasks in the Armor group. SR, AS, AT denote success rate, average number of steps, and average time (seconds), respectively.}
\label{tab:armor_result}
\resizebox{\textwidth}{!}{%
\renewcommand\arraystretch{1.2}
\begin{tabular}{cccccc}
\toprule[1.5pt]
\textbf{Task} & \textbf{Sub Goal Num.} & \textbf{SR} & \textbf{AS} & \textbf{AT(s)} & \textbf{Eval Times} \\ \hline
Craft shield             & 14 & 43.33 & 7229.00  & 861.45 & 30 \\
Craft iron chestplate    & 14 & 47.22 & 7230.24  & 851.51 & 36 \\
Craft iron boots         & 14 & 23.81 & 6597.33  & 729.87 & 42 \\
Craft iron leggings      & 14 & 6.67  & 9279.00  & 763.95 & 30 \\
Craft iron helmet        & 14 & 58.14 & 6287.11  & 814.36 & 43 \\
Craft diamond helmet     & 17 & 2.08  & 7342.00  & 867.10 & 48 \\
Craft diamond chestplate & 17 & 2.70  & 7552.00  & 777.60 & 37 \\
Craft diamond leggings   & 17 & 9.68  & 7664.67  & 883.23 & 31 \\
Craft diamond boots      & 17 & 16.67 & 10065.60 & 803.28 & 30 \\
Craft golden helmet      & 17 & 12.50 & 11563.25 & 778.16 & 32 \\
Craft golden leggings    & 17 & 14.60 & 10107.33 & 805.37 & 41 \\
Craft golden boots       & 17 & 6.06  & 10311.00 & 915.55 & 33 \\
Craft golden chestplate  & 17 & 9.67  & 10407.58 & 820.38 & 31 \\
\bottomrule[1.5pt]
\end{tabular}%
}
\end{table}

%% file: table/other_benchmark/mp5_result.tex
\begin{table}[]
\centering
\caption{Result on Process-Dependent Tasks compared with MP5 \cite{qin2023mp5}. SR, AS, AT denote success rate, average number of steps, and average time (seconds), respectively.}
\label{tab:mp5_result}
\resizebox{\textwidth}{!}{%
\renewcommand\arraystretch{1.3}
\begin{tabular}{cc|c|ccc}
\toprule[1.3pt]
\multicolumn{2}{c|}{} &
  MP5 \cite{qin2023mp5} &
  \multicolumn{3}{c}{Optimus-1} \\
\multicolumn{2}{c|}{\multirow{-2}{*}{Task Level}} &
  SR &
  SR &
  \multicolumn{1}{c}{AS} &
  \multicolumn{1}{c}{AT(s)} \\ \hline
 &
  log &
  96.67 &
  \textbf{100.00} &
  586.58 &
  29.33 \\
 &
  sand &
  \textbf{96.67} &
  94.32 &
  1540.33 &
  77.02 \\
 &
  planks &
  96.67 &
  \textbf{100.00} &
  571.06 &
  28.55 \\
 &
  stick &
  96.67 &
  \textbf{97.14} &
  372.97 &
  18.65 \\
 &
  crafting table &
  93.33 &
  \textbf{98.55} &
  448.63 &
  22.43 \\
\multirow{-6}{*}{Basic Level} &
  \cellcolor[HTML]{EFEFEF}Average &
  \cellcolor[HTML]{EFEFEF}96.00 &
  \cellcolor[HTML]{EFEFEF}\textbf{98.00} &
  \cellcolor[HTML]{EFEFEF}703.91 &
  \cellcolor[HTML]{EFEFEF}35.20 \\ \hline
 &
  bowl &
  93.33 &
  \textbf{100.00} &
  744.30 &
  37.21 \\
 &
  boat &
  \textbf{93.33} &
  92.86 &
  1170.00 &
  58.50 \\
 &
  chest &
  90.00 &
  \textbf{100.00} &
  573.80 &
  28.69 \\
 &
  wooden sword &
  86.67 &
  \textbf{100.00} &
  1214.90 &
  60.74 \\
 &
  wooden pickaxe &
  80.00 &
  \textbf{100.00} &
  1153.91 &
  57.70 \\
\multirow{-6}{*}{Wooden Level} &
  \cellcolor[HTML]{EFEFEF}Average &
  \cellcolor[HTML]{EFEFEF}88.67 &
  \cellcolor[HTML]{EFEFEF}\textbf{98.57} &
  \cellcolor[HTML]{EFEFEF}971.38 &
  \cellcolor[HTML]{EFEFEF}48.56 \\ \hline
 &
  cobblestone &
  80.00 &
  \textbf{95.29} &
  1492.00 &
  74.60 \\
 &
  furnace &
  80.00 &
  \textbf{93.75} &
  2842.71 &
  142.14 \\
 &
  stone pickaxe &
  80.00 &
  \textbf{96.77} &
  2310.09 &
  115.50 \\
 &
  iron ore &
  \textbf{60.00} &
  50.00 &
  3017.00 &
  150.85 \\
 &
  glass &
  80.00 &
  \textbf{81.11} &
  3870.75 &
  193.54 \\
\multirow{-6}{*}{Stone Level} &
  \cellcolor[HTML]{EFEFEF}Average &
  \cellcolor[HTML]{EFEFEF}76.00 &
  \cellcolor[HTML]{EFEFEF}\textbf{83.38} &
  \cellcolor[HTML]{EFEFEF}2706.51 &
  \cellcolor[HTML]{EFEFEF}135.32 \\ \hline
 &
  iron ingot &
  56.67 &
  \textbf{59.42} &
  4697.23 &
  634.86 \\
 &
  shield $*$ &
  \textbf{56.67} &
  43.33 &
  7229.00 &
  661.45 \\
 &
  bucket &
  53.33 &
  \textbf{54.29} &
  6124.61 &
  606.23 \\
 &
  iron pickaxe &
  50.00 &
  \textbf{59.42} &
  6157.39 &
  607.87 \\
 &
  iron door &
  43.33 &
  \textbf{48.28} &
  5528.00 &
  676.40 \\
\multirow{-6}{*}{Iron Level} &
  \cellcolor[HTML]{EFEFEF}Average &
  \cellcolor[HTML]{EFEFEF}52.00 &
  \cellcolor[HTML]{EFEFEF}\textbf{52.94} &
  \cellcolor[HTML]{EFEFEF}5947.25 &
  \cellcolor[HTML]{EFEFEF}637.36 \\ \hline
 &
  diamond ore $*$ &
  \textbf{30.00} &
  9.09 &
  20782.13 &
  1039.10 \\
 &
  mind redstone &
  20.00 &
  \textbf{25.12} &
  6787.87 &
  739.39 \\
 &
  compass &
  \textbf{16.67} &
  15.00 &
  14908.67 &
  745.43 \\
 &
  diamond pickaxe &
  \textbf{23.33} &
  15.71 &
  32189.50 &
  1609.48 \\
 &
  piston &
  20.00 &
  \textbf{28.57} &
  6457.10 &
  622.85 \\
\multirow{-6}{*}{Diamond Level} &
  \cellcolor[HTML]{EFEFEF}Average &
  \cellcolor[HTML]{EFEFEF}\textbf{22.00} &
  \cellcolor[HTML]{EFEFEF}18.70 &
  \cellcolor[HTML]{EFEFEF}14963.83 &
  \cellcolor[HTML]{EFEFEF}948.19 \\ \bottomrule[1.3pt]

\end{tabular}%
}
\end{table}

%% file: table/other_benchmark/deps_result.tex
\begin{table}[]
\centering
\caption{Results (success rate) on 8 META TASK groups compared with DEPS \cite{mc-planner}.}
\label{tab:deps_result}
\resizebox{\textwidth}{!}{%
\renewcommand\arraystretch{1.1}
\begin{tabular}{cccccc}
\toprule[1.3pt]
Meta-Task &
  Task Object &
  InnerMonologue &
  Code-as-Policy &
  DEPS \cite{mc-planner} &
  Ours \\ \hline
\multirow{7}{*}{\begin{tabular}[c]{@{}c@{}}Basic\\ MT1\end{tabular}} &
  planks &
  83.3 &
  83.3 &
  83.3 &
  \textbf{100.0} \\
 & stick               & 83.3 & 83.3 & 86.7 & \textbf{97.1}  \\
 & chest               & 0.0  & 50.0 & 76.7 & \textbf{100.0} \\
 & sign                & 0.0  & 43.3 & 86.7 & \textbf{94.3}  \\
 & boat                & 26.7 & 56.7 & 73.3 & \textbf{92.9}  \\
 & trapdoor            & 56.7 & 56.7 & 76.7 & \textbf{96.2}  \\
 & bowl                & 23.3 & 46.7 & 80.0 & \textbf{100.0} \\ \hline
\multirow{10}{*}{\begin{tabular}[c]{@{}c@{}}Tool(Simple) \\ MT2\end{tabular}} &
  crafting\_table &
  70.0 &
  70.0 &
  90.0 &
  \textbf{98.5} \\
 & wooden\_pickaxe     & 80.0 & 80.0 & 80.0 & \textbf{100.0} \\
 & wooden\_sword       & 83.3 & 83.3 & 86.7 & \textbf{100.0} \\
 & wooden\_shovel      & 76.7 & 76.7 & 90.0 & \textbf{95.0}  \\
 & furnace             & 40.0 & 40.0 & 66.7 & \textbf{93.7}  \\
 & stone\_pickaxe      & 36.7 & 53.3 & 73.3 & \textbf{96.7}  \\
 & stone\_axe          & 30.0 & 30.0 & 70.0 & \textbf{96.8}  \\
 & stone\_hoe          & 36.7 & 56.7 & 66.7 & \textbf{94.6}  \\
 & stone\_shovel       & 36.7 & 36.7 & 66.7 & \textbf{90.3}  \\
 & stone\_sword        & 53.3 & 36.7 & 80.0 & \textbf{94.2}  \\ \hline
\multirow{6}{*}{\begin{tabular}[c]{@{}c@{}}Hunt and Food\\ MT3\end{tabular}} &
  bed &
  6.7 &
  6.7 &
  43.3 &
  \textbf{90.0} \\
 & painting            & 16.7 & 16.7 & 86.7 & \textbf{92.2}  \\
 & carpet              & 0.0  & 13.3 & 43.3 & \textbf{91.3}  \\
 & cooked\_porkchop    & 0.0  & 0.0  & 50.0 & \textbf{90.0}  \\
 & cooked\_beef        & 0.0  & 0.0  & 63.3 & \textbf{90.0}  \\
 & cooked\_mutton      & 0.0  & 0.0  & 66.7 & \textbf{90.0}  \\ \hline
\multirow{5}{*}{\begin{tabular}[c]{@{}c@{}}Dig-down \\ MT4\end{tabular}} &
  stone\_stairs &
  36.7 &
  16.7 &
  66.7 &
  \textbf{90.3} \\
 & stone\_slab         & 16.7 & 33.3 & 73.3 & \textbf{91.2}  \\
 & lever               & 46.7 & 46.7 & 83.3 & \textbf{91.0}  \\
 & coal                & 6.7  & 0.0  & 20.0 & \textbf{86.5}  \\
 & torch               & 6.7  & 0.0  & 13.3 & \textbf{85.7}  \\ \hline
\multirow{9}{*}{\begin{tabular}[c]{@{}c@{}}Equipment \\ MT5\end{tabular}} &
  leather\_boots &
  13.3 &
  13.3 &
  60.0 &
  \textbf{68.2} \\
 & leather\_chestplate & 0.0  & 6.7  & 36.7 & \textbf{64.2}  \\
 & leather\_helmet     & 6.7  & 0.0  & 70.0 & \textbf{65.9}  \\
 & leather\_leggings   & 20.0 & 0.0  & 56.7 & \textbf{65.5}  \\
 & iron\_chestplate    & 0.0  & 0.0  & 0.0  & \textbf{47.2}  \\
 & iron\_leggings      & 0.0  & 0.0  & 3.3  & \textbf{6.6}  \\
 & iron\_helmet        & 0.0  & 0.0  & 3.3  & \textbf{58.1}  \\
 & iron\_boots         & 0.0  & 0.0  & 20.0 & \textbf{23.8}  \\
 & shield              & 0.0  & 6.7  & 13.3 & \textbf{43.3}  \\ \hline
\multirow{7}{*}{\begin{tabular}[c]{@{}c@{}}Tool Complex \\ MT6\end{tabular}} &
  bucket &
  0.0 &
  3.3 &
  6.7 &
  \textbf{54.3} \\
 & shears              & 0.0  & 0.0  & 30.0 & \textbf{53.6}  \\
 & iron\_pickaxe       & 6.7  & 0.0  & 10.0 & \textbf{59.4}  \\
 & iron\_axe           & 0.0  & 0.0  & 16.7 & \textbf{54.3}  \\
 & iron\_hoe           & 0.0  & 0.0  & 13.3 & \textbf{52.7}  \\
 & iron\_shovel        & 0.0  & 0.0  & 13.3 & \textbf{57.8}  \\
 & iron\_sword         & 0.0  & 3.3  & 6.7  & \textbf{54.7}  \\ \hline
\multirow{5}{*}{\begin{tabular}[c]{@{}c@{}}Iron-Stage \\ MT7\end{tabular}} &
  iron\_bars &
  0.0 &
  0.0 &
  6.7 &
  \textbf{42.0} \\
 & hopper              & 0.0  & 0.0  & 6.7  & \textbf{46.7}  \\
 & iron\_door          & 0.0  & 0.0  & 3.3  & \textbf{48.3} \\
 & tripwire\_hook      & 6.7  & 0.0  & 30.0 & \textbf{48.6}  \\
 & rail                & 0.0  & 0.0  & 6.7  & \textbf{42.2}  \\ \hline
\begin{tabular}[c]{@{}c@{}}Challenge\\ MT8\end{tabular} &
  diamond &
  0.0 &
  0.0 &
  0.6 &
  \multicolumn{1}{c}{\textbf{9.1}} \\ \bottomrule[1.3pt]
\end{tabular}%
}
\end{table}

%% file: table/other_benchmark/tech-tree.tex
\begin{figure*}[!h]
    \centering
    \includegraphics[width=1.0\textwidth]{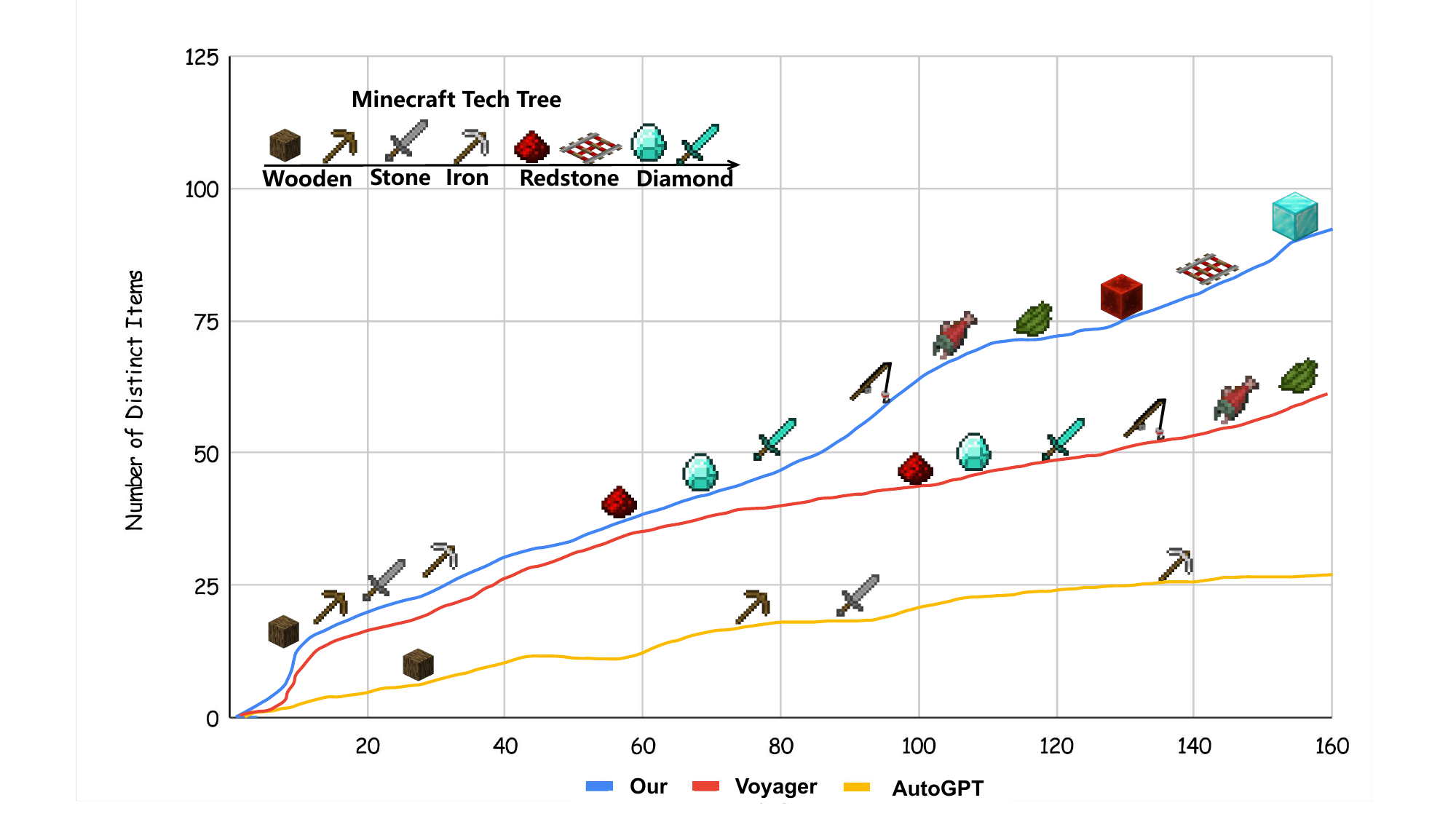}
    \caption{An illustration of Optimus-1 unlocking the tech tree in Minecraft.}
    \label{tech_tree}
\end{figure*}

%% file: picture/case-study.tex
\begin{figure}[ht]
  \centering
  \includegraphics[width=1\linewidth]{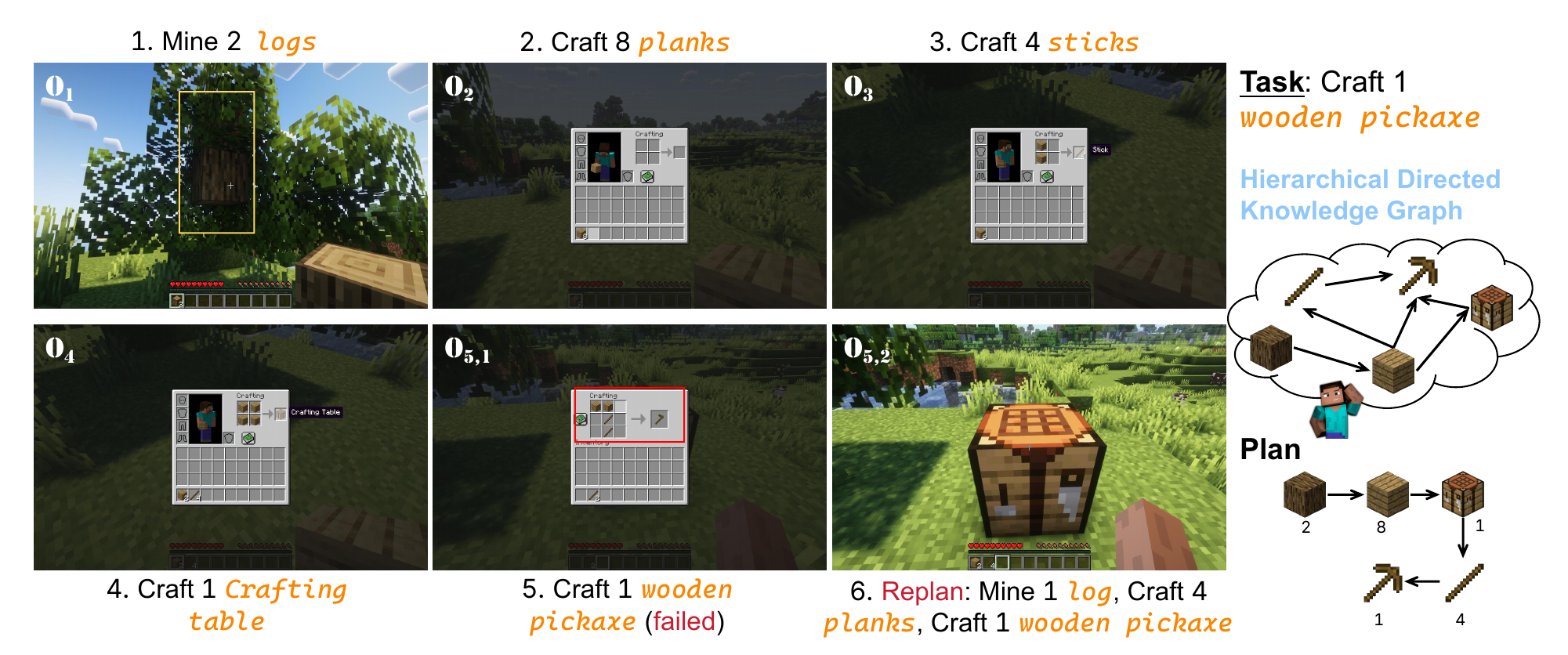}
  \caption{The process of completing the task "Craft 1 wooden pickaxe". Optimus-1 gives wrong planning. When Optimus-1 realizes it cannot complete the task, it will replan the current task. }
  \label{fig:case-replan}
\end{figure}

\begin{figure}[ht]
  \centering
  \includegraphics[width=1\linewidth]{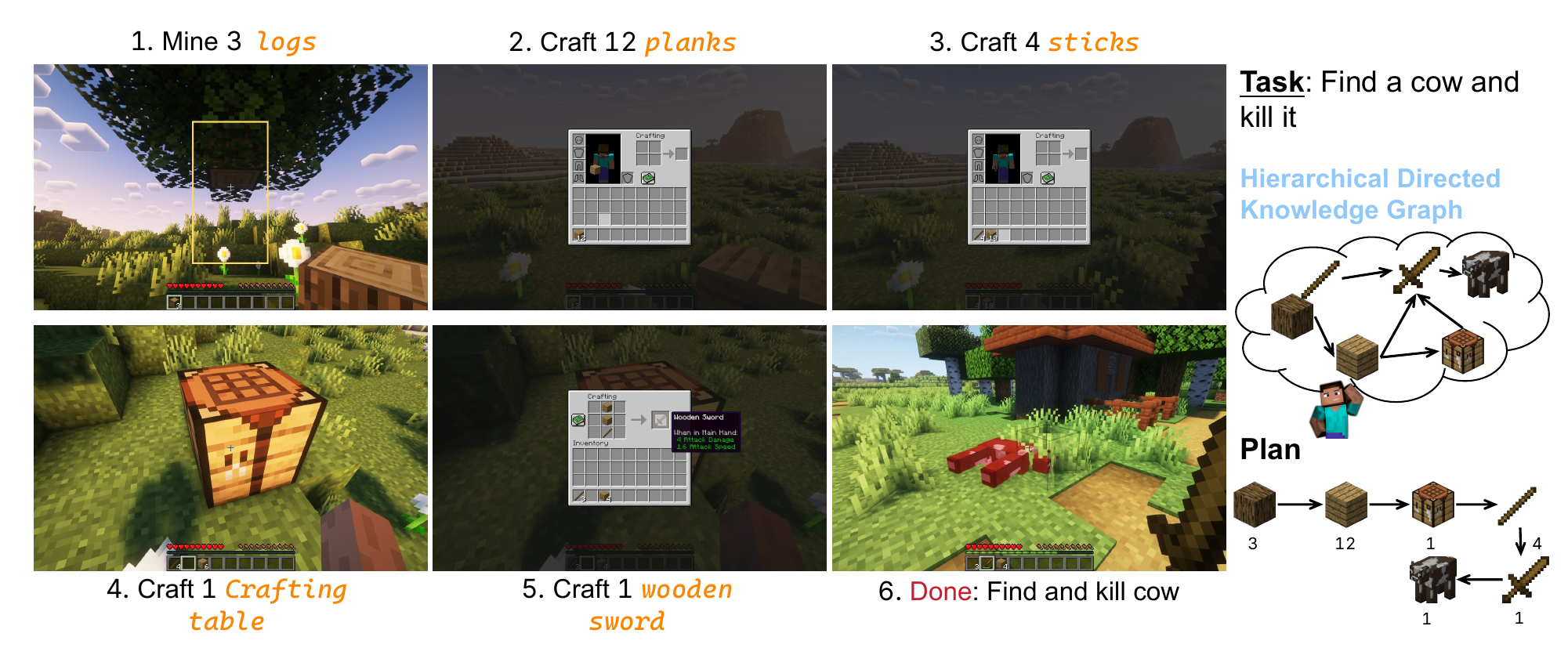}
  \caption{The process of completing the task "Find a cow and kill it". Hierarchical Directed Knowledge Graph indicates that having a wooden sword will make the task easier to complete. Therefore, Optimus-1 first crafts a wooden sword and then proceeds to find and kill a cow.}
  \label{fig:case-done}
\end{figure}

\begin{figure}[ht]
  \centering
  \includegraphics[width=1\linewidth]{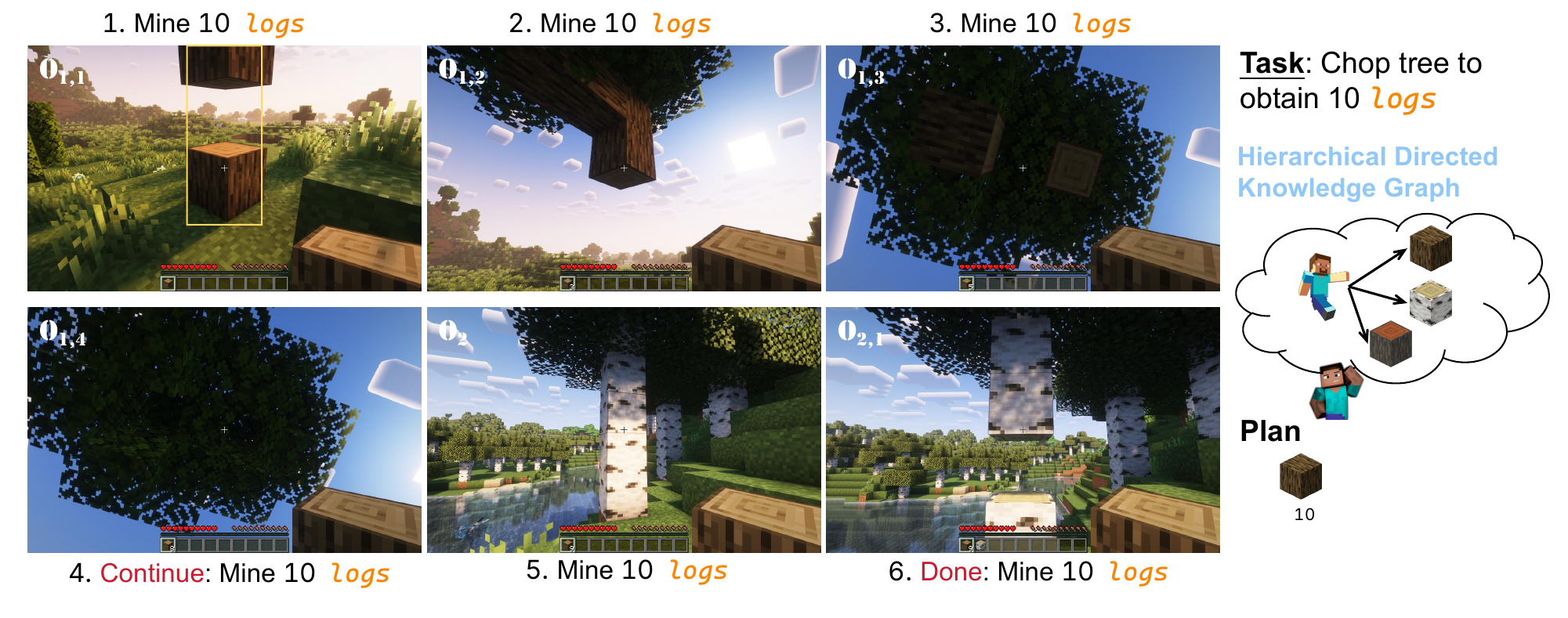}
  \caption{The process of completing the task "Chop tree to obtain 10 logs". Hierarchical Directed Knowledge graph indicates that no tools are needed to complete this goal. After finding a tree, Optimus-1 starts chopping it down. The task requires a substantial amount of wood, so midway through, Optimus-1 performs a reflection. The task is not yet complete but is progressing smoothly, and the result of the reflection is to continue.}
  \label{fig:case-continue}
\end{figure}

\begin{figure}[ht]
  \centering
  \includegraphics[width=1\linewidth]{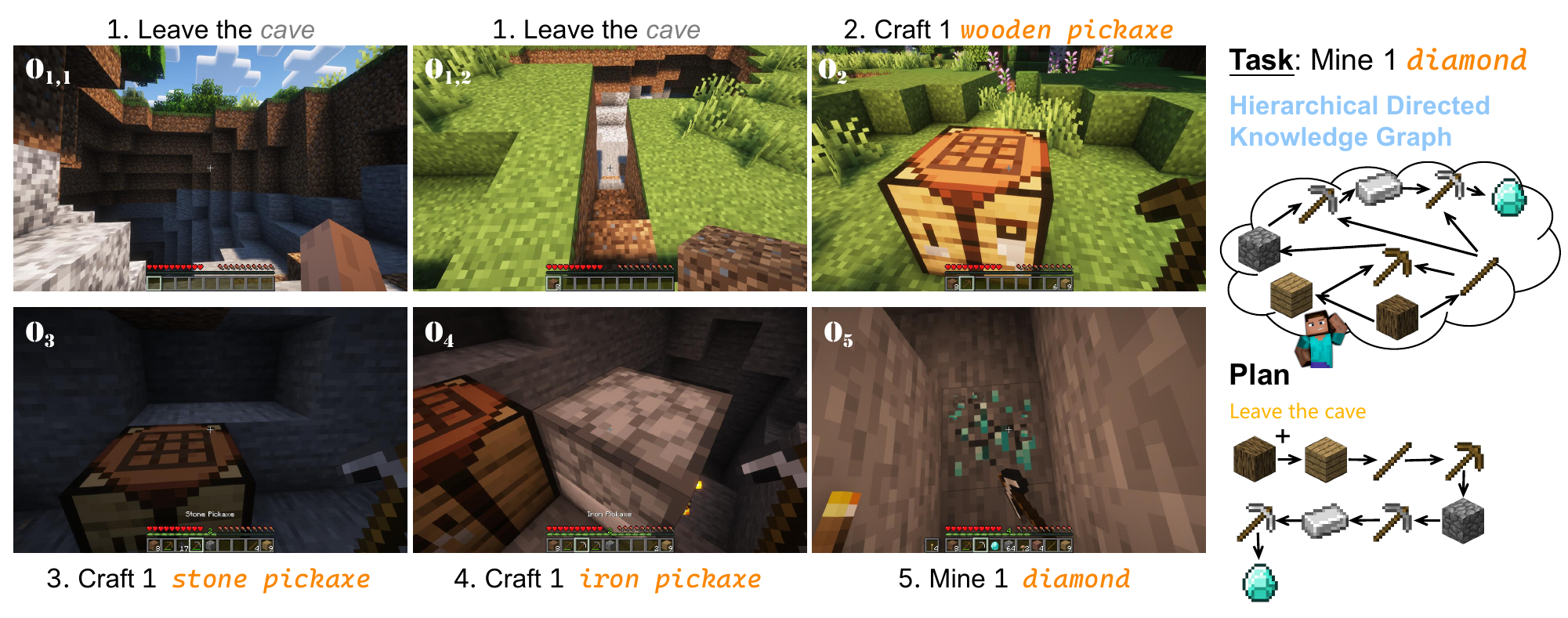}
  \caption{The process of completing the task "Mine 1 diamond \includegraphics[width=0.3cm]{picture/logo/diamond.pdf}". Mining diamonds \includegraphics[width=0.3cm]{picture/logo/diamond.pdf} is a highly complex task. Diamonds can only be mined with an iron pickaxe \includegraphics[width=0.3cm]{picture/logo/iron_pickaxe.pdf}, so an iron pickaxe \includegraphics[width=0.3cm]{picture/logo/iron_pickaxe.pdf} must be crafted first. Crafting an iron pickaxe \includegraphics[width=0.3cm]{picture/logo/iron_pickaxe.pdf} requires iron ingots \includegraphics[width=0.3cm]{picture/logo/iron_ingot.pdf}, which are smelted from iron ore \includegraphics[width=0.3cm]{picture/logo/iron_ore.pdf}. Mining iron ore \includegraphics[width=0.3cm]{picture/logo/iron_ore.pdf} requires a stone pickaxe \includegraphics[width=0.3cm]{picture/logo/stone_pickaxe.pdf}. Crafting a stone pickaxe \includegraphics[width=0.3cm]{picture/logo/stone_pickaxe.pdf} requires stone \includegraphics[width=0.3cm]{picture/logo/cobblestone.pdf}, which in turn must be mined with a wooden pickaxe \includegraphics[width=0.3cm]{picture/logo/wooden_pickaxe.pdf}. Crafting a wooden pickaxe \includegraphics[width=0.3cm]{picture/logo/wooden_pickaxe.pdf} requires wooden planks \includegraphics[width=0.3cm]{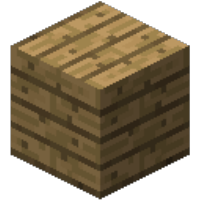}and sticks \includegraphics[width=0.3cm]{picture/logo/stick.pdf}. All these crafting processes require a crafting table \includegraphics[width=0.3cm]{picture/logo/crafting_table.pdf}, and smelting requires a furnace \includegraphics[width=0.3cm]{picture/logo/furnace.pdf}. In this case, the agent spawns at a cave, so Optimus-1 must leave the cave to chop logs.}
  \label{fig:case-plan-knowledge}
\end{figure}

\begin{figure}[ht]
  \centering
  \includegraphics[width=1\linewidth]{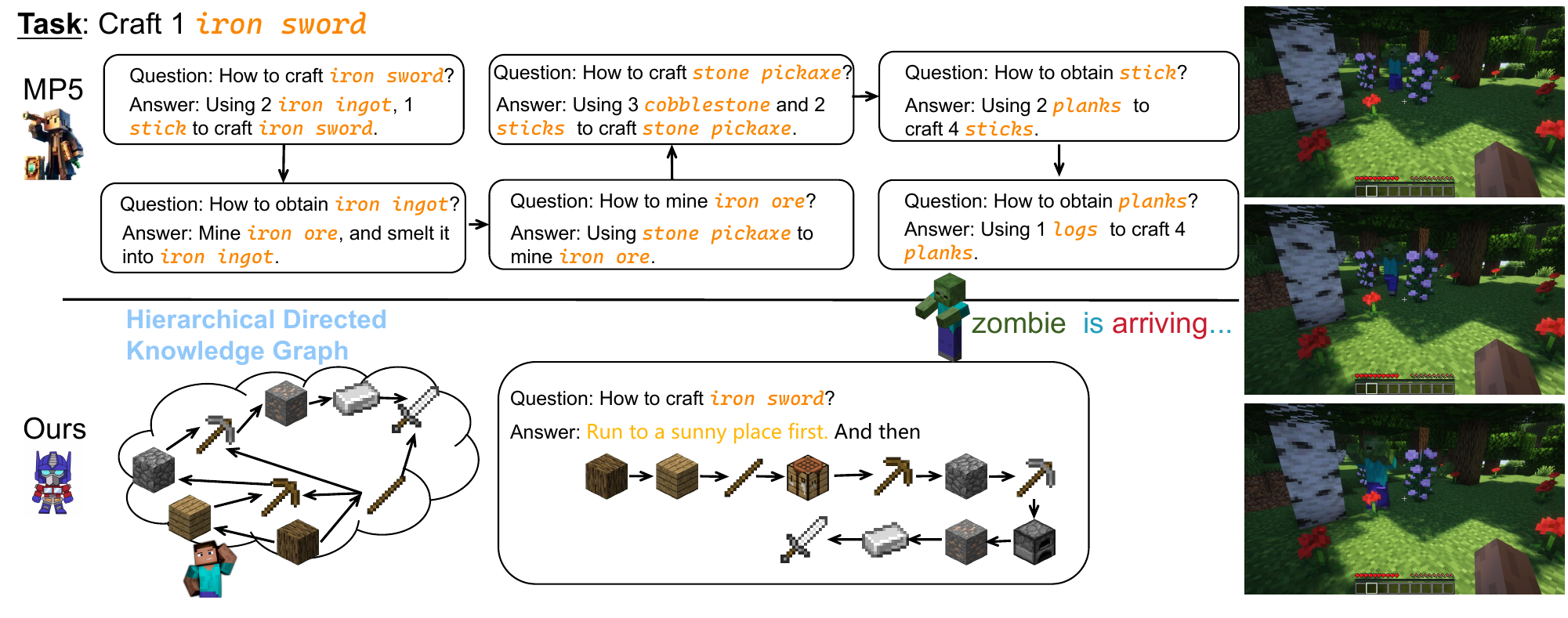}
  \caption{In this example, a zombie is slowly approaching the agent. Agents like MP5 \cite{qin2023mp5} and Voyager \cite{wang2023voyager} uses an iterative planning strategy to generate the plan, which consumes a great deal of time and puts the agent in danger. While Optimus-1 directly generates a plan in one step based on the knowledge graph. Using the current visual information, it makes a plan to "run to a sunny place," allowing the agent to avoid danger then begin to achieve sub-goals.}
  \label{fig:case-plan-one}
\end{figure}